\documentclass{article}

\usepackage[main, final]{neurips_2026}

\usepackage[utf8]{inputenc}
\usepackage[T1]{fontenc}
\usepackage{hyperref}
\usepackage{url}
\usepackage{booktabs}
\usepackage{amsfonts}
\usepackage{nicefrac}
\usepackage{amsthm}
\usepackage{microtype}
\usepackage{xcolor}
\usepackage{amsmath}
\usepackage{amssymb}
\usepackage{amsthm}
\usepackage{algorithm}
\usepackage{algorithmic}
\usepackage{bm}
\usepackage{enumitem}
\usepackage{multirow}
\usepackage{graphicx}
\usepackage{subcaption}
\usepackage{caption}
\usepackage{float}
\usepackage{adjustbox}
\usepackage[normalem]{ulem}
\usepackage{placeins}
\usepackage{svg}
\usepackage{subcaption}

\newtheorem{theorem}{Theorem}
\newtheorem{lemma}{Lemma}

\newtheorem{corollary}{Corollary}
\newtheorem{assumption}{Assumption}

\title{Reusable Conditional Resampling via Flow Matching for Constraint-Based Causal Discovery}

\author{%
Shunyu Zhao\thanks{Equal contribution.}, \ 
  Yanfeng Yang$^*$ \\
  Graduate University for Advanced Studies, SOKENDAI \& 
  The Institute of Statistical  Mathematics \\
  Tokyo, Japan \\
  \AND
  Shuai Li\thanks{Corresponding authors.} \\
  School of Mathematics and Statistics, Hunan Normal University \\
  Changsha, China \\
  \texttt{shuaili.cq@foxmail.com} 
  \AND
  Kenji Fukumizu$^\dag$ \\
  The Institute of Statistical  Mathematics \\
  Tokyo, Japan \\
  \texttt{fukumizu@ism.ac.jp}
}

\begin{document}

\maketitle

\begin{abstract}

Constraint-based causal discovery requires a large number of conditional independence (CI) tests. Generative CI tests are attractive in nonlinear and high-dimensional settings because they can model complex conditional distributions, but often suffer from high per test cost and limited reusability across the many query dependent conditional resampling tasks arising in graph search. We propose a generative nonlinear CI testing and graph search framework for PC-style causal discovery. 
We first propose the Langevin-RePaint algorithm for conditional imputation. Building on Langevin-RePaint and joint flow matching modeling, we propose the Flow Matching based Conditional Independence Test (FMCIT), which reformulates generative CI testing from query-specific conditional modeling into a reusable conditional resampling framework. We further establish theoretical guarantees for the type-I error control of FMCIT.
FMCIT allows a single trained generative model to repeatedly generate the randomized copies required for CI testing throughout the entire graph search procedure. 
We then introduce GPC‑FMCIT, which embeds FMCIT into a guided PC pipeline with screening, guidance, refinement, and orientation. By using low-cost screening to shrink the search space, constructing edge-specific candidate conditioning sets from local graph structure, and invoking FMCIT only in a budget-controlled refinement stage, GPC-FMCIT reduces both the deployment cost of individual nonlinear CI queries and the overall query complexity. On synthetic nonlinear structural equation models, GPC-FMCIT improves CPDAG recovery over PC-stable baselines and requires far fewer FMCIT queries than direct FMCIT-based PC search, and on hESC gene-expression data it yields smaller graphs than PC/FisherZ in a descriptive comparison. Overall, our results suggest that, through the joint design of reusable conditional resampling and controlled graph search, generative nonlinear CI testing can become a practical component of PC-style causal discovery.

\end{abstract}

\section{Introduction}

Conditional independence (CI) testing is a fundamental problem in statistics and machine learning, with broad applications in tasks such as causal discovery \citep{pearl1988probabilistic,spirtes2000causation} and variable selection \citep{dai2022significance,yang_2025_cdcit}. In constraint-based causal discovery, it serves as a core computational primitive. Given $p$ variables $X=(X_1,\hdots,X_p)$, where $X_i\in\mathbb{R}$ for all $i$, the goal is to test
\begin{equation}
\label{eq_cit_def}
H_0: X_i \perp \!\!\!\perp X_j \mid X_S
\quad \text{versus} \quad
H_1: X_i \not\perp\!\!\!\perp X_j \mid X_S,
\end{equation}
where $X_i$ and $X_j$ are any two variables, $S\subseteq [p]\setminus\{i,j\}$ is an index set that may be empty, and $X_S:=\{X_k:k\in S\}$. When $S=\varnothing$, the problem reduces to an unconditional independence test. 


Constraint-based methods such as PC and its variants repeatedly invoke such tests over a large number of variable pairs and conditioning sets, leading to a combinatorial explosion in the number of queries during the skeleton learning stage \citep{kalisch2007estimating,colombo2014order}. As the number of variables and the size of candidate conditioning sets increase, the computational cost of each individual test together with the cumulative number of tests becomes a major practical bottleneck for constraint-based causal discovery. Therefore, high-dimensional causal discovery calls for a CI testing procedure that is not only statistically effective in a single test, but also computationally scalable under the repeated queries required by large-scale graph search.

Existing methods for CI testing are already quite rich \citep{li2020nonparametric}. Kernel-based methods can capture general nonlinear dependence structures and have been widely used for CI testing \citep{fukumizu2007kernel,zhang2011kernel,Rajen_2020_aos,scetbon2022asymptotic,guan_2025_ekci,he_2025_onthehardness,miyazaki_2025_Spectralgcm}, while metric-based approaches characterize conditional dependence through carefully designed test statistics \citep{Wang_2015_cdc,runge2018conditional}. However, although these classical methods are valuable for handling complex nonlinear relationships, they often face limitations in sample size, conditioning-set dimension, or system level scalability: kernel-based methods are typically computationally expensive, especially in large sample settings, whereas metric-based methods may suffer substantial performance degradation in high-dimensional settings or with complex conditioning sets. In contrast, the conditional randomization test (CRT) offers a particularly appealing perspective by reducing CI testing to a conditional resampling problem \citep{candes2018panning}. Its core idea is to generate random samples from the conditional distribution $P_{X_i \mid X_S}$ and compare the test statistic computed from the resampled data with the original one \citep{li2023nearest,li2024k}. As a result, the practical viability of CRT largely depends on whether one can sample efficiently from $P_{X_i \mid X_S}$. Recent advances in generative modeling have opened up a new technical route for this problem \citep{goodfellow2014generative,ho2020denoising,song2021score}, leading to a growing line of generative CI testing methods \citep{bellot2019conditional,shi2021double,yang_2025_cdcit,ren_2025_sgmcit,he_2025_transportmap,yang_2026_dacit}.

However, once we move to the setting of constraint-based causal discovery, the key challenge is no longer whether a single fixed conditional distribution can be learned accurately, but whether the conditional sampling mechanism can be efficiently reused as the target variable and conditioning set vary across CI queries. Existing methods such as GCIT \cite{bellot2019conditional}, CDCIT \citep{yang_2025_cdcit} and SGMCIT \citep{ren_2025_sgmcit}, which are based on conditional generative models, have shown strong flexibility and competitive performance for a single CI test. But in the large-scale graph search carried out by PC and its variants, one must handle a massive number of different CI queries. If a separate conditional generative model has to be fitted for each $(i,S)$ pair, the overall computational cost grows rapidly with the number of queries, substantially limiting the practical utility of such methods for causal discovery \citep{yang_2025_cdcit,ren_2025_sgmcit}. Therefore, the core problem we address is not only how to accelerate each individual CI test, but also how to avoid repeated model training across the large number of CI queries arising in causal discovery, thereby achieving both per-test efficiency and cross-query scalability.

Motivated by the above observations, we propose a reusable conditional resampling framework for constraint-based causal discovery, with an emphasis on cross-query reuse. The key idea is to avoid learning a separate conditional distribution $P_{X_i \mid X_S}$ for each $(i,S)$ query. Instead, we learn the joint distribution of all variables only once, and reformulate conditional sampling from $P_{X_i \mid X_S}$ as a global imputation problem: given $X_S$, we impute the full variable vector and then extract the target coordinate $X_i$. 
Based on this reformulation, we develop a Langevin-RePaint algorithm for conditional imputation. Building on this algorithm, we propose the Flow Matching based Conditional Independence Test (FMCIT), which allows a single flow matching (FM) model to be reused across queries throughout the causal discovery pipeline, thereby avoiding repeated training of conditional generative models for different target variables and conditioning sets. 
This design shifts the focus of generative CI testing from constructing a conditional sampler for a single test to providing a reusable conditional resampling mechanism for large scale graph search. In addition, we aim to reduce the total number of queries required by constraint-based causal discovery. To this end, we develop a guided PC-style skeleton learning framework that prioritizes CI tests on more informative variable pairs and conditioning sets. This allows us to control both per-test cost and overall query complexity, yielding a more practical accuracy-efficiency trade-off in high-dimensional causal discovery.

\noindent\textbf{Our contributions are summarized as follows.} \textit{(1) Reusable conditional resampling framework.} We introduce a reusable conditional resampling framework for generative CRT-based causal discovery, replacing query-specific conditional generative modeling with a single joint model that can be shared across CI queries. \textit{(2) FMCIT and GPC-FMCIT.} We propose FMCIT, a flow matching based CI test that performs conditional resampling via mask-based joint imputation, and embed it into GPC-FMCIT, a screen--guide--refine--orient PC-style pipeline with budgeted nonlinear refinement. 
\textit{(3) Langevin-RePaint and the type-I error of FMCIT.} We  develop Langevin-RePaint for conditional imputation. Building on Langevin-RePaint, we further establish theoretical guarantees for the type-I error control of FMCIT.
\textit{(4) Empirical evaluation.} We evaluate the method on synthetic CI benchmarks, synthetic CPDAG recovery tasks, and hESC gene expression data. The results demonstrate improved CPDAG level recovery on nonlinear structural equation models, and on hESC data the method yields smaller graphs than PC/FisherZ in a descriptive comparison.

Section~\ref{ff_method} introduces Langevin-RePaint, FMCIT and the guided GPC-FMCIT causal graph search procedure. Section~\ref{sec_simulation} presents the simulation study, and Section~\ref{sec:real} studies real gene expression data. The Appendix provides background on CRT and PC, detailed algorithms and proofs, full experimental settings, standalone CI testing benchmarks, and supplementary real data analyses.

\section{Methods}  \label{ff_method}

\subsection{Flow matching based conditional independence test (FMCIT)}
\label{sec_fmcit}
Our proposed Flow Matching based Conditional Independence Test (FMCIT) is built on the CRT framework \citep{candes2018panning}, which is detailed in Appendix~\ref{sec_crt}. For a given CI query $(i,j,S)$, the key step of CRT is to generate random samples from the conditional distribution $P_{X_i \mid X_S}$ and compare the statistic computed from the original samples with that computed from the randomized samples. However, in constraint-based causal discovery, the target variable $X_i$ and the conditioning set $X_S$ vary across CI queries. If a separate conditional generative model is fitted for each $(i,S)$ pair, the overall computational cost grows rapidly with the number of queries, severely limiting the scalability of generative CI testing in graph search. Therefore, the goal of FMCIT is not merely to accelerate a single conditional sampling step, but to construct a conditional resampling mechanism that can be repeatedly invoked throughout the entire causal discovery pipeline.

The core idea of FMCIT is to avoid learning a separate conditional distribution $P_{X_i \mid X_S}$ for each $(i,S)$. Instead of directly modeling a query-dependent conditional generator, we learn a joint generative model for $P_X$ once and reformulate sampling from $P_{X_i \mid X_S}$ as a joint imputation problem: given $X_S$, we impute the full variable vector $X$ and then extract the $i$-th coordinate. In this way, the same generative model can be shared across different $(i,j,S)$ queries, turning generative CI testing from query-specific conditional sampling into query-reusable imputation. For constraint-based causal discovery, this reformulation fundamentally alleviates the repeated retraining issue caused by changing conditioning sets.

\subsubsection{Introduction to flow matching (FM)}
To realize the reusable imputation mechanism described above, we adopt flow matching \citep[FM,][]{lipman_2023_fm} as the implementation of the joint generative model. The motivation for using FM is twofold. First, its training procedure is relatively fast and stable. Second, its generation process is based on deterministic ordinary differential equations (ODE) and is therefore much faster than the stochastic counterpart of diffusion models \cite{liu_2023_recflow}, making it better suited for repeated use in a large number of CRT resampling tasks. In our setting, the role of FM is not merely to replace one generative model with another, but to provide an effective foundation for a train once, reuse-across-queries conditional resampling framework that balances high generation quality and inference efficiency.

We now describe the construction of the joint FM model. Let $X{(1)}=X$ denote the target variable, with $X{(0)}\sim N(0,I_p)$ representing the initial noise. FM constructs a continuous-time path between $X{(0)}$ and $X{(1)}$, and learns a corresponding vector field $\widehat v_t$ that transports the noise distribution to the joint data distribution. Specifically, we use the linear interpolation path 
\begin{equation} \label{eq_fm_ode}
    \frac{dX(t)}{dt}=X{(1)}-X{(0)}, \quad X(0) \perp \! \! \! \perp X(1), \quad t\in[0,1],
\end{equation}
which has the closed-form representation 
    $X(t)=tX{(1)}+(1-t)X{(0)}$. 
Since $X{(1)}$ is unknown at generation time, we use a neural network $\widehat v_t$ to approximate the conditional vector field along this path, and train it by minimizing the flow matching loss
\begin{equation}
    \mathbb{E}_{X(0) \sim N(0,I_p),\, X(1) \sim P_X,\, t \sim U[0,1)} \left\| X(1) - X(0) - \widehat v_t \left(t X(1) + (1-t) X(0)\right)  \right\|_2^2.
\end{equation}
After training, we can generate samples approximately following the joint distribution $P_X$ by solving the ODE
\begin{equation}
    \frac{d X(t)}{dt} = \widehat v_t(X(t) ),  \quad X(0) \sim N(0, I_p), \quad t \in [0,1).
\end{equation}
In other words, this step yields a generator that models the full joint structure of all variables, rather than a local conditional model tailored to a fixed query.

\subsubsection{Conditional imputation via Langevin-RePaint}
Given the joint FM model, we further transform the query-specific conditional sampling task into a global imputation problem. For any CI query $(i,j,S)$, we treat the coordinates indexed by the conditioning set $S$ as observed and enforce the observed coordinates at each imputation step; the remaining coordinates are treated as missing and updated by the FM generation process. As a result, sampling from $P_{X_i\mid X_S}$ is reformulated as follows: first generate an imputed sample approximately from $P_{X\mid X_S}$, and then extract the $i$-th coordinate as the randomized copy required by CRT. The key advantage of this reformulation is that, regardless of how the $i$ and conditioning set $S$ vary, we always invoke the same joint generative model and only change the mask specifying which coordinates are observed and fixed.

To enforce the conditioning information $X_S$ throughout generation, we develop Langevin-RePaint, inspired by the vanilla RePaint \citep{Lugmayr_2022_repaint,hu_2025_FlowTS}. We consider Langevin-RePaint on the uniform grid $t_\ell = \ell/L, \ell = 0,\ldots,L$.
Starting from the initial noise
$\widetilde X_{t_0}\sim N(0,I_p)$, we construct an estimate
$\widehat X_{t_\ell}$ of the terminal state
$X(1)$ at each sampling step $\ell=0,\ldots,L-1$.
Specifically, we use the Picard-based terminal
prediction in the early stage and introduce a
half-score correction in the terminal stage:
\begin{equation}
\label{eq_lr_update}
\begin{aligned}
\widehat X_{t_\ell}
&=
\begin{cases}
\widetilde X_{t_\ell}+(1-t_\ell)
\widehat v_{t_\ell}(\widetilde X_{t_\ell}),
& t_\ell<1-L^{-1/4}, \\[2mm]
\frac{1+t_\ell}{2 t_\ell}\widetilde X_{t_\ell}
+\frac{1-t_\ell}{2}
\widehat v_{t_\ell}(\widetilde X_{t_\ell}),
& t_\ell\geq 1-L^{-1/4},
\end{cases}, \quad \ell=0,\ldots,L-1 \\
\widetilde X_{t_{\ell+1}}
&=
t_{\ell+1} \mathbb{P}^S_{X_S}   \big( \widehat{X}_{t_\ell} \big)
+(1-t_{\ell+1})\varepsilon_{\ell+1}, \quad \varepsilon_{\ell+1} \sim N(0,I_p), \quad \ell=0,\ldots,L-2,
\end{aligned}
\end{equation}
where $\mathbb{P}^S_{X_S} (\cdot)$ denotes the projection operator that replaces the coordinates indexed by $S$ with the observed values $X_S$. The final output of Langevin-RePaint is $ \mathbb{P}^S_{X_S} \big( \widehat X_{t_{L-1}} \big)$. 

For $t_\ell<1-L^{-1/4}$, the update follows the vanilla RePaint-style sampling procedure \cite{Lugmayr_2022_repaint}. However, as shown in Appendix \ref{sec_repaint_fails}, without correction, the resulting imputation distribution converges to a biased distribution proportional to $p_{X_{S^c}\mid X_S}^2$, where $p_{X_{S^c}\mid X_S}$ is the density of $P_{X_{S^c}\mid X_S}$. To avoid this, the half-score correction for $t_\ell\geq 1-L^{-1/4}$ in (\ref{eq_lr_update}) restores convergence to the target conditional distribution under suitable regularity conditions. The following theorem establishes the convergence of Langevin-RePaint in total variation (TV) distance.

\begin{theorem}
\label{thm_LR_tv}
Let $\widehat v_t$ denote the velocity field trained on dataset $\mathcal D$ with $N$ samples, and $v_t^*$ the true velocity field. We define the estimation error of FM velocity as 
\begin{equation}
\label{eq_thm_tv}
    \mathcal E_{N}^{\mathrm{FM}} := \mathbb E_{\mathcal D} \mathbb E_t \mathbb E_{X(t)}  \left\| \widehat v_{t}(X(t)) -  v^*_{t}(X(t)) \right\|^2
\end{equation}
Let $\widehat P_{X_{S^c}\mid X_S}$ and $ P_{X_{S^c}\mid X_S}$ denote
the conditional distribution generated by
Langevin-RePaint and the true conditional distribution.
Under Assumptions \ref{ass_lr_joint_reg}-\ref{ass_lr_time_transfer},
there exist constants $c,C>0$ such that
\begin{equation}
\label{eq_lr_tv}
\begin{aligned}
&\mathbb E_{\mathcal D}\mathbb E_{X_S}
\left[
d_{\mathrm{TV}}\left(
\widehat P_{X_{S^c}\mid X_S},
P_{X_{S^c}\mid X_S}
\right)
\right] \leq C\left[
\sqrt{\log L}\,e^{-cL^{1/4}}
+L^{-1/4}
+\sqrt{L^{3/4} \mathcal E_{N}^{\mathrm{FM}} }
\right].
\end{aligned}
\end{equation}

\end{theorem}

The proposed Langevin-RePaint injects randomness through Gaussian perturbations while preserving the conditioning information via projection, thereby enabling more accurate and diverse conditional imputations. More importantly, Langevin-RePaint avoids repeated repainting \citep{Lugmayr_2022_repaint} and the computational overhead associated with query-specific conditional generative sampling \citep{yang_2025_cdcit}, allowing efficient conditional imputation with relatively few sampling steps, typically 15–50.

Based on the above conditional imputation mechanism, FMCIT follows the standard CRT pipeline to construct the test. For a given query $(i,j,S)$, we fix the observed conditioning variables $X_S$ and generate $B$ imputed samples in parallel. We then extract the target coordinates $\mathbf{X_i}^{(1)},\ldots,\mathbf{X_i}^{(B)}$ from these imputations as randomized copies. Let $T(\cdot)$ be a given statistic. We compute the original statistic $T\big(\mathbf{X_i},\mathbf{X_j},\mathbf{X_S}\big)$, as well as the randomized statistics $T\big(\mathbf{X_i}^{(b)},\mathbf{X_j},\mathbf{X_S}\big), b=1,\ldots,B$. The $p$-value in FMCIT is then defined as,
\begin{align}
    p_{\mathrm{FMCIT}} :=
\frac{
1+\sum_{b=1}^B \mathbf{1}\!\left\{
T(\mathbf{X_i}^{(b)},\mathbf{X_j},\mathbf{X_S})\ge T(\mathbf{X_i},\mathbf{X_j},\mathbf{X_S})
\right\}
}{
1+B
}.
\end{align}
In our implementation, the $B$ randomized copies are generated in parallel in a batched manner, which substantially reduces the conditional resampling cost of each individual test. At the same time, because all queries share the same joint FM model, FMCIT requires training the generator only once throughout the entire causal discovery pipeline, making it particularly well suited for constraint-based graph search with a large number of CI queries. Building on Theorem~\ref{thm_LR_tv}, the following theorem establishes an upper bound on the type-I error of FMCIT.

\begin{theorem}[Type-I Error of FMCIT \cite{berrett2020conditional}]
\label{thm_FMCIT_typeI}
Let $\alpha\in(0,1)$ be the significance level,
$p_{\mathrm{FMCIT}}$ denote the p-value of FMCIT,
and $n$ be the sample size used for the CRT.
Assume that the training dataset $\mathcal D$ is
independent of the CRT samples.
Under the null hypothesis, the type-I error
of FMCIT satisfies
\begin{equation}
\label{eq_FMCIT_typeI}
\begin{aligned}
\mathbb P(p_{\mathrm{FMCIT}}\leq\alpha)
\leq \alpha
+n\mathbb E_{\mathcal D}\mathbb E_{X_S}
\left[
d_{\mathrm{TV}}\left(
\widehat P_{X_{S^c}\mid X_S},
P_{X_{S^c}\mid X_S}
\right)
\right].
\end{aligned}
\end{equation}
\end{theorem}
Suppose $\mathcal D$ contains
$N$ samples and the learning error satisfies
$\mathcal E_{N}^{\mathrm{FM}}=O(N^{-\kappa})$
for some $\kappa>0$. When $N\gg n^{5/\kappa}$, by choosing $L\asymp N^{4\kappa/5}$, the type-I error of FMCIT is asymptotically
controlled at the nominal significance level:
$\mathbb P(p_{\mathrm{FMCIT}}\leq\alpha)
\leq\alpha+o(1).$

\begin{figure}[htbp]
    \centering
    \includegraphics[width=1.0\linewidth]{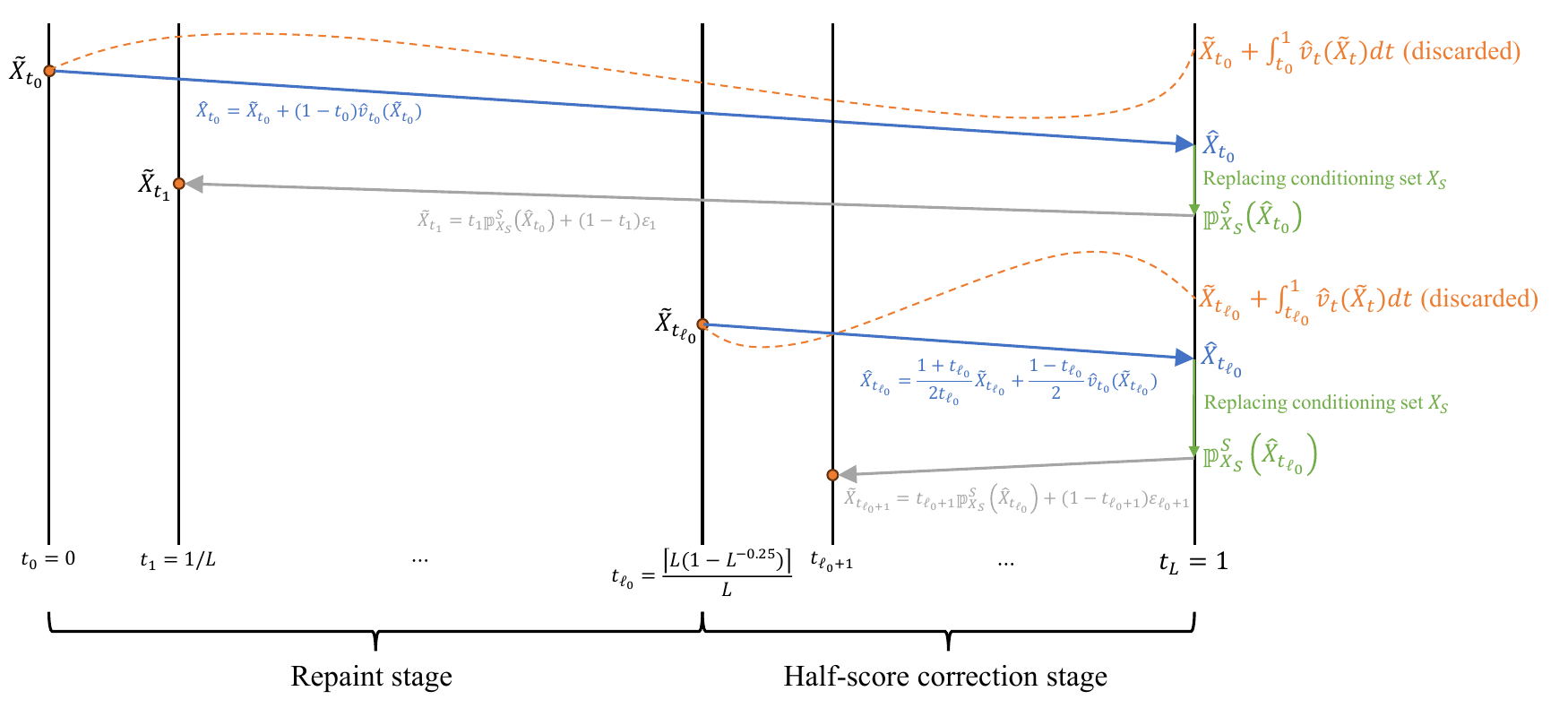}
    \caption{An illustration of Langevin-RePaint. }
    \label{fig_picard_and_repaint}
\end{figure}

In summary, by combining joint modeling, conditional imputation, and CRT, FMCIT transforms generative CI testing from query-specific conditional modeling into a conditional resampling mechanism that is reusable across queries. This design retains the flexibility of CRT in complex high-dimensional settings while substantially reducing the overall deployment cost of generative CI testing in causal discovery. The FM training procedure, the algorithm of Langevin-RePaint and FMCIT are provided in Algorithms~\ref{alg_train_fm}, \ref{alg_sample_fm}, and \ref{alg_fmcit} in Appendix~\ref{all_algorithms}. Appendix~\ref{app:fmcit_theory} provides the proofs of Theorem \ref{thm_LR_tv} and \ref{thm_FMCIT_typeI}.
Figure~\ref{fig_picard_and_repaint} illustrates the Langevin-RePaint conditional imputation mechanism used by FMCIT.

\subsection{GPC-FMCIT: guided CPDAG learning with FMCIT}
\label{sec_fmcit_pc}

The previous section introduced FMCIT as a reusable approximate nonlinear CI oracle. 
Its key advantage is that a single joint FM model can be trained once and then reused across different target variables and conditioning sets through masked conditional imputation. 
Because FMCIT relies on approximate learned conditional resampling, we treat it as an empirical nonlinear CI oracle whose calibration is evaluated in the standalone CI experiments.
However, directly using FMCIT for all CI queries in constraint-based graph learning is still computationally unattractive. The bottleneck comes not only from the cost of one CI query, but also from the large number of variable pairs and conditioning sets that must be considered during graph search. Therefore, to make FMCIT practical at the graph-learning level, we need a mechanism that controls where FMCIT is invoked and which conditioning sets are tested.

We propose \textbf{GPC-FMCIT}, a modular framework for guided CPDAG learning. By ``guided CPDAG learning'', we mean the following four-stage procedure: first construct an undirected candidate graph, then use this graph to guide the search over conditioning variables, then refine the candidate graph by CI-based edge deletion, and finally orient the refined skeleton to obtain a CPDAG. Thus, GPC-FMCIT follows a \emph{screen--guide--refine--orient} design. The screening module constructs a candidate graph \(G_{\mathrm{scr}}\); the guidance module converts \(G_{\mathrm{scr}}\) into edge-specific conditioning pools; the refinement module invokes FMCIT on a budgeted subset of candidate conditioning sets to remove edges; and the orientation module applies conservative collider orientation followed by Meek propagation rules. This formulation keeps the standard constraint-based principle of deleting edges by conditional independence evidence, but replaces exhaustive search with a guided and budgeted refinement process.


\textbf{Screening module. }
The screening module constructs an undirected candidate graph \(G_{\mathrm{scr}}=([p],E_{\mathrm{scr}})\), which defines the initial edge set for downstream FMCIT refinement. Given a screening CI test \(\mathcal T_{\mathrm{scr}}\), a significance level \(\alpha_{\mathrm{scr}}\), and a maximum screening depth \(d_{\mathrm{scr}}\), the screening procedure starts from the complete undirected graph and removes edges by testing conditional independence relations with increasing conditioning-set size. At level \(\ell=0,1,\ldots,d_{\mathrm{scr}}\), for each current edge \((i,j)\), it tests whether \(X_i\perp\!\!\!\perp X_j\mid X_S\) for candidate sets \(S\) of size \(\ell\) drawn from the current neighborhoods of \(i\) or \(j\). If the screening test accepts conditional independence for some \(S\), the edge \(i-j\) is removed. The remaining graph after this preliminary deletion stage is used as \(G_{\mathrm{scr}}\). Importantly, \(G_{\mathrm{scr}}\) is not the final learned skeleton; it is only a candidate graph that restricts the search space for the more expensive FMCIT-based refinement stage.

The screening module is plug-in: FisherZ screening gives a scalable low-cost candidate graph, while KCI screening gives a stronger nonlinear candidate graph at higher cost. 
In the KCI-screened version, KCI is used only to construct the candidate screening graph \(G_{\mathrm{scr}}\); the subsequent guidance, FMCIT refinement, and conservative orientation modules remain the same as in the FisherZ-screened version.

\textbf{Guidance module. }
Given \(G_{\mathrm{scr}}\), the guidance module converts the screening graph into edge-specific conditioning pools. 
For each current edge \((i,j)\), we define the raw local candidate set as 
\(\mathcal Q_{ij}=(\mathrm{Adj}_{G_{\mathrm{scr}}}(i)\cup \mathrm{Adj}_{G_{\mathrm{scr}}}(j))\setminus\{i,j\}\), 
where \(\mathrm{Adj}_{G_{\mathrm{scr}}}(i)\) denotes the neighbors of \(i\) in the screening graph. 
This restricts potential separating variables to the local neighborhood of the two endpoints. 
The guidance rule then ranks variables in \(\mathcal Q_{ij}\) and truncates the list to an ordered pool \(\mathcal P_{ij}\) with \(|\mathcal P_{ij}|\le K\).

We consider two guidance rules. 
\emph{Stability guidance} reruns the screening procedure over a small grid of screening levels \(\mathcal A=\{\alpha_1,\ldots,\alpha_R\}\). 
For each pair \((u,v)\), it computes the edge-selection frequency 
\(\mathrm{freq}_{uv}=R^{-1}\sum_{r=1}^R\mathbf 1\{(u,v)\in E(G_{\alpha_r})\}\). 
These frequencies are used to construct a coarse support graph and a global priority order, so that variables with stable screening support are tested earlier. 
\emph{Triangle guidance} uses only the topology of \(G_{\mathrm{scr}}\). 
For each edge \((u,v)\), it computes the triangle support 
\(\Delta_{uv}=|\mathrm{Adj}_{G_{\mathrm{scr}}}(u)\cap\mathrm{Adj}_{G_{\mathrm{scr}}}(v)|\), 
and uses the resulting triangle-supported structure to prioritize variables that are locally informative for separating adjacent regions of the screening graph.

Both rules output an ordered pool \(\mathcal P_{ij}\) for each edge. 
They do not change the CI oracle or make edge-deletion decisions; they only determine which conditioning variables are considered first. 
The actual refinement decisions are made by FMCIT in the next stage.

\textbf{FMCIT-based refinement module. }
Starting from \(G^{(0)}\leftarrow G_{\mathrm{scr}}\), GPC-FMCIT performs CI-based edge deletion using FMCIT as the nonlinear CI oracle.
At conditioning level \(\ell\), for each current edge \((i,j)\), the algorithm constructs candidate separating sets \(S\subseteq\mathcal P_{ij}\) with \(|S|=\ell\), and tests \(X_i\perp\!\!\!\perp X_j\mid X_S\) using FMCIT. 
If the null is accepted, the edge \(i-j\) is removed and the separating set is recorded as \(\mathrm{Sep}(i,j)\leftarrow S\) and \(\mathrm{Sep}(j,i)\leftarrow S\). 
This preserves the edge-deletion logic of PC, but replaces exhaustive conditioning-set enumeration by guided and budgeted search.

More precisely, for each edge and level, we test at most \(M\) candidate conditioning sets from the guided pool. 
If a separating set is found, the search for that edge stops immediately at the current level. 
The refinement depth can either be fixed in advance or chosen adaptively. 
In the adaptive mode, the algorithm increases \(\ell\) until no current edge admits any feasible conditioning set of size \(\ell\) from its guided pool. 
Let \(E^{(\ell)}\) denote the current edge set at level \(\ell\), and let \(d_{\mathrm{eff}}\) be the largest level reached. 
Since refinement only removes edges, \(E^{(\ell)}\subseteq E_{\mathrm{scr}}\). 
Because each edge is tested on at most \(M\) sets per level, the number of FMCIT queries satisfies \(N_{\mathrm{ref}}\le M\sum_{\ell=0}^{d_{\mathrm{eff}}}|E^{(\ell)}|\le M(d_{\mathrm{eff}}+1)|E_{\mathrm{scr}}|\). 
This bound shows how the nonlinear query complexity is controlled by screening sparsity, guided candidate pools, and the per-edge search budget.

All CI queries in this stage reuse the same joint FM model trained for FMCIT. 
For a query \((i,j,S)\), FMCIT only changes the mask specifying which coordinates are fixed as \(X_S\), performs conditional imputation using the shared model, and extracts the randomized copy of \(X_i\) required by the CRT. 
Thus, GPC-FMCIT avoids training a separate conditional generator for each target variable and conditioning set.

\textbf{Conservative CPDAG orientation. }
After skeleton refinement, GPC-FMCIT outputs a CPDAG rather than only an undirected skeleton. 
For each unshielded triple \(i-k-j\), we perform a conservative collider search: if the separating-set evidence consistently supports \(k\notin S\), the triple is oriented as \(i\to k\leftarrow j\); if the evidence is ambiguous, the triple is left unoriented. 
We then apply Meek propagation rules to orient as many remaining edges as possible without introducing directed cycles or invalid new v-structures. Since finite-sample CI decisions can yield an acyclic graph that is not a valid completed PDAG, all synthetic metrics are computed after a deterministic, ground-truth-independent completion step (Appendix~\ref{app:cpdag_validity}).

This final stage is important because strong skeleton recovery alone does not guarantee strong CPDAG recovery. 
In our framework, the screening module controls which candidate edges enter the search, the guidance and refinement modules decide which edges survive nonlinear CI testing, and the conservative orientation module determines which surviving adjacencies can be reliably directed. 
The default FisherZ-screened version of GPC-FMCIT emphasizes scalability, whereas the KCI-screened version is an accuracy-oriented strong-screen instantiation. 
Both versions share the same downstream FMCIT-based refinement and orientation pipeline. The pseudocode of the full framework is provided in Algorithm~\ref{alg:gpc_fmcit} in Appendix~\ref{all_algorithms}.

\section{Simulation study}
\label{sec_simulation}

\subsection{Empirical null calibration of FMCIT}
\label{sec:null_calibration}

Before evaluating graph-level causal discovery, we first examine the null calibration of FMCIT in a post-nonlinear CI benchmark (\ref{eq_ci_benchmark}).
Because FMCIT uses approximate learned conditional resampling, this experiment is intended as an empirical calibration check rather than a proof of exact finite-sample type-I error control.
We fix the conditioning dimension to \(d_z=50\), the noise scale to \(s=0.75\), and the nominal level to \(\alpha=0.05\), and vary the sample size \(n\in\{250,500,1000\}\).
As shown in Table~\ref{tab:type1_n_calibration_main}, the empirical type-I error is mildly above the nominal level for \(n=250\) and \(n=500\), and is close to the nominal level at \(n=1000\).
With \(100\) repetitions, finite-sample fluctuations should not be over-interpreted; this is an empirical calibration check, not an exact finite-sample validity guarantee.

\begin{table}[t]
\centering
\caption{
Null calibration of FMCIT in the post-nonlinear CI benchmark.
We fix \(d_z=50\), \(s=0.75\), and nominal level \(\alpha=0.05\), and report empirical type-I error over \(100\) independent null trials.
The \(\pm\) value is the binomial Monte Carlo standard error of the empirical rejection rate.
Runtime is measured per CI test.
}
\label{tab:type1_n_calibration_main}
\small
\setlength{\tabcolsep}{5.0pt}
\begin{adjustbox}{max width=\linewidth}
\begin{tabular}{lcccc}
\toprule
\(n\)
& Type-I error \(\downarrow\)
& Mean \(p\)-value
& Median \(p\)-value
& Runtime(s) \(\downarrow\) \\
\midrule
250
& \(0.09 \pm 0.029\)
& \(0.489\)
& \(0.465\)
& \(1.04 \pm 0.31\) \\
500
& \(0.09 \pm 0.029\)
& \(0.441\)
& \(0.425\)
& \(1.10 \pm 0.22\) \\
1000
& \(0.04 \pm 0.020\)
& \(0.518\)
& \(0.545\)
& \(1.30 \pm 0.25\) \\
\bottomrule
\end{tabular}
\end{adjustbox}
\end{table}

\subsection{Graph-level causal discovery}
\label{sec:graph_level_simulation}

We next evaluate the proposed framework on synthetic causal discovery tasks where the ground-truth graph is known.
Our goal is to assess whether reusable FMCIT, when embedded into the proposed screen--guide--refine--orient framework, improves full CPDAG recovery under nonlinear and heavy-tailed data-generating mechanisms.

\textbf{Data generation. }
We generate synthetic observational data from random additive nonlinear structural equation models (SEMs) with heavy-tailed noise. For each repetition, we first sample a random DAG with $p$ nodes under a fixed topological ordering. Given the DAG, the variables are then generated sequentially following this ordering: $X_j=\sum_{i\in\mathrm{Pa}(j)}w_{ij}f_{ij}(X_i)+\varepsilon_j$, where $\mathrm{Pa}(j)$ denotes the set of parents of node $j$, $w_{ij}$ is a signed edge weight, $f_{ij}$ is an edge-specific nonlinear transformation, and $\varepsilon_j$ is heavy-tailed noise. This data-generating process yields nonlinear and non-Gaussian observational data with a known ground-truth causal structure. Since PC-style methods aim to recover the Markov equivalence class, we compare the graph learned by each method against the CPDAG induced by the ground-truth DAG, rather than evaluating only the skeleton. Full details of the data-generating process and all numerical settings are provided in Appendix~\ref{app:exp_pc}.





\textbf{Compared methods. }
We compare two operating regimes of GPC-FMCIT against their corresponding PC-style baselines. 
In the FisherZ family, GPC-FMCIT (FisherZ screen) improves over PC-stable (FisherZ) in all four accuracy metrics at both levels, at the cost of additional FMCIT-based refinement.
This shows that even with a low-cost linear screening graph, the reusable nonlinear refinement stage can improve CPDAG-level recovery.

In the KCI family, PC-stable (KCI) serves as a strong nonlinear CI baseline, while GPC-FMCIT (KCI screen) uses KCI only as a nonlinear screening module and then applies the same FMCIT-based refinement and orientation pipeline. 
This grouping separates the scalable FisherZ-screened regime from the accuracy-oriented KCI-screened regime. 
A direct PC-stable(FMCIT) baseline and additional component ablations are reported in Table~\ref{tab:component_ablation}.
Implementation details and hyperparameter settings for all methods are provided in Appendix~\ref{app:exp_pc}.

\textbf{Evaluation metrics. }
We evaluate causal graph recovery at three levels. 
At the skeleton level, we report skeleton F1. 
At the orientation level, we report strict directed F1, where a directed edge is counted as correct only if its orientation exactly matches the ground-truth CPDAG, and relaxed directed F1, which gives partial credit when a true directed edge is recovered as undirected. 
At the full-graph level, we report CPDAG pairwise error, which counts the number of unordered node pairs for which the learned CPDAG state differs from the ground-truth CPDAG state, including missing edges, extra edges, and orientation errors. 
Runtime is measured as wall-clock time in seconds.
Detailed definitions of the evaluation metrics are provided in Appendix~\ref{app:exp_pc}.

\textbf{Results. }
The main CPDAG recovery results are shown in Table~\ref{tab:sim_pc_main}. 
The table highlights two operating regimes of the proposed framework. 
In the FisherZ family, PC-stable (FisherZ) serves as a scalable linear-Gaussian baseline, while GPC-FMCIT (FisherZ screen) uses FisherZ only to construct a low-cost candidate graph and then applies FMCIT-based nonlinear refinement and conservative orientation.

In the KCI family, both methods attain similar skeleton F1 (PC-stable (KCI) is higher at \(\alpha=0.05\), GPC-FMCIT at \(\alpha=0.10\)), confirming that KCI is effective for recovering adjacencies.
However, GPC-FMCIT (KCI screen) achieves clearly higher directed and relaxed directed F1 and the lowest CPDAG error at both levels; the directed-F1 and CPDAG-error gains already hold before the final completion step (Appendix~\ref{app:cpdag_validity}).
This indicates that strong skeleton recovery alone does not necessarily imply strong CPDAG recovery: after obtaining a nonlinear screening graph, FMCIT-based refinement and conservative orientation improve the balance between adjacency recovery and orientation quality. 
 
The KCI-screened variant should therefore be viewed as an accuracy-oriented instantiation of the same screen--guide--refine--orient framework, whereas the FisherZ-screened version uses a simpler screening oracle.

Overall, the two instantiations highlight different operating points of the same framework.
The FisherZ-screened version uses a less computationally demanding screening test and, relative to direct FMCIT-based PC search, uses far fewer FMCIT queries (Table~\ref{tab:component_ablation}), whereas the KCI-screened version emphasizes CPDAG accuracy by starting from a stronger nonlinear candidate graph.
Additional component and hyperparameter ablations are reported below to examine the effects of screening, nonlinear refinement, candidate ordering, and budgeted search.

\begin{table*}[t]
\centering
\caption{
Main CPDAG recovery results on additive nonlinear SEMs
($n=500$, $p=100$, $\texttt{exp\_edges}=150$,
$\mathrm{df}=3$).
Mean $\pm$ sample standard deviation over ten matched seeds.
For GPC-FMCIT,
$\alpha_{\mathrm{scr}}=\alpha_{\mathrm{fm}}=\alpha$.
Metrics are computed on valid completed PDAGs (PC-stable after repair) against the ground-truth CPDAG (Appendix~\ref{app:cpdag_validity}).
Bold: best accuracy within each family.
}
\label{tab:sim_pc_main}
\small
\setlength{\tabcolsep}{3.5pt}
\begin{adjustbox}{max width=\textwidth}
\begin{tabular}{llccccc}
\toprule
$\alpha$
& Method
& Skel-F1 $\uparrow$
& Dir-F1 $\uparrow$
& Relaxed Dir-F1 $\uparrow$
& CPDAG Error $\downarrow$
& Recorded Runtime (s) \\
\midrule

\multirow{4}{*}{$0.05$}
& PC-stable (FisherZ)
& $0.558 \pm 0.041$
& $0.073 \pm 0.026$
& $0.095 \pm 0.026$
& $231.9 \pm 9.9$
& $0.64 \pm 0.07$ \\

& GPC-FMCIT (FisherZ screen)
& $\mathbf{0.685 \pm 0.049}$
& $\mathbf{0.286 \pm 0.057}$
& $\mathbf{0.301 \pm 0.055}$
& $\mathbf{136.4 \pm 12.0}$
& $517.7 \pm 220.4$ \\

\addlinespace[2pt]

& PC-stable (KCI)
& $\mathbf{0.876 \pm 0.031}$
& $0.240 \pm 0.066$
& $0.287 \pm 0.048$
& $126.0 \pm 9.5$
& $1435.1 \pm 261.8$ \\

& GPC-FMCIT (KCI screen)
& $0.861 \pm 0.030$
& $\mathbf{0.379 \pm 0.038}$
& $\mathbf{0.387 \pm 0.038}$
& $\mathbf{104.2 \pm 5.0}$
& $1985.8 \pm 440.3$ \\

\midrule

\multirow{4}{*}{$0.10$}
& PC-stable (FisherZ)
& $0.499 \pm 0.028$
& $0.054 \pm 0.016$
& $0.068 \pm 0.017$
& $285.6 \pm 9.4$
& $1.28 \pm 0.25$ \\

& GPC-FMCIT (FisherZ screen)
& $\mathbf{0.682 \pm 0.044}$
& $\mathbf{0.286 \pm 0.042}$
& $\mathbf{0.294 \pm 0.042}$
& $\mathbf{147.4 \pm 10.4}$
& $1128.1 \pm 531.2$ \\

\addlinespace[2pt]

& PC-stable (KCI)
& $0.848 \pm 0.033$
& $0.248 \pm 0.045$
& $0.277 \pm 0.043$
& $137.2 \pm 8.8$
& $2522.4 \pm 316.0$ \\

& GPC-FMCIT (KCI screen)
& $\mathbf{0.883 \pm 0.031}$
& $\mathbf{0.405 \pm 0.019}$
& $\mathbf{0.409 \pm 0.020}$
& $\mathbf{102.6 \pm 3.7}$
& $3308.5 \pm 579.4$ \\

\bottomrule
\end{tabular}
\end{adjustbox}
\end{table*}

\begin{table*}[t]
\centering
\caption{
Component ablation of GPC-FMCIT on additive nonlinear SEMs
(\(n=500\), \(p=50\), \(\texttt{exp\_edges}=75\),
\(\mathrm{df}=3\), \(\alpha_{\mathrm{scr}}=\alpha_{\mathrm{fm}}=0.10\)).
Mean \(\pm\) sample standard deviation over five matched seeds.
Bold: best mean in each column.
}
\label{tab:component_ablation}
\small
\setlength{\tabcolsep}{3.5pt}
\begin{adjustbox}{max width=\textwidth}
\begin{tabular}{lccccccccc}
\toprule
Variant
& Screen
& Candidate order
& Budget
& Refine CI
& Skel-F1 $\uparrow$
& Dir-F1 $\uparrow$
& CPDAG Error $\downarrow$
& \#FMCIT queries $\downarrow$
& Runtime(s) $\downarrow$ \\
\midrule
PC-FMCIT direct & None & Exhaustive & None & FMCIT & $0.663 \pm 0.033$ & $0.345 \pm 0.100$ & $104.6 \pm 15.0$ & $149952.0 \pm 96756.3$ & $5972.72 \pm 3947.85$ \\
Screen only & FisherZ & None & None & None & $0.585 \pm 0.047$ & $0.078 \pm 0.035$ & $117.0 \pm 9.2$ & $\mathbf{0.0 \pm 0.0}$ & $\mathbf{0.33 \pm 0.06}$ \\
Screen + exhaustive local FMCIT & FisherZ & Exhaustive local & None & FMCIT & $0.704 \pm 0.047$ & $0.343 \pm 0.058$ & $\mathbf{65.4 \pm 3.2}$ & $16245.0 \pm 7299.8$ & $632.31 \pm 290.63$ \\
Screen + random pool & FisherZ & Random & \(M=2\) & FMCIT & $\mathbf{0.716 \pm 0.057}$ & $0.332 \pm 0.043$ & $67.8 \pm 3.3$ & $3822.0 \pm 1096.7$ & $149.35 \pm 44.00$ \\
Screen + degree order & FisherZ & Degree & \(M=2\) & FMCIT & $0.707 \pm 0.053$ & $0.330 \pm 0.046$ & $69.0 \pm 2.9$ & $4087.2 \pm 1491.3$ & $159.27 \pm 56.59$ \\
Screen + triangle guide & FisherZ & Triangle & \(M=2\) & FMCIT & $0.706 \pm 0.054$ & $0.325 \pm 0.053$ & $69.6 \pm 4.0$ & $4147.4 \pm 1462.0$ & $159.45 \pm 59.51$ \\
Screen + guided pool, no budget & FisherZ & Stability & None & FMCIT & $0.715 \pm 0.053$ & $\mathbf{0.350 \pm 0.042}$ & $66.2 \pm 2.4$ & $4353.0 \pm 1508.2$ & $174.93 \pm 60.66$ \\
Full GPC-FMCIT & FisherZ & Stability & \(M=2\) & FMCIT & $0.712 \pm 0.049$ & $0.345 \pm 0.042$ & $67.6 \pm 2.9$ & $4172.6 \pm 1289.6$ & $164.33 \pm 50.95$ \\
\bottomrule
\end{tabular}
\end{adjustbox}
\end{table*}

Table~\ref{tab:component_ablation} shows that most of the gain comes from combining screening with FMCIT-based refinement: screening alone attains directed F1 \(0.078\) and CPDAG error \(117.0\).
Compared with direct PC-FMCIT, full GPC-FMCIT uses \(97.2\%\) fewer FMCIT queries and \(164.3\)\,s instead of \(5972.7\)\,s, with higher skeleton F1, lower CPDAG error and the same strict directed F1.
The full method is not best on every metric, and the candidate-ordering rules perform similarly; since most queries arise in orientation, the budget reduces refinement queries by \(46.6\%\) but total queries by only \(4.1\%\).
Appendix~\ref{app:budget_pool_sensitivity} shows that accuracy changes little with \(M\) and \(K\), so we use \(M=2,K=4\) as a moderate default.

\paragraph{Additional synthetic analyses.}
Appendix~\ref{app:scaling_sensitivity} reports sample-size sensitivity: for KCI-screened GPC-FMCIT, skeleton F1 increases with \(n\), whereas directed F1 is highest at \(n=500\).
Appendix~\ref{app:dgp_robustness} further evaluates post-nonlinear and interaction SEMs, where KCI screening gives higher mean accuracy than FisherZ screening within GPC-FMCIT; matched PC-stable baselines for these SEMs are not included.

\section{Real data analysis}
\label{sec:real}

The previous section evaluated graph-level causal discovery on synthetic data.
We next evaluate GPC-FMCIT on the processed hESC single-cell gene-expression benchmark from BEELINE \citep{pratapa2020beeline}.
The processed matrix contains 758 cells and 17,735 genes before additional preprocessing.
Detailed preprocessing and gene-selection protocols are given in Appendix~\ref{app:hesc_sensitivity}.

Since no exact causal graph is available, we report graph compactness and computational cost rather than causal recovery accuracy.
We compare PC/FisherZ with stability-guided and triangle-guided GPC-FMCIT on the same 758-cell, 100-gene expression matrix, using full-joint FMCIT in the B39 configuration (39 CRT replicates, 15 sampling steps) and unbounded conditioning depth (\(\texttt{mcs}=-1\)) for all methods.

\textbf{CPDAG comparison. }

Table~\ref{tab:hesc_cpdag_main} summarizes the results.
PC/FisherZ uses \(\alpha=0.02\), whereas GPC-FMCIT screens at \(\alpha_{\mathrm{scr}}=0.10\) and refines at \(\alpha_{\mathrm{fm}}=0.05\), so the comparison is descriptive.
Both guided variants yield much smaller graphs than PC/FisherZ (136 and 130 versus 332 skeleton edges), and triangle guidance takes 4488.56\,s versus 7256.98\,s for stability guidance; a smaller graph does not by itself imply better biological recovery.
Appendix~\ref{app:hesc_sensitivity} shows the directed edges of the triangle-guided graph (Figure~\ref{fig:hesc_alpha005_directed}) and that its density increases with \(\alpha_{\mathrm{fm}}\) (from 92 to 162 skeleton edges for \(\alpha_{\mathrm{fm}}\) from 0.03 to 0.08).

\begin{table}[t]
\centering
\caption{
Main CPDAG comparison on the hESC gene-expression data (758 cells, 100 genes; full-joint B39, \(\texttt{mcs}=-1\)).
Since no ground-truth causal graph is available, we report graph-size and runtime metrics rather than causal recovery accuracy.
PC/FisherZ uses \(\alpha=0.02\) and GPC-FMCIT \(\alpha_{\mathrm{scr}}=0.10\), \(\alpha_{\mathrm{fm}}=0.05\), so the comparison is descriptive. 
}
\label{tab:hesc_cpdag_main}
\small
\setlength{\tabcolsep}{5pt}
\begin{adjustbox}{max width=\linewidth}
\begin{tabular}{lcccc}
\toprule
Method & Skeleton Edges & Directed & Undirected & Runtime (s) \\
\midrule
PC / FisherZ & 332 & 328 & 4 & 39.63 \\
GPC-FMCIT (stability guide) & 136 & 128 & 8 & 7256.98 \\
GPC-FMCIT (triangle guide) & {130} & {124} & 6 & 4488.56 \\
\bottomrule
\end{tabular}
\end{adjustbox}
\end{table}

\section{Conclusion}
We proposed FMCIT, a reusable conditional resampling CI test based on joint flow matching and Langevin-RePaint imputation, and GPC-FMCIT, a screen--guide--refine--orient framework for PC-style causal discovery.
The key design is to train one joint generative model and reuse it across many CI queries through masking, while controlling where expensive nonlinear refinement is invoked during graph search.
Experiments show improved CPDAG-level recovery over PC-stable baselines on additive SEMs, further gains from KCI screening on post-nonlinear and interaction SEMs, and compact graph estimates on hESC data.
Appendix~\ref{app:fmcit_theory} provides the theoretical proof; future work will focus on broader real-data evaluation.

\section*{Acknowledgments}
This work has been partially supported by JSPS KAKENHI 26K02990 and JSPS Grant-in-Aid for Transformative Research Areas (A) 22H05106. Shunyu Zhao and Yanfeng Yang are supported by JST SPRING, Japan Grant Number JPMJSP2104.  Shuai Li is partially supported by the Research Matching Grant from the Research Grants Council of the Hong Kong Special Administrative Region (project: 700006 Applications of SAS Viya in Big Data Analytics). The authors declare no competing interests.

\bibliographystyle{plain}
\bibliography{fmcit_references_neurips}

@book{spirtes2000causation,
	title={Causation, prediction, and search},
	author={Spirtes, Peter and Glymour, Clark N and Scheines, Richard},
	year={2000},
	publisher={MIT Press}
}

@inproceedings{zhang2011kernel,
	title={Kernel-based conditional independence test and application in causal discovery},
	author={Zhang, Kun and Peters, Jonas and Janzing, Dominik and Sch{\"o}lkopf, Bernhard},
	booktitle={Conference on Uncertainty in Artificial Intelligence},
	pages={804--813},
	year={2011},
}

@inproceedings{fukumizu2007kernel,
  title={Kernel measures of conditional dependence},
  author={Fukumizu, Kenji and Gretton, Arthur and Sun, Xiaohai and Sch{\"o}lkopf, Bernhard},
  booktitle={Advances in Neural Information Processing Systems},
  volume={20},
  year={2007},
}

@inproceedings{bellot2019conditional,
	title={Conditional independence testing using generative adversarial networks},
	author={Bellot, Alexis and van der Schaar, Mihaela},
	booktitle={Advances in Neural Information Processing Systems},
	volume={32},
	year={2019},
}

@article{shi2021double,
	title={Double generative adversarial networks for conditional independence testing},
	author={Shi, Chengchun and Xu, Tianlin and Bergsma, Wicher and Li, Lexin},
	journal={Journal of Machine Learning Research},
	volume={22},
        number={285},
	pages={1--32},
	year={2021},
publisher = {JMLR.org},
}

@article{li2020nonparametric,
  title={On nonparametric conditional independence tests for continuous variables},
  author={Li, Chun and Fan, Xiaodan},
  journal={Wiley Interdisciplinary Reviews: Computational Statistics},
  volume={12},
  number={3},
  pages={e1489},
  year={2020},
  publisher={Wiley Online Library}
}

@inproceedings{goodfellow2014generative,
  title={Generative adversarial nets},
  author={Goodfellow, Ian J and Pouget-Abadie, Jean and Mirza, Mehdi and Xu, Bing and Warde-Farley, David and Ozair, Sherjil and Courville, Aaron and Bengio, Yoshua},
  booktitle={Advances in Neural Information Processing Systems},
  volume={2},
  year={2014}
}

@inproceedings{scetbon2022asymptotic,
  title={An asymptotic test for conditional independence using analytic kernel embeddings},
  author={Scetbon, Meyer and Meunier, Laurent and Romano, Yaniv},
  booktitle={International Conference on Machine Learning},
  pages={19328--19346},
  year={2022},
}

@inproceedings{li2023nearest,
author = {Li, Shuai and Chen, Ziqi and Zhu, Hongtu and Wang, Christina Dan and Wen, Wang},
year = {2023},
month = {06},
pages = {8631-8639},
title = {Nearest-neighbor sampling based conditional independence testing},
volume = {37},
booktitle = {AAAI Conference on Artificial Intelligence},
}

@article{berrett2020conditional,
	title={The conditional permutation test for independence while controlling for confounders},
	author={Berrett, Thomas B and Wang, Yi and Barber, Rina Foygel and Samworth, Richard J},
	journal={Journal of the Royal Statistical Society: Series B (Statistical Methodology)},
	volume={82},
	number={1},
	pages={175--197},
	year={2020},
	publisher={Wiley Online Library}
	}

@inproceedings{runge2018conditional,
	title={Conditional independence testing based on a nearest-neighbor estimator of conditional mutual information},
	author={Runge, Jakob},
	booktitle={International Conference on Artificial Intelligence and Statistics},
	pages={938--947},
	year={2018}
	}

@article{candes2018panning,
	title={Panning for gold: `model-X' knockoffs for high dimensional controlled variable selection},
	author={Cand{\`e}s, Emmanuel and Fan, Yingying and Janson, Lucas and Lv, Jinchi},
	journal={Journal of the Royal Statistical Society: Series B (Statistical Methodology)},
	volume={80},
	number={3},
	pages={551--577},
	year={2018},
	publisher={Wiley Online Library}
}

@book{pearl1988probabilistic,
  title={Probabilistic reasoning in intelligent systems: networks of plausible inference},
  author={Pearl, Judea},
  year={1988},
  publisher={Morgan kaufmann}
}

@article{dai2022significance,
  title={Significance tests of feature relevance for a black-box learner},
  author={Dai, Ben and Shen, Xiaotong and Pan, Wei},
  journal={IEEE Transactions on Neural Networks and Learning Systems},
  volume={35},
  number={2},
  pages={1898--1911},
  year={2022},
  publisher={IEEE}
}

@inproceedings{li2024k,
  title={K-nearest-neighbor local sampling based conditional independence testing},
  author={Li, Shuai and Zhang, Yingjie and Zhu, Hongtu and Wang, Christina and Shu, Hai and Chen, Ziqi and others},
  booktitle={Advances in Neural Information Processing Systems},
  volume={36},
  year={2024}
}

@inproceedings{ho2020denoising,
  title={Denoising diffusion probabilistic models},
  author={Ho, Jonathan and Jain, Ajay and Abbeel, Pieter},
  booktitle={Advances in Neural Information Processing Systems},
  volume={33},
  year={2020}
}

@inproceedings{song2021score,
  title={Score-based generative modeling through stochastic differential equations},
  author={Song, Yang and Sohl-Dickstein, Jascha and Kingma, Diederik P and Kumar, Abhishek and Ermon, Stefano and Poole, Ben},
  booktitle={International Conference on Learning Representations},
  year={2021},
}

@inproceedings{yang_2025_cdcit,
author = {Yang, Yanfeng and Li, Shuai and Zhang, Yingjie and Sun, Zhuoran and Shu, Hai and Chen, Ziqi and Zhang, Renming},
title = {Conditional diffusion models based conditional independence testing},
year = {2025},
booktitle = {AAAI Conference on Artificial Intelligence},
articleno = {2456},
numpages = {9},
}

@article{guan_2025_ekci,
  title   = {Efficient Ensemble Conditional Independence Test Framework for Causal Discovery},
  author  = {Guan, Zhengkang and Kuang, Kun},
  journal = {arXiv:2509.21021},
  year    = {2025}
}

@article{he_2025_transportmap,
  title   = {From Conditional to Unconditional Independence: Testing Conditional Independence via Transport Maps},
  author  = {He, Chenxuan and Gao, Yuan and Zhu, Liping and Huang, Jian},
  journal = {arXiv:2504.09567},
  year    = {2025}
}

@article{miyazaki_2025_spectralgcm,
  title   = {Testing Conditional Independence via the Spectral Generalized Covariance Measure: Beyond Euclidean Data},
  author  = {Miyazaki, Ryunosuke and Uematsu, Yoshimasa},
  journal = {arXiv:2511.15453},
  year    = {2025}
}

@inproceedings{
he_2025_onthehardness,
title={On the Hardness of Conditional Independence Testing In Practice},
author={Zheng He and Roman Pogodin and Yazhe Li and Namrata Deka and Arthur Gretton and Danica J. Sutherland},
booktitle={Advances in Neural Information Processing Systems},
year={2025},
}

@article{Rajen_2020_aos,
author = {Rajen D. Shah and Jonas Peters},
title = {{The hardness of conditional independence testing and the generalised covariance measure}},
volume = {48},
journal = {The Annals of Statistics},
number = {3},
pages = {1514 -- 1538},
year = {2020},
}

@article{Wang_2015_cdc,
author = {Xueqin Wang and Wenliang Pan and Wenhao Hu and Yuan Tian and Heping Zhang},
title = {Conditional Distance Correlation},
journal = {Journal of the American Statistical Association},
volume = {110},
number = {512},
pages = {1726--1734},
year = {2015},
}

@inproceedings{ren_2025_sgmcit,
author = {Ren, Yixin and Jin, Chenghou and Xia, Yewei and Ke, Li and Huang, Longtao and Xue, Hui and Zhang, Hao and Guan, Jihong and Zhou, Shuigeng},
title = {Score-based Generative Modeling for Conditional Independence Testing},
year = {2025},
booktitle = {ACM SIGKDD Conference on Knowledge Discovery and Data Mining},
pages = {2410–2419},
numpages = {10},
}

@article{Gretton_2012_twosammpletest,
author = {Gretton, Arthur and Borgwardt, Karsten M. and Rasch, Malte J. and Sch\"{o}lkopf, Bernhard and Smola, Alexander},
title = {A kernel two-sample test},
year = {2012},
volume = {13},
issn = {1532-4435},
journal = {Journal of Machine Learning Research},
pages = {723–773},
numpages = {51},
}

@inproceedings{
lipman_2023_fm,
title={Flow Matching for Generative Modeling},
author={Yaron Lipman and Ricky T. Q. Chen and Heli Ben-Hamu and Maximilian Nickel and Matthew Le},
booktitle={International Conference on Learning Representations },
year={2023},
}

@article{hu_2025_FlowTS,
  title   = {{FlowTS}: Time Series Generation via Rectified Flow},
  author  = {Hu, Yang and Wang, Xiao and Ding, Zezhen and Wu, Lirong and Zhang, Huatian and Li, Stan Z. and Wang, Sheng and Zhang, Jiheng and Li, Ziyun and Chen, Tianlong},
  journal = {arXiv 2411.07506},
  year    = {2025}
}

@article{kalisch2007estimating,
  title={Estimating high-dimensional directed acyclic graphs with the PC-algorithm.},
  author={Kalisch, Markus and B{\"u}hlman, Peter},
  journal={Journal of Machine Learning Research},
  volume={8},
  number={3},
  year={2007}
}

@article{colombo2014order,
  title={Order-independent constraint-based causal structure learning},
  author={Colombo, Diego and Maathuis, Marloes H},
  journal={Journal of Machine Learning Research},
  volume={15},
  number={1},
  pages={3741--3782},
  year={2014}
}

@INPROCEEDINGS{Lugmayr_2022_repaint,
  author={Lugmayr, Andreas and Danelljan, Martin and Romero, Andres and Yu, Fisher and Timofte, Radu and Van Gool, Luc},
  booktitle={Computer Vision and Pattern Recognition}, 
  title={RePaint: Inpainting using Denoising Diffusion Probabilistic Models}, 
  year={2022},
  volume={},
  number={},
  pages={11451-11461},
}

@inproceedings{Ramsey_2006_cpc,
author = {Ramsey, Joseph and Spirtes, Peter and Zhang, Jiji},
title = {Adjacency-faithfulness and conservative causal inference},
year = {2006},
booktitle = {Uncertainty in Artificial Intelligence},
pages = {401–408},
numpages = {8},
}

@article{ramsey2016improving,
  title={Improving accuracy and scalability of the pc algorithm by maximizing p-value},
  author={Ramsey, Joseph},
  journal={arXiv:1610.00378},
  year={2016}
}

@inproceedings{
liu_2023_recflow,
title={Flow Straight and Fast: Learning to Generate and Transfer Data with Rectified Flow},
author={Xingchao Liu and Chengyue Gong and qiang liu},
booktitle={International Conference on Learning Representations },
year={2023},
}

@inproceedings{Spirtes_1995_fci,
author = {Spirtes, Peter and Meek, Christopher and Richardson, Thomas},
title = {Causal inference in the presence of latent variables and selection bias},
year = {1995},
booktitle = {Uncertainty in Artificial Intelligence},
pages = {499–506},
numpages = {8},
}

@inproceedings{Meek_1995_Meekrules,
author = {Meek, Christopher},
title = {Causal inference and causal explanation with background knowledge},
year = {1995},
booktitle = {Uncertainty in Artificial Intelligence},
pages = {403–410},
numpages = {8},
}

@inproceedings{Lopez_2013_rdc,
author = {Lopez-Paz, David and Hennig, Philipp and Sch\"{o}lkopf, Bernhard},
title = {The Randomized Dependence Coefficient},
year = {2013},
booktitle = {Advances in Neural Information Processing Systems},
pages = {1–9},
numpages = {9},
}

@article{pratapa2020beeline,
  title={Benchmarking algorithms for gene regulatory network inference from single-cell transcriptomic data},
  author={Pratapa, Aditya and Jalihal, Amogh P. and Law, Jeffrey N. and Bharadwaj, Aditya and Murali, T. M.},
  journal={Nature Methods},
  volume={17},
  pages={147--154},
  year={2020},
  doi={10.1038/s41592-019-0690-6}
}

@article{yang_2026_dacit,
      title={Domain-Adapted Diffusion Models for Conditional Independence Testing}, 
      author={Yanfeng Yang and Junda Zhao and Yijie Gao and Jiaqi Yang and Xinyu Shi and Ziqi Chen and Shunyu Zhao and Shuai Li and Wei Huang and Eshant English and Kenji Fukumizu},
      year={2026},
      journal={arXiv:2609.33490},
}

@article{li_2026_jit,
      title={Back to Basics: Let Denoising Generative Models Denoise}, 
      author={Tianhong Li and Kaiming He},
      year={2026},
      journal={arXiv:2511.13720},
}

@article{roberts_tweedie_1996,
 author = {Gareth O. Roberts and Richard L. Tweedie},
 journal = {Bernoulli},
 number = {4},
 pages = {341--363},
 title = {Exponential Convergence of Langevin Distributions and Their Discrete Approximations},
 urldate = {2026-10-27},
 volume = {2},
 year = {1996}
}

@article{Brigati_2026_log_sobolev,
  author  = {Brigati, Giovanni and Dolbeault, Jean and Simonov, Nikita},
  title   = {Logarithmic Sobolev Inequalities: A Review on Stability and Instability Results},
  journal = {La Matematica},
  year    = {2026},
  volume  = {5},
  number  = {1},
  pages   = {5},
}

@inproceedings{Vempala_2019_log_sobolev,
 author = {Vempala, Santosh and Wibisono, Andre},
 booktitle = {Advances in Neural Information Processing Systems},
 editor = {H. Wallach and H. Larochelle and A. Beygelzimer and F. d\textquotesingle Alch\'{e}-Buc and E. Fox and R. Garnett},
 pages = {},
 title = {Rapid Convergence of the Unadjusted Langevin Algorithm: Isoperimetry Suffices},
 volume = {32},
 year = {2019}
}

\newpage
\appendix

\section{Preliminaries}  
\label{sec_preliminary}
In this section, we introduce the framework of CRT, PC algorithm, and its variants. 

\subsection{Conditional randomization test (CRT)}
\label{sec_crt}

Note that the definition of CI in (\ref{eq_cit_def}) is equivalent to $P_{X_i | X_j,X_S} = P_{X_i | X_S}$, where $P_{X_i | X_j,X_S}$ denotes the conditional distribution of $X_i$ given $(X_j,X_S)$. According to Bayes' rule, CI is also equivalent to $P_{X_i,X_j,X_S} = P_{X_i | X_S}  P_{X_j,X_S}$, where $P_{X_i,X_j,X_S}$ and $P_{X_j,X_S}$ are the joint distributions of $(X_i,X_j,X_S)$ and $(X_j,X_S)$, respectively.  
Suppose we have $n$ i.i.d. samples from the distribution $P_X$ of $X$, denoted by $\mathbf{X} \in \mathbb{R}^{n \times p}$. \citep{candes2018panning} reformulates the CI testing problem as a two-sample testing problem \citep{Gretton_2012_twosammpletest}. 
Specifically, it compares the original samples $(\mathbf{X}_i, \mathbf{X}_j, \mathbf{X}_S)$ with the randomized samples $(\mathbf{X}_i^{(b)}, \mathbf{X}_j, \mathbf{X}_S)$ for $b = 1, \ldots, B$. 
Here, $\mathbf{X}_i$, $\mathbf{X}_j$, and $\mathbf{X}_S$ denote samples drawn from $X_i$, $X_j$, and $X_S$, respectively, while $\mathbf{X}_i^{(b)}$ are independent samples drawn from the conditional distribution $P_{X_i | X_S = \mathbf{X}_S}$.  
Let $T: \mathbb{R}^{n} \times \mathbb{R}^{n} \times \mathbb{R}^{n \times |S|} \rightarrow \mathbb{R}$ be a test statistic that measures the conditional dependence between the $n$ samples of $X_i$ and $X_j$ given $X_S$, where larger values of $T$ indicate stronger conditional dependence. 
When the null hypothesis $H_0$ holds, that is, when $X_i$ and $X_j$ are conditionally independent given $X_S$, the value of $T(\mathbf{X}_i, \mathbf{X}_j, \mathbf{X}_S)$ should be comparable to that of $T(\mathbf{X}_i^{(b)}, \mathbf{X}_j, \mathbf{X}_S)$. 
In contrast, when the alternative hypothesis $H_1$ holds, i.e., when $X_i$ and $X_j$ are conditionally dependent given $X_S$, the value of $T(\mathbf{X}_i, \mathbf{X}_j, \mathbf{X}_S)$ should be significantly larger than that of $T(\mathbf{X}_i^{(b)}, \mathbf{X}_j, \mathbf{X}_S)$, since $\mathbf{X}_i^{(b)}$ are drawn from the conditionally independent distribution $P_{X_i | X_S = \mathbf{X}_S}$. Finally, the $p$-value of the CRT is given by
\begin{equation}
\label{eq_CRT_pval}    
    \frac{1+\sum_{b=1}^{B} \mathbf{I} \{ T(\mathbf{X}_i^{(b)}, \mathbf{X}_j, \mathbf{X}_S) \geq T(\mathbf{X}_i, \mathbf{X}_j, \mathbf{X}_S) \}}{1+B},
\end{equation}
where $\bm{\mbox{I}}\{ \cdot\}$ is the indicator function. In our CI testing procedure, we adopt the Randomized Dependence Coefficient (RDC) as the test statistics $T$ \citep{Lopez_2013_rdc}.

\subsection{PC algorithm and its variants}
\label{sec_pc}

PC algorithm \citep{spirtes2000causation} is a constraint-based causal discovery method that estimates a Markov equivalence class of DAGs, represented as a CPDAG, from CI relations inferred from data.

\paragraph{Assumptions.}
PC algorithm is typically presented under the \emph{Causal Markov} and \emph{Causal Faithfulness} assumptions with respect to an underlying DAG, together with i.i.d.\ sampling.
In addition, the vanilla PC algorithm assumes \emph{causal sufficiency} (no latent confounders); extensions such as FCI relax this assumption (see ``Variants'' below).

\paragraph{Skeleton learning (edge deletion).}
PC algorithm starts from the complete undirected graph $G^{(0)}$ over $[p]$ and iteratively removes edges using conditional independence tests.
Let $\mathrm{Adj}_{G}(i)$ denote the current adjacency set of node $i$ in an undirected graph $G$.
At conditioning level $\ell=0,1,2,\dots$, for each adjacent pair $(i,j)$ in the current graph, PC searches for a separating set
\begin{equation}
S \subseteq \mathrm{Adj}_{G}(i)\setminus\{j\}, \qquad |S|=\ell,
\end{equation}
such that the CI test accepts
\begin{equation}
\label{eq:pc_ci_test}
X_i \perp\!\!\!\perp X_j | X_S.
\end{equation}
If such an $S$ is found, the edge $i-j$ is removed and the separating set is stored as $\mathrm{Sep}(i,j)\leftarrow S$ (and symmetrically $\mathrm{Sep}(j,i)\leftarrow S$).
The algorithm increases $\ell$ until a predefined maximum conditioning size is reached or no further edges can be removed.
Since the search over conditioning sets is combinatorial, the skeleton stage is typically the computational bottleneck in practice, especially in high dimensions \citep{kalisch2007estimating}.

\paragraph{Orientation (v-structures and propagation).}
After skeleton learning, PC orients edges in two steps.
First, for every unshielded triple $i-k-j$ (i.e., $i$ adjacent to $k$, $k$ adjacent to $j$, but $i$ and $j$ non-adjacent), PC orients a v-structure $i \rightarrow k \leftarrow j$ whenever $k \notin \mathrm{Sep}(i,j)$.
Second, PC applies sound orientation propagation rules (e.g., Meek rules) to orient as many remaining edges as possible without introducing directed cycles or new v-structures.

\paragraph{Order-independence and variants.}
A well-known practical issue is that the adjacency search in skeleton learning can depend on the order in which CI tests are performed.
The PC-stable variant  \citep{colombo2014order} addresses this by using an order-independent deletion scheme (often referred to as a ``stable'' PC skeleton stage).
Other variants include conservative PC (CPC) \citep{Ramsey_2006_cpc} and PC-Max \citep{ramsey2016improving}, which modify v-structure orientation rules to reduce false orientations, and FCI \citep{Spirtes_1995_fci}, which extends the framework to settings with latent confounding and selection bias.
In this work, FMCIT is used both for skeleton refinement and in the conservative orientation stage, which uses the recorded separating sets \citep{Meek_1995_Meekrules}; in our experiments, orientation accounts for most FMCIT queries (Table~\ref{tab:component_ablation}).

\section{Algorithm of FMCIT}  \label{all_algorithms}
In this section, we introduce the algorithm of training a FM model, sampling from a trained FM model, and the complete FMCIT procedure.

\begin{algorithm}[htbp]
\caption{Training a flow matching model}
\label{alg_train_fm}
\begin{algorithmic}[1]
\REQUIRE Data matrix $\mathbf{X} \in \mathbb{R}^{n\times p}$.
\ENSURE A trained neural network $\widehat v_t$.

\STATE Initialize a neural network $\widehat v_t$

\WHILE{Not converge}
\STATE Draw $t \sim U(0,1)$
\STATE Draw $\mathbf{X}(0) \sim N(0 , I)$
\STATE Calculate $L_{FM} = \| \mathbf{X} - \mathbf{X}(0) - \widehat v_t( t \mathbf{X} + (1-t) \mathbf{X}(0) ) \|^2$
\STATE Calculate $\nabla_{\theta} L_{FM}$ and update the parameters of $\widehat v_t$ 

\ENDWHILE
\STATE \textbf{Output} $\widehat v_t$
\end{algorithmic}
\end{algorithm}

\begin{algorithm}[htbp]
\caption{Conditional sampling via Langevin-RePaint,
\textsc{FM-Impute}($\mathbf X_S$)}
\label{alg_sample_fm}
\begin{algorithmic}[1]
\REQUIRE Observed coordinates $\mathbf X_S$,
trained velocity field $\widehat v_t$,
sampling steps $L$, and uniform grid
$t_\ell=\ell/L$.

\ENSURE An approximate sample from
$P_{X\mid X_S=\mathbf X_S}$.

\STATE Set $\delta_L=L^{-1/4}$ and
$\ell_0=\lceil L(1-\delta_L)\rceil$.
\STATE Draw $\mathbf X(t_0)\sim\mathcal N(0,I_p)$.

\FOR{$\ell=0,\ldots,L-1$}

    \IF{$\ell<\ell_0$}
        \STATE \textbf{Picard warm-up:}
        \STATE $\displaystyle
        \widehat{\mathbf X}_{t_\ell}(1)
        \gets
        \mathbf X(t_\ell)
        +(1-t_\ell)
        \widehat v_{t_\ell}(\mathbf X(t_\ell))$.
    \ELSE
        \STATE \textbf{Half-score terminal correction:}
        \STATE $\displaystyle
        \widehat{\mathbf X}_{t_\ell}(1)
        \gets
        \frac{1+t_\ell}{2t_\ell}\mathbf X(t_\ell)
        +\frac{1-t_\ell}{2}
        \widehat v_{t_\ell}(\mathbf X(t_\ell))$.
    \ENDIF

    \STATE Replace the coordinates of
    $\widehat{\mathbf X}_{t_\ell}(1)$ indexed
    by $S$ with the observed values $\mathbf X_S$.

    \IF{$\ell<L-1$}
        \STATE Draw
        $\varepsilon_{\ell+1}\sim\mathcal N(0,I_p)$.
        \STATE $\displaystyle
        \mathbf X(t_{\ell+1})
        \gets
        t_{\ell+1}\widehat{\mathbf X}_{t_\ell}(1)
        +(1-t_{\ell+1})\varepsilon_{\ell+1}$.
    \ENDIF

\ENDFOR

\STATE \textbf{Return}
$\widehat{\mathbf X}_{t_{L-1}}(1)$.

\end{algorithmic}
\end{algorithm}

\begin{algorithm}[htb!]
\caption{Flow matching based conditional independence test, \textsc{FMCIT}$(\mathbf{X}_i \perp\!\!\!\perp \mathbf{X}_j | \mathbf{X}_S)$}
\label{alg_fmcit}
\begin{algorithmic}[1]
\REQUIRE Samples of whole dataset $\mathbf{X}$, variable index $(i,j)$, conditioning set $S$ and conditioning samples $\mathbf{X}_S$, number of CRT randomizations $B$, test statistics $T(\cdot,\cdot,\cdot)$, significance level $\alpha$. 

\ENSURE The CRT $p$-value $p_{\mathrm{FMCIT}}$.

\STATE Repeat $\mathbf{X}_S$ for $B$ times, and let the repeated sample be $\mathbf{X}_S^{(1:B)} \in \mathbb{R}^{nB \times |S| }$

\STATE Generate ${\mathbf{X}}^{(1:B)}$ =  \textsc{FM-Impute}($\mathbf{X}_S^{(1:B)}$)

\STATE Separate ${\mathbf{X}}^{(1:B)}$ into $B$ samples ${\mathbf{X}}^{(1)}, \hdots, {\mathbf{X}}^{(B)}$
 
\STATE Let the $i$-th and $j$-th variable in $\mathbf{X}$ be $ \mathbf{X}_i$ and $\mathbf{X}_j$
\STATE $T = T(\mathbf{X}_i,\mathbf{X}_j,\mathbf{X}_S)$
\FOR{$b=1,\hdots B$} 
\STATE Let the $i$-th variable in ${\mathbf{X}}^{(b)}$ be $\mathbf{X}_i^{(b)}$
\STATE Calculate $T^{(b)} = T(\mathbf{X}_i^{(b)},\mathbf{X}_j,\mathbf{X}_S)$
\ENDFOR

\STATE Calculate $p = \frac{1+\sum_{b=1}^{B} \mathbf{I} \{ T^{(b)} \geq T \}}{1+B}$
\STATE \textbf{Return} $p$
\end{algorithmic}
\end{algorithm}

\begin{algorithm}[htbp]
\caption{GPC-FMCIT: plug-in screened CPDAG learning with FMCIT}
\label{alg:gpc_fmcit}
\begin{algorithmic}[1]
\REQUIRE Data matrix \(\mathbf X\in\mathbb R^{n\times p}\); screening module \(\mathcal S_{\mathrm{scr}}\); guidance rule \(\mathrm{rule}\); pool size \(K\); per-edge budget \(M\); FMCIT level \(\alpha_{\mathrm{fm}}\); refinement depth \(d_{\mathrm{ref}}\); orientation depth \(d_{\mathrm{ori}}\).
\ENSURE Estimated CPDAG \(\widehat{\mathcal C}\), refined skeleton \(\widehat G\), and separating sets \(\mathrm{Sep}\).

\vspace{2pt}
\STATE \textbf{Screen.} Obtain an undirected candidate graph \(G_{\mathrm{scr}}=([p],E_{\mathrm{scr}})\leftarrow \mathcal S_{\mathrm{scr}}(\mathbf X)\).
\STATE \(\triangleright\) Examples: FisherZ screening for the scalable version; a KCI-based screening module for the strong-screen version.
\STATE Initialize \(G\leftarrow G_{\mathrm{scr}}\) and \(\mathrm{Sep}\leftarrow\emptyset\).

\vspace{2pt}
\STATE \textbf{Guide.} From \(G_{\mathrm{scr}}\), compute guidance information according to \(\mathrm{rule}\in\{\texttt{stab},\texttt{tri}\}\), including component labels and a global node priority order.

\vspace{2pt}
\STATE \textbf{Refine.}
\FOR{\(\ell=0,1,2,\ldots\)}
    \IF{\(d_{\mathrm{ref}}\ge 0\) and \(\ell>d_{\mathrm{ref}}\)}
        \STATE \textbf{break}
    \ENDIF
    \STATE \(\mathcal R\leftarrow\emptyset\), \(\mathrm{feasible}\leftarrow\mathrm{false}\).
    \FOR{each current edge \((i,j)\in E(G)\)}
        \STATE Form \(\mathcal Q_{ij}=(\mathrm{Adj}_{G_{\mathrm{scr}}}(i)\cup \mathrm{Adj}_{G_{\mathrm{scr}}}(j))\setminus\{i,j\}\).
        \STATE Build an ordered pool \(\mathcal P_{ij}\subseteq\mathcal Q_{ij}\) with \(|\mathcal P_{ij}|\le K\) using the guidance rule.
        \STATE Select at most \(M\) conditioning sets \(\mathcal S_{ij}^{(\ell)}\subseteq\binom{\mathcal P_{ij}}{\ell}\).
        \IF{\(\mathcal S_{ij}^{(\ell)}\neq\emptyset\)}
            \STATE \(\mathrm{feasible}\leftarrow\mathrm{true}\).
        \ENDIF
        \FOR{each \(S\in\mathcal S_{ij}^{(\ell)}\)}
            \STATE Compute \(p_{ij\mid S}\leftarrow \textsc{FMCIT}(\mathbf X_i\perp\!\!\!\perp \mathbf X_j\mid \mathbf X_S)\).
            \IF{\(p_{ij\mid S}>\alpha_{\mathrm{fm}}\)}
                \STATE Add \((i,j,S)\) to \(\mathcal R\) and \textbf{break}.
            \ENDIF
        \ENDFOR
    \ENDFOR
    \FOR{each \((i,j,S)\in\mathcal R\)}
        \STATE Remove \(i-j\) from \(G\), and set \(\mathrm{Sep}(i,j)=\mathrm{Sep}(j,i)=S\).
    \ENDFOR
    \IF{\(d_{\mathrm{ref}}<0\) and \(\mathrm{feasible}=\mathrm{false}\)}
        \STATE \textbf{break}
    \ENDIF
\ENDFOR
\STATE Set \(\widehat G\leftarrow G\).

\vspace{2pt}
\STATE \textbf{Orient.} Initialize \(\widehat{\mathcal C}\leftarrow \widehat G\). Apply conservative collider orientation using \(\mathrm{Sep}\) and optional FMCIT checks up to depth \(d_{\mathrm{ori}}\), then apply Meek propagation rules.
\STATE \textbf{Complete.} If \(\widehat{\mathcal C}\) is not a valid completed PDAG, apply the deterministic completed-PDAG projection of Appendix~\ref{app:cpdag_validity}.
\STATE \textbf{Return} \(\widehat{\mathcal C}\), \(\widehat G\), and \(\mathrm{Sep}\).
\end{algorithmic}
\end{algorithm}


\section{Experimental settings}
\label{app:exp_settings}

\subsection{Synthetic conditional independence testing benchmark}
\label{app:sim_cit}

In this subsection, we provide a complete description of the synthetic CI testing benchmark used to evaluate FMCIT as a standalone CI test. Our goal is to examine whether FMCIT can simultaneously maintain near-nominal type-I error control and high testing power under nonlinear relationships, relatively high-dimensional conditioning sets, and settings where the conditioning information may gradually weaken.

\paragraph{Baselines and metrics.}
We compare FMCIT with representative CI testing methods, including NNLSCIT \citep{li2024k}, KCIT \citep{zhang2011kernel}, EKCI \citep{guan_2025_ekci}, GCIT \citep{bellot2019conditional}, and CDCIT \citep{yang_2025_cdcit}. These baselines cover several major technical routes, including $k$-nearest neighbor resampling, kernel-based methods, and generative CRT-based approaches. All methods are evaluated under the same data-generating configurations and significance level. For each configuration, we repeat the experiment independently 100 times and report the empirical type-I error and testing power at significance level $\alpha=0.05$.

\paragraph{Data generation.}
We construct a synthetic benchmark using a post-nonlinear data-generating mechanism that is widely used in the CI testing literature \citep{scetbon2022asymptotic,li2024k,yang_2025_cdcit,ren_2025_sgmcit}. Let $Z=(Z_1,\ldots,Z_{d_z})^\top$ denote the conditioning vector, where the components are independently drawn from the standard normal distribution, and define $m=\frac{1}{d_z}\sum_{k=1}^{d_z} Z_k$.
We then generate $X$ and $Y$ and test whether $X \perp\!\!\!\perp Y \mid Z$. More specifically, the data-generating processes under the null and alternative hypotheses are
\begin{align}
H_0:&\quad X = f_1(m + s\varepsilon_x), \qquad
Y = f_2(m + s\varepsilon_y), \notag\\
H_1:&\quad X = f_1(m + s\varepsilon_x) + 0.5 e_b, \quad
Y = f_2(m + s\varepsilon_y) + 0.5 e_b,  \label{eq_ci_benchmark}
\end{align}
where $s$ is a hyperparameter controlling the noise intensity, $\varepsilon_x,\varepsilon_y,e_b \stackrel{\mathrm{i.i.d.}}{\sim} N(0,1)$, and $f_1,f_2$ are randomly selected from $\{x,\;x^2,\;\cos(x),\;x^3,\;\tanh(x)\}$.
This construction introduces nonlinear transformations, stochastic perturbations, and variation in the conditioning-set dimension, thereby providing a systematic testbed for evaluating the robustness of different CI methods under complex conditioning structures. In the experiments, we consider conditioning-set dimensions $d_z\in\{10,50\}$, fix the sample size at $n=1000$, and vary the noise scale over $s\in\{0.50,0.75,1.00\}$.


\textbf{The flow matching model.}
For FMCIT, the vector field $\widehat v_t$ is parameterized by a multilayer perceptron (MLP) with two hidden layers, each of width $256$. We train $\widehat v_t$ for $200$ epochs using Adam with learning rate $10^{-3}$ and batch size $1024$. Model selection is done by the smallest training loss.

\textbf{Conditional sampler.}
Conditional resamples are generated by the Langevin-RePaint FM imputation sampler with a uniform time grid
\[
t_\ell = \ell/L, \ell = 0,\ldots,L
\]
and $L=50$ sampling steps.

\textbf{CRT configuration and test statistic.}
Each CI query is evaluated using CRT with $B=100$ randomizations, and the $p$-value is computed as in Eq.~(\ref{eq_CRT_pval}). We use the randomized dependence coefficient (RDC) as the dependence statistic inside CRT. Unless otherwise stated, we use RDC hyperparameters $k=10$, $s=1/3$, and $n=1$, and fix the projection seed for reproducibility.

\paragraph{Results.}
Table~\ref{tab:cit_post_nonlinear_transposed} reports the results. Overall, FMCIT exhibits stable statistical behavior under both conditioning-set dimensions: its empirical type-I error remains close to the nominal level, while achieving high testing power in most settings. In particular, under the high-dimensional conditioning-set regime with $d_z=50$, FMCIT still maintains strong power, suggesting that its conditional resampling mechanism does not deteriorate substantially as the conditioning dimension increases. By contrast, GCIT, KCIT, and EKCI show more noticeable deviations from the nominal level in terms of type-I error control in this benchmark, whereas NNLSCIT and CDCIT suffer a more pronounced loss of power as the noise scale increases. 

\begin{table}[t]
\centering
\caption{Post-nonlinear synthetic CI benchmark ($n=1000$, $\alpha=0.05$, 100 trials). We report empirical type-I error and power.}
\label{tab:cit_post_nonlinear_transposed}
\small
\setlength{\tabcolsep}{4pt}
\begin{adjustbox}{max width=\linewidth}
\begin{tabular}{llcccccc}
\toprule
\multirow{2}{*}{Method} & \multirow{2}{*}{Metric}
& \multicolumn{3}{c}{$d_z=10$}
& \multicolumn{3}{c}{$d_z=50$} \\
\cmidrule(lr){3-5}\cmidrule(lr){6-8}
& & $s=0.50$ & $s=0.75$ & $s=1.00$ & $s=0.50$ & $s=0.75$ & $s=1.00$ \\
\midrule
\multirow{2}{*}{FMCIT (Ours)}
& Type-I $\downarrow$ & 0.08 & 0.05 & 0.04 & 0.05 & 0.04 & 0.05 \\
& Power $\uparrow$    & 1.00 & 1.00 & 1.00 & 1.00 & 1.00 & 0.98 \\
\midrule
\multirow{2}{*}{GCIT}
& Type-I $\downarrow$ & 0.25 & 0.18 & 0.17 & 0.21 & 0.22 & 0.18 \\
& Power $\uparrow$    & 1.00 & 1.00 & 0.98 & 1.00 & 0.99 & 1.00 \\
\midrule
\multirow{2}{*}{NNLSCIT}
& Type-I $\downarrow$ & 0.09 & 0.08 & 0.08 & 0.11 & 0.06 & 0.05 \\
& Power $\uparrow$    & 0.90 & 0.55 & 0.32 & 0.88 & 0.52 & 0.30 \\
\midrule
\multirow{2}{*}{KCIT}
& Type-I $\downarrow$ & 0.24 & 0.24 & 0.23 & 0.14 & 0.14 & 0.15 \\
& Power $\uparrow$    & 1.00 & 0.97 & 0.88 & 1.00 & 1.00 & 0.99 \\
\midrule
\multirow{2}{*}{EKCI}
& Type-I $\downarrow$ & 0.47 & 0.45 & 0.44 & 0.20 & 0.23 & 0.24 \\
& Power $\uparrow$    & 1.00 & 0.98 & 0.91 & 1.00 & 1.00 & 0.99 \\
\midrule
\multirow{2}{*}{CDCIT}
& Type-I $\downarrow$ & 0.07 & 0.04 & 0.04 & 0.08 & 0.07 & 0.05 \\
& Power $\uparrow$    & 0.94 & 0.74 & 0.62 & 0.97 & 0.80 & 0.69 \\
\bottomrule
\end{tabular}
\end{adjustbox}
\end{table}

The null-calibration experiment for FMCIT is reported in the main text in Table~\ref{tab:type1_n_calibration_main}. Overall, FMCIT delivers a more balanced empirical performance by maintaining near-nominal empirical type-I error and strong testing power under complex nonlinear conditioning structures.

\subsection{Time complexity of individual CITs}

In this section, we investigate the computational efficiency of individual FisherZ, KCIT, GCIT and CDCIT. The time complexity of FMCIT is reported in Table \ref{tab:type1_n_calibration_main}.
We evaluate these CIT methods on the simulation example described in Appendix~\ref{app:sim_cit}. 
Each CIT is repeated 100 times with $d_z=50$ and $n=500$. 

\begin{table}[htp]
\centering
\caption{Average computational time for a single CIT test over 100 repetitions with $d_z=50$ and $n=500$.}
\label{tab:cit_runtime}
\begin{tabular}{lc}
\toprule
Method & Time (s) \\
\midrule
FisherZ & $0.01$ \\
KCIT    & $0.38$ \\
GCIT    & $2.36$ \\
CDCIT   & $14.36$ \\
\bottomrule
\end{tabular}
\end{table}

From Table~\ref{tab:cit_runtime}, it can be observed that the runtime of a single FMCIT test is even shorter than that of GCIT, which uses only one sampling step, and is second only to FisherZ and KCIT.

\subsection{Causal discovery experiments}
\label{app:exp_pc}

\paragraph{Data-generating process.}
We generate synthetic observational data from random additive nonlinear structural equation models with heavy-tailed noise.
For each trial, we first sample a random DAG over $p$ nodes by drawing $\texttt{exp\_edges}$ directed edges that respect a fixed topological ordering, so that edges are only allowed from lower-indexed nodes to higher-indexed nodes.
Given the sampled DAG, variables are generated sequentially.
For each edge $i\to j$, we sample an edge weight $w_{ij}=s_{ij}u_{ij}$, where $u_{ij}\sim\mathrm{Unif}(0.7,1.3)$ and $s_{ij}\in\{+1,-1\}$ is chosen uniformly at random.
For each edge, we independently select a nonlinear transformation from $\{\tanh(x),\ \sin(x),\ 0.5x^2,\ \mathrm{sign}(x)\sqrt{|x|+10^{-8}},\ \max(x,0)+0.1\min(x,0)\}$.
Each variable is generated as $X_j=\sum_{i\in\mathrm{Pa}(j)}w_{ij}f_{ij}(X_i)+\varepsilon_j$, where $\varepsilon_j$ is sampled independently from a Student-$t$ distribution with degrees of freedom $\texttt{df}$.

\paragraph{Default setting.}
Unless otherwise stated, we set \(n=500\), \(p=100\), \(\texttt{exp\_edges}=150\), and \(\texttt{df}=3\).
The main causal discovery benchmark in Table~\ref{tab:sim_pc_main} uses ten matched seeds.
The sample-size experiment uses three seeds for computational reasons, and the component ablation, the budget and pool-size sensitivity study, and the alternative-SEM robustness study use five seeds each, as stated in their captions.
All methods are evaluated on the same generated datasets within each seed.
\paragraph{Graph-search setting.}
For the main causal discovery benchmark, we use the auto-adaptive conditioning-depth setting for both skeleton learning and CPDAG orientation.
In our implementation, this corresponds to setting $\texttt{mcs}=-1$ and $\texttt{orient\_max\_cond\_set}=-1$.
The adaptive search continues increasing the conditioning level until no feasible conditioning sets remain for the current graph.

\paragraph{Baseline methods.}
The main benchmark compares GPC-FMCIT with PC-stable(FisherZ) and PC-stable(KCI); the latter uses a kernel-based conditional independence test in the stable-PC skeleton stage followed by standard CPDAG orientation.
The component ablation additionally includes a direct PC-FMCIT baseline, which uses FMCIT as the CI oracle without screening or guided refinement, to isolate the effect of the proposed guidance mechanism.

\paragraph{GPC-FMCIT configuration.}
The FisherZ-screened GPC-FMCIT variant uses FisherZ-based screening, and the KCI-screened variant uses a KCI-based screening module to construct the candidate graph, while keeping the same downstream guidance, FMCIT refinement, and conservative orientation pipeline.
In the main benchmark, the screening and refinement levels are matched to the baseline level, \(\alpha_{\mathrm{scr}}=\alpha_{\mathrm{fm}}=\alpha\in\{0.05,0.10\}\); all other synthetic graph-level experiments use \(\alpha_{\mathrm{scr}}=\alpha_{\mathrm{fm}}=0.10\).
Unless otherwise stated, we use stability-based guidance with \(\texttt{stab\_alphas}=(0.08,0.10,0.12)\), a candidate pool of size \(K=4\) and a per-edge budget of \(M=2\).
The final graph is oriented using conservative CPDAG orientation followed by Meek propagation rules and the completed-PDAG projection of Appendix~\ref{app:cpdag_validity}.

\paragraph{FMCIT configuration for graph search.}
For FMCIT-based methods, a single FM model is trained on the full joint data and reused across CI queries.
Conditional samples are generated through the Langevin-RePaint imputation sampler.
Unless otherwise stated, we use RDC as the dependence statistic at both unconditional and conditional levels in the synthetic causal discovery experiments.
All synthetic graph-level experiments (main benchmark, component ablation, sample-size, budget and pool-size, and alternative-SEM experiments) use the same FMCIT configuration: the joint FM model is an MLP with width 256 and three hidden layers, trained for up to 300 epochs (with early stopping where applicable) with Adam (learning rate \(10^{-3}\), batch size 2048); the sampler uses 15 steps; and each FMCIT query uses 39 CRT replicates.
The standalone CI benchmark (Appendix~\ref{app:sim_cit}) and the hESC experiments (Appendix~\ref{app:hesc_sensitivity}) use their own settings, described there.

\paragraph{Evaluation metrics.}
We compare learned CPDAGs with the CPDAG induced by the ground-truth DAG.
All synthetic graph-level tables are scored with this CPDAG-based comparison.
At the skeleton level, we report precision, recall, and F1.
For orientation, we report strict directed precision, recall, and F1, as well as relaxed directed F1, where a true directed edge recovered as undirected receives partial credit.
We also report CPDAG pairwise error and wall-clock runtime in seconds; the timing boundary and hardware of each table are listed in Appendix~\ref{app:compute_environment}.

\paragraph{CPDAG validity and completion.}
\label{app:cpdag_validity}
A graph without directed cycles is not necessarily a valid completed PDAG.
Before scoring, every learned graph is checked for a valid encoding, directed acyclicity, the existence of a consistent DAG extension, and completed-PDAG validity; a graph that passes all four checks is left unchanged.
Neither of the two procedures below uses the ground-truth graph.

\emph{Completed-PDAG projection (GPC-FMCIT outputs).}
For an acyclic mixed graph that fails the checks, the projection computes a deterministic topological order consistent with the existing directed edges. At each step, it greedily favors an available vertex that creates fewer additional unshielded colliders among already ordered neighbors, breaking ties by total collider count, number of preceding adjacent vertices, and vertex index. It orients the remaining undirected edges according to this order and applies DAG-to-CPDAG conversion. This is a heuristic completion, not a globally optimal collider-minimization procedure.
It verifies that the skeleton is preserved and that no existing directed edge is reversed; the 40 audited main GPC outputs also preserve all original directed arrows. The projection may orient previously undirected edges and may change the unshielded-collider structure, so its output is not necessarily equivalent to the raw CPC--Meek output.
The projection does not accept graphs with directed cycles.

\emph{Baseline repair (PC-stable outputs).}
Raw PC-stable outputs can contain directed cycles.
For these baselines, a separate deterministic repair repeatedly finds a directed cycle and changes one arrow on that cycle into an undirected edge; it selects the arrow with the fewest supported unshielded colliders, using lexicographic tie-breaking. Once no directed cycle remains, the same acyclic completion and validity checks are applied. Valid raw baseline CPDAGs are left unchanged. This is not claimed to find a globally minimal repair.
The method names in Table~\ref{tab:sim_pc_main} refer to the PC-stable baselines followed by this repair, and we do not report them as unmodified native PC output.

Table~\ref{tab:projection_counts} reports how many final graphs were changed.
Because the projection changed nearly all GPC-FMCIT outputs in Table~\ref{tab:sim_pc_main}, the reported orientation metrics reflect conservative orientation followed by deterministic completion, not conservative orientation alone.
Across the 40 main GPC runs, completion preserves the skeleton and all 4,350 original directed edge instances and orients 300 previously undirected edge instances (7.5 per graph on average), of which 121 are correct and 179 are incorrect with respect to the true CPDAG.
To measure its effect, we replayed all 40 runs and scored the pre-completion mixed graphs with the same evaluation (Table~\ref{tab:orientation_counts}).
These pre-completion scores are orientation diagnostics rather than scores of valid CPDAGs, since 39 of the 40 pre-completion graphs are not valid completed PDAGs.
Completion raises mean strict directed F1 by 0.009 to 0.022 and lowers mean CPDAG pairwise error by 1.3 to 2.7, but it lowers strict directed F1 in 8 of the 40 runs.
In all four settings, the pre-completion strict directed F1 and pairwise error of GPC-FMCIT are already better than those of the corresponding repaired PC-stable baseline, so the advantages in Table~\ref{tab:sim_pc_main} are not produced by the completion step.
In the component ablation, all eight variants are scored on completed outputs, the completion time is included in the reported runtimes, and all 40 final graphs pass the four checks.

\begin{table}[t]
\centering
\caption{Number of final graphs changed by CPDAG completion (changed / total runs). GPC-FMCIT rows use the completed-PDAG projection; PC-stable rows use the separate baseline repair.}
\label{tab:projection_counts}
\small
\setlength{\tabcolsep}{4pt}
\begin{tabular}{llc}
\toprule
Experiment & Group & Changed / total \\
\midrule
Table~\ref{tab:sim_pc_main} & GPC-FMCIT, FisherZ screen, \(\alpha=0.05\) & 10/10 \\
 & GPC-FMCIT, KCI screen, \(\alpha=0.05\) & 10/10 \\
 & GPC-FMCIT, FisherZ screen, \(\alpha=0.10\) & 9/10 \\
 & GPC-FMCIT, KCI screen, \(\alpha=0.10\) & 10/10 \\
 & PC-stable (FisherZ and KCI), both levels & 12/40 \\
\midrule
Table~\ref{tab:kci_screen_n_sweep} & \(n=200\) / \(500\) / \(1000\) & 2/3, 3/3, 3/3 \\
Table~\ref{tab:dgp_robustness} & FisherZ screen / KCI screen & 0/10, 0/10 \\
Table~\ref{tab:budget_pool_ablation} & budget sweep / pool-size sweep & 15/20, 27/35 \\
\bottomrule
\end{tabular}
\end{table}

\begin{table}[t]
\centering
\caption{Effect of the completion step on GPC-FMCIT in Table~\ref{tab:sim_pc_main} (means over ten seeds). ``New (correct)'': edges oriented by completion and how many of them are correct; ``pre'' and ``post'': replayed pre-completion graph and final completed PDAG. PC-stable values are from Table~\ref{tab:sim_pc_main}.}
\label{tab:orientation_counts}
\small
\setlength{\tabcolsep}{3.5pt}
\begin{adjustbox}{max width=\linewidth}
\begin{tabular}{llccccccc}
\toprule
& & & \multicolumn{3}{c}{Strict Dir-F1 $\uparrow$} & \multicolumn{3}{c}{Pairwise error $\downarrow$} \\
\cmidrule(lr){4-6}\cmidrule(lr){7-9}
$\alpha$ & Screen & New (correct) & Pre & Post & PC-stable & Pre & Post & PC-stable \\
\midrule
0.05 & FisherZ & 10.1 (3.8) & 0.264 & 0.286 & 0.073 & 137.8 & 136.4 & 231.9 \\
0.05 & KCI & 6.1 (3.1) & 0.362 & 0.379 & 0.240 & 106.9 & 104.2 & 126.0 \\
0.10 & FisherZ & 6.0 (2.0) & 0.276 & 0.286 & 0.054 & 148.7 & 147.4 & 285.6 \\
0.10 & KCI & 7.8 (3.2) & 0.391 & 0.405 & 0.248 & 104.3 & 102.6 & 137.2 \\
\bottomrule
\end{tabular}
\end{adjustbox}
\end{table}

\paragraph{Component-ablation protocol.}
\label{app:ablation_protocol}
The component ablation uses five matched datasets (seeds 0, 1000, 2000, 3000, 4000) with \(n=500\), \(p=50\) and 75 DAG edges.
All FMCIT variants of a seed reuse the same pretrained full-joint FM checkpoint; no model is fitted during the measured runs.
Conditioning-set limits are unbounded.
Queries are distinct unordered variable-pair and conditioning-set tests, deduplicated across refinement and orientation.
All timings use one NVIDIA RTX 6000 Ada Generation GPU with 15 statistical worker processes, one main process and one native numerical thread per process; the 40 measurements run sequentially on the same GPU and on the same NUMA-local CPU cores.
Runs use fresh query caches and include screening, stability guidance, refinement, orientation and final CPDAG completion; worker start-up, input verification, evaluation and output serialization are excluded, and kernel initialization uses separate warm-up queries whose caches are discarded before measurement.
Each variant and seed is measured once.

\subsection{Scaling and sensitivity experiments}
\label{app:scaling_sensitivity}

\paragraph{Sample-size sensitivity.}
Table~\ref{tab:kci_screen_n_sweep} fixes \(p=100\) and varies the sample size \(n\) for KCI-screened GPC-FMCIT.
The \(n=500\) row uses the same three seeds as the other rows and is therefore a three-seed subset of the ten-seed benchmark in Table~\ref{tab:sim_pc_main}, not the same summary.
Skeleton F1 increases with \(n\) (\(0.779\), \(0.904\), \(0.952\)), whereas strict and relaxed directed F1 are highest at \(n=500\) and CPDAG error is similar at \(n=500\) and \(n=1000\) (\(100.0\) and \(101.3\)).
More samples therefore mainly improve adjacency recovery; with three seeds we do not interpret the small differences in orientation accuracy between \(n=500\) and \(n=1000\). Runtime is not reported because the \(n=200\) and \(n=1000\) runs created their FM checkpoints inside the timed stage, whereas the \(n=500\) runs reused existing checkpoints.

\begin{table}[t]
\centering
\caption{
Sample-size sensitivity of KCI-screened GPC-FMCIT
(\(p=100\), \(\texttt{exp\_edges}=150\), \(\mathrm{df}=3\),
\(\alpha_{\mathrm{scr}}=\alpha_{\mathrm{fm}}=0.10\)).
Mean \(\pm\) sample standard deviation over seeds 0, 1000 and 2000.
}
\label{tab:kci_screen_n_sweep}
\small
\setlength{\tabcolsep}{4pt}
\begin{adjustbox}{max width=\linewidth}
\begin{tabular}{ccccc}
\toprule
\(n\) & Skel-F1 $\uparrow$ & Dir-F1 $\uparrow$ & Relaxed Dir-F1 $\uparrow$ & CPDAG Error $\downarrow$ \\
\midrule
200 & $0.779 \pm 0.023$ & $0.383 \pm 0.033$ & $0.392 \pm 0.033$ & $108.7 \pm 2.3$ \\
500 & $0.904 \pm 0.045$ & $0.417 \pm 0.018$ & $0.422 \pm 0.019$ & $100.0 \pm 4.4$ \\
1000 & $0.952 \pm 0.029$ & $0.389 \pm 0.012$ & $0.390 \pm 0.010$ & $101.3 \pm 3.1$ \\
\bottomrule
\end{tabular}
\end{adjustbox}
\end{table}

\subsection{Robustness to graph-level nonlinear SEMs}
\label{app:dgp_robustness}

Table~\ref{tab:dgp_robustness} reports results on post-nonlinear and interaction SEMs for the two screening variants of GPC-FMCIT.
The post-nonlinear SEM applies a node-specific nonlinear transformation to the sum of nonlinear parent effects and noise; the interaction SEM adds transformed products of pairs of existing parents (up to three pairs per node).
Within each SEM, both variants are run on the same datasets and share the same FM checkpoint.
In both SEMs, KCI screening gives higher mean skeleton F1, strict and relaxed directed F1, and lower mean CPDAG error than FisherZ screening, with the largest skeleton-level gain on the interaction SEM (\(0.938\) versus \(0.838\)).
This table compares two variants of our own method; it contains no matched PC-stable baselines, so it does not by itself show an advantage over PC-stable on these SEMs.
Runtime is not reported: the FM checkpoint was created during the FisherZ-screened run, so its time includes FM training, whereas the KCI-screened run reused that checkpoint, and the two times are therefore not comparable.

\begin{table*}[t]
\centering
\caption{
Robustness of GPC-FMCIT under alternative
nonlinear SEMs (\(n=500\), \(p=100\),
\(\texttt{exp\_edges}=150\), \(\mathrm{df}=3\),
\(\alpha_{\mathrm{scr}}=\alpha_{\mathrm{fm}}=0.10\)).
Mean \(\pm\) sample standard deviation over five matched seeds.
Bold: better mean within each SEM.
}
\label{tab:dgp_robustness}
\small
\setlength{\tabcolsep}{4pt}
\begin{adjustbox}{max width=\textwidth}
\begin{tabular}{llcccc}
\toprule
DGP & Method
& Skel-F1 $\uparrow$
& Dir-F1 $\uparrow$
& Relaxed Dir-F1 $\uparrow$
& CPDAG Error $\downarrow$ \\
\midrule
\multirow{2}{*}{Post-nonlinear}
& GPC-FMCIT (FisherZ screen)
& $0.813 \pm 0.045$ & $0.364 \pm 0.047$ & $0.370 \pm 0.046$ & $116.6 \pm 8.6$ \\
& GPC-FMCIT (KCI screen)
& $\mathbf{0.888 \pm 0.022}$ & $\mathbf{0.400 \pm 0.039}$ & $\mathbf{0.404 \pm 0.037}$ & $\mathbf{102.4 \pm 7.1}$ \\
\midrule
\multirow{2}{*}{Interaction}
& GPC-FMCIT (FisherZ screen)
& $0.838 \pm 0.034$ & $0.364 \pm 0.034$ & $0.367 \pm 0.030$ & $124.0 \pm 8.1$ \\
& GPC-FMCIT (KCI screen)
& $\mathbf{0.938 \pm 0.013}$ & $\mathbf{0.418 \pm 0.034}$ & $\mathbf{0.420 \pm 0.036}$ & $\mathbf{98.6 \pm 5.2}$ \\
\bottomrule
\end{tabular}
\end{adjustbox}
\end{table*}

\subsection{Budget and pool-size sensitivity}
\label{app:budget_pool_sensitivity}

We study the two main search-control hyperparameters in GPC-FMCIT: the per-edge budget \(M\) and the candidate-pool size \(K\).
The budget \(M\) controls how many candidate conditioning sets are tested for each edge at each conditioning level, whereas \(K\) controls how many locally guided candidate variables are retained for constructing those conditioning sets.
Table~\ref{tab:budget_pool_ablation} reports the resulting accuracy under the FisherZ-screened stability-guided GPC-FMCIT pipeline, run on the H100 server. Runtime is not reported because each configuration and seed trained its own FM checkpoint inside the timed run.

\begin{table}[t]
\centering
\caption{
Budget and pool-size sensitivity of FisherZ-screened
Full-Joint GPC-FMCIT ($n=500$, $p=50$,
$\texttt{exp\_edges}=75$, $\mathrm{df}=3$,
$\alpha_{\mathrm{scr}}=\alpha_{\mathrm{fm}}=0.10$).
Mean $\pm$ sample standard deviation over five matched seeds.
The budget sweep fixes $K=4$ and the pool-size sweep fixes $M=2$, so the $M=2$ and $K=4$ rows coincide.
Bold: best mean within each sweep.
}
\label{tab:budget_pool_ablation}
\small
\setlength{\tabcolsep}{4pt}
\begin{adjustbox}{max width=\linewidth}
\begin{tabular}{lccc}
\toprule
Setting
& Skel-F1 $\uparrow$
& Dir-F1 $\uparrow$
& CPDAG Error $\downarrow$ \\
\midrule

$M=1$
& $0.711 \pm 0.047$
& $\mathbf{0.307 \pm 0.063}$
& $\mathbf{71.2 \pm 5.3}$ \\

$M=2$
& $0.708 \pm 0.050$
& $0.299 \pm 0.051$
& $71.4 \pm 4.3$ \\

$M=4$
& $0.710 \pm 0.048$
& $0.287 \pm 0.048$
& $\mathbf{71.2 \pm 3.9}$ \\

$M=8$
& $\mathbf{0.712 \pm 0.049}$
& $0.276 \pm 0.022$
& $71.4 \pm 2.7$ \\

\midrule

$K=1$
& $0.707 \pm 0.043$
& $\mathbf{0.306 \pm 0.040}$
& $72.4 \pm 4.0$ \\

$K=2$
& $0.710 \pm 0.047$
& $0.288 \pm 0.046$
& $72.6 \pm 4.9$ \\

$K=3$
& $0.709 \pm 0.049$
& $0.284 \pm 0.075$
& $72.4 \pm 5.9$ \\

$K=4$
& $0.708 \pm 0.050$
& $0.299 \pm 0.051$
& $71.4 \pm 4.3$ \\

$K=8$
& $\mathbf{0.716 \pm 0.045}$
& $0.294 \pm 0.062$
& $\mathbf{70.2 \pm 4.9}$ \\

$K=12$
& $0.713 \pm 0.049$
& $0.272 \pm 0.052$
& $71.6 \pm 4.9$ \\

$K=16$
& $0.706 \pm 0.052$
& $0.285 \pm 0.043$
& $71.0 \pm 4.3$ \\

\bottomrule
\end{tabular}
\end{adjustbox}
\end{table}

Table~\ref{tab:budget_pool_ablation} shows that, within the tested ranges, GPC-FMCIT is only moderately sensitive to both search-control parameters, and the differences are small and not consistently monotonic.
Across \(M\in\{1,2,4,8\}\), mean skeleton F1 stays between \(0.708\) and \(0.712\) and mean CPDAG error between \(71.2\) and \(71.4\); mean strict directed F1 decreases from \(0.307\) at \(M=1\) to \(0.276\) at \(M=8\).
Across \(K\in\{1,\ldots,16\}\), mean skeleton F1 stays between \(0.706\) and \(0.716\) and mean CPDAG error between \(70.2\) and \(72.6\), without a monotonic trend; \(K=8\) has the best mean skeleton F1 and CPDAG error, and \(K=1\) the best mean directed F1.
All of these differences are well within one standard deviation across five seeds.
We therefore use \(M=2\) and \(K=4\) as a moderate default rather than as a setting that is optimal for every metric.

The same \(M=2,K=4\) setting also appears as the full method in Table~\ref{tab:component_ablation}, with a different mean strict directed F1 (\(0.299\) here versus \(0.345\) there). These are not controlled replications: the experiments use matching five seed-level datasets, but different FM checkpoints and graph-learning implementations; the component study also uses 15 statistical worker processes on an RTX 6000 Ada, whereas the sensitivity sweep was run on an H100. Different hardware alone does not explain an accuracy difference. The individual effects of checkpoint training, sampling and implementation have not been isolated. Accordingly, the tables support only within-series comparisons; we do not attribute their discrepancy to any single factor.

\subsection{Additional real-data analyses on hESC}
\label{app:hesc_sensitivity}

This subsection gives the implementation details of the hESC experiments in Section~\ref{sec:real} and a sensitivity analysis for the FMCIT refinement level.
All hESC results use the full-joint B39 configuration: one joint FM model reused across queries, 39 CRT replicates per FMCIT query, 15 sampling steps, and unbounded conditioning depth (\(\texttt{mcs}=-1\)) for every method.

\paragraph{hESC data source and matrix format.}
We use the processed hESC single-cell gene-expression benchmark from BEELINE \citep{pratapa2020beeline}. 
The expression matrix is provided in CSV format, where the first column contains gene names and the remaining columns contain cell-wise expression values. 
Before our additional preprocessing, the processed matrix contains 17,735 genes and 758 cells.

\paragraph{Default gene selection.}
Starting from the processed hESC matrix, we first apply the transformation \(x \mapsto \log_{10}(\max(x,0)+10^{-2})\). 
We then remove duplicate gene names by keeping the first occurrence, discard genes with zero empirical variance, and remove genes whose nonzero proportion across cells is below \(0.10\). 
After these filtering steps, 13,059 genes remain. 
For the default hESC CPDAG experiments, we rank the remaining genes by empirical variance and retain the top 100 genes. 
All methods use this 100-feature matrix, in both the main CPDAG comparison and the \(\alpha_{\mathrm{fm}}\) sensitivity experiment; it contains 93 genes and 7 ERCC spike-in controls, which are retained by the variance-based selection.


\paragraph{PC / FisherZ CPDAG baseline.}
The real-data CPDAG baseline is obtained by first running a stable-PC style skeleton learning stage with Fisher's \(Z\) test, followed by the same downstream CPDAG orientation pipeline used in our implementation.
The FisherZ significance level is fixed at \(\alpha = 0.02\).
The maximum conditioning-set size is unbounded (\(\texttt{mcs}=-1\)).
After skeleton recovery, we apply collider orientation and Meek propagation rules to produce the final CPDAG.
Its final graph passes the completed-PDAG checks of Appendix~\ref{app:cpdag_validity}.

\paragraph{Guided GPC-FMCIT CPDAG learner.}
For all GPC-FMCIT-based hESC results, we use the same overall pipeline:
(i) a FisherZ screening stage,
(ii) a guided FMCIT-based refinement stage,
(iii) CPC-style conservative collider orientation, and
(iv) Meek propagation, followed by the completion check of Appendix~\ref{app:cpdag_validity}.
Unless otherwise stated, the screening significance level is fixed at \(\alpha_{\mathrm{scr}}=0.10\) and the refinement significance level at \(\alpha_{\mathrm{fm}}=0.05\); the conditioning-set size is unbounded (\(\texttt{mcs}=-1\)) in both screening and refinement.
All final GPC-FMCIT graphs are valid completed PDAGs.

\paragraph{Flow matching and CRT settings.}
All FMCIT-based results in this real-data subsection use the same FM configuration unless explicitly varied.
The train/test split ratio inside FMCIT is set to \(0.5/0.5\).
The joint FM model is parameterized by an MLP with hidden width \(128\) and \(2\) hidden layers.
We train it for \(30\) epochs using Adam with learning rate \(10^{-3}\) and batch size \(2048\).
The conditional sampler uses \(15\) FM sampling steps.
The data are standardized internally with numerical stabilization parameters \(\texttt{eps\_std}=10^{-6}\) and \(\texttt{clip\_value}=10.0\).
For the CRT stage, we use \(\texttt{crt\_repeat}=39\).
We use RDC as the dependence statistic at conditioning level \(0\) and a correlation-based statistic at higher conditioning levels.
The same joint FM model is reused across CI queries through lazy training and cached loading whenever possible. Throughout the paper, ``PC-stable (KCI)'' denotes the baseline that uses KCI throughout the PC skeleton search, whereas ``GPC-FMCIT (KCI screen)'' denotes our variant that uses KCI only to construct the initial screening graph before FMCIT-based refinement and orientation.

\paragraph{Guidance settings.}
For the triangle-guided variant, we set \(\texttt{tri\_T}=2\), \(\texttt{pool\_k}=12\), and \(\texttt{max\_sets\_per\_edge}=2\).
For the repaired stability-guided variant, we use \(\texttt{stab\_alphas}=(0.08,0.10,0.12)\) together with \((\texttt{stab\_low},\texttt{stab\_high})=(0.20,0.80)\), the same pool size \(\texttt{pool\_k}=12\), and the same per-edge budget \(\texttt{max\_sets\_per\_edge}=2\).
In both guided variants, candidate conditioning variables are derived from the screening graph and the same CPC-style orientation stage is used after refinement.

\paragraph{Guidance comparison protocol.}
In the main hESC CPDAG comparison, we compare three methods on the same default hESC expression matrix:
(i) PC / FisherZ,
(ii) GPC-FMCIT (stability guide), and
(iii) GPC-FMCIT (triangle guide).
For fairness, the two GPC-FMCIT variants share the same screening level, FM parameters, and orientation pipeline; only the guidance rule is changed.
The purpose of this comparison is to isolate the effect of the guidance mechanism on the compactness and computational behavior of the final CPDAG.

\paragraph{Reported metrics for the hESC CPDAG study.}
Since no exact causal ground truth is available, we report the numbers of skeleton, directed and undirected edges of the final CPDAG and the recorded graph-learning runtime, rather than ground-truth recovery scores.

\begin{figure}[t]
    \centering
    \includegraphics[width=0.72\linewidth]{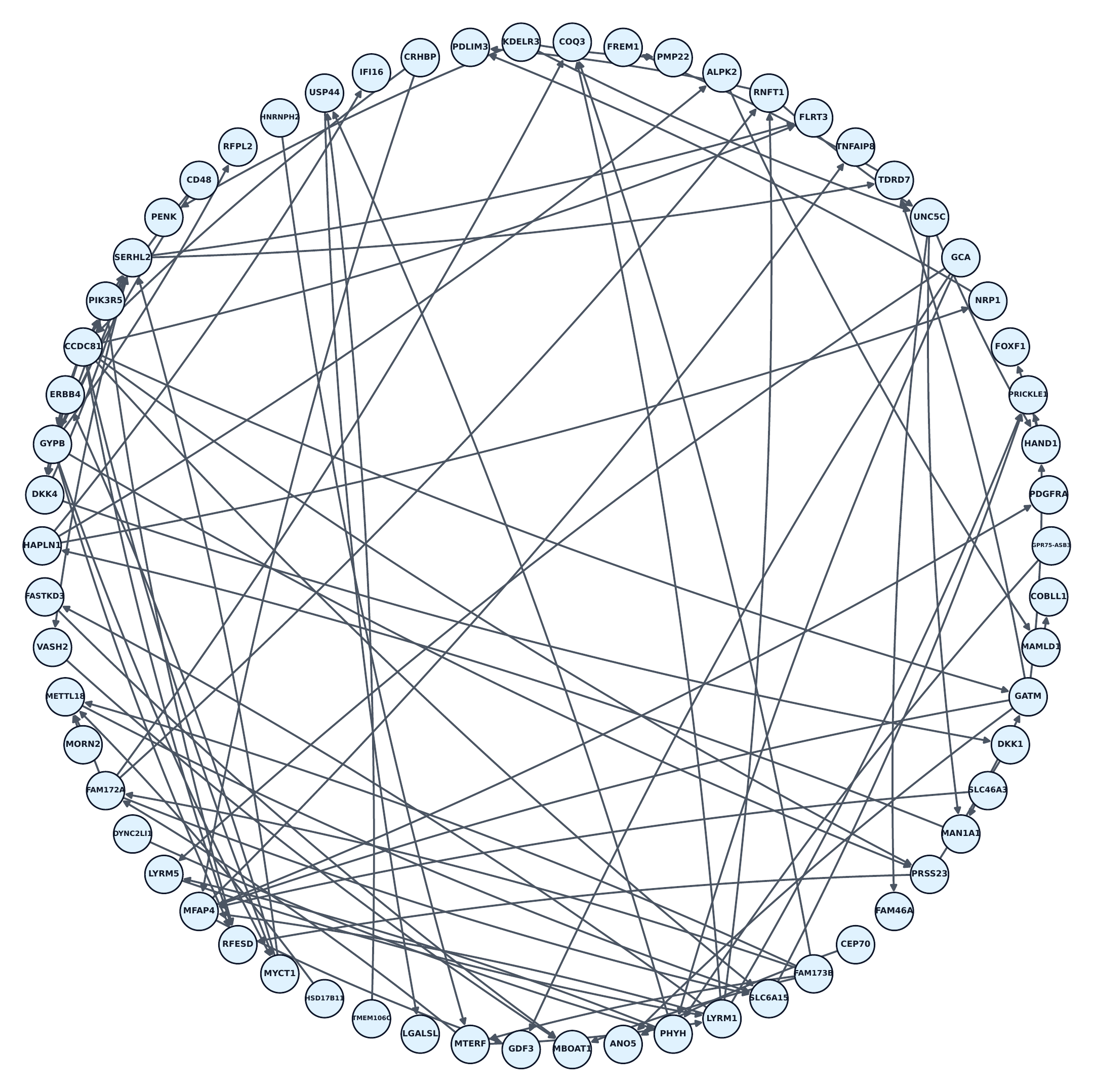}
    \caption{
    Directed edges of the hESC CPDAG learned by triangle-guided GPC-FMCIT (full-joint B39, \(\texttt{mcs}=-1\), \(\alpha_{\mathrm{scr}}=0.10\), \(\alpha_{\mathrm{fm}}=0.05\)). Of the 124 directed edges, the 30 that involve ERCC spike-in controls are omitted, as are the six undirected edges.
    }
    \label{fig:hesc_alpha005_directed}
\end{figure}

\paragraph{Sensitivity to the FMCIT significance level.}
We vary the refinement level \(\alpha_{\mathrm{fm}}\) for triangle-guided GPC-FMCIT while keeping all other settings fixed.
Table~\ref{tab:hesc_alpha_sensitivity} shows that the learned graph becomes denser as \(\alpha_{\mathrm{fm}}\) increases, from 92 to 162 skeleton edges, and that the runtime increases with it.
This is expected, because a larger \(\alpha_{\mathrm{fm}}\) rejects more conditional independence hypotheses and therefore retains more edges, so \(\alpha_{\mathrm{fm}}\) is the main control of graph density in this pipeline.
Figure~\ref{fig:hesc_alpha005_directed} shows the directed edges of the triangle-guided graph at \(\alpha_{\mathrm{fm}}=0.05\).

\begin{table}[t]
\centering
\caption{
Sensitivity to the FMCIT significance level under triangle-guided GPC-FMCIT on hESC (full-joint B39, \(\texttt{mcs}=-1\), \(\alpha_{\mathrm{scr}}=0.10\)).
}
\label{tab:hesc_alpha_sensitivity}
\small
\setlength{\tabcolsep}{5pt}
\begin{adjustbox}{max width=\linewidth}
\begin{tabular}{ccccc}
\toprule
\(\alpha_{\mathrm{fm}}\) & Skeleton Edges & Directed & Undirected & Runtime (s) \\
\midrule
0.03 & 92 & 86 & 6 & 1614.48 \\
0.05 & 130 & 124 & 6 & 4488.56 \\
0.08 & 162 & 157 & 5 & 7650.29 \\
\bottomrule
\end{tabular}
\end{adjustbox}
\end{table}

\paragraph{Overall conclusion on hESC.}
Both guided variants learn much sparser CPDAGs than PC/FisherZ, which uses a different significance level, so this comparison is descriptive; triangle guidance does so at lower cost than stability guidance, and the density of the learned graph is governed mainly by \(\alpha_{\mathrm{fm}}\).
Without a verified causal ground truth, these results describe graph compactness and computational cost and do not establish biological accuracy.

\subsection{Computing environment and runtime measurement}
\label{app:compute_environment}

The main server has two Intel Xeon Gold 6526Y CPUs, 1.0 TiB RAM, and one NVIDIA H100 NVL GPU with 95,830 MiB memory, and uses Python 3.10.12, PyTorch 2.6.0, NumPy 1.26.4, SciPy 1.11.0, scikit-learn 1.6.1, and pandas 2.2.3.
The component ablation was run on a different machine with an NVIDIA RTX 6000 Ada Generation GPU (Appendix~\ref{app:ablation_protocol}).
Reported runtimes are wall-clock times.
Graph-level runtime measurements use their implementation-specific stopwatch boundaries; standalone CI-test runtimes are measured per query. For the main GPC-FMCIT table, the stopwatch surrounds the graph-learning call (screening, guidance, refinement, orientation, and CPDAG completion). The run configuration permits model training upon a missing checkpoint, and archived timestamps show that the first FisherZ-screened run created an FM checkpoint within the run for seeds 2000--9000, whereas seeds 0 and 1000 already had checkpoints. Later runs at the same seed reuse that checkpoint; hence main GPC runtimes are not uniformly training-excluded. For PC-stable (FisherZ and KCI) baselines, the reported runtime is the original CPU graph-learning duration plus a separately measured post-hoc CPDAG repair duration, not a single timed end-to-end replay. PC-stable runs were restricted to CPU cores, whereas GPC used the H100; absolute times across these methods are descriptive rather than a controlled comparison.
Table~\ref{tab:runtime_protocol} lists the boundary and hardware of each table.
The wall-clock numbers do not include a uniform accounting of FM pretraining, cross-seed hardware contention, or parallel worker time, and should not be used as an estimate of total hardware-hours.

\begin{table}[t]
\centering
\caption{Runtime measurement boundary and hardware of each table.}
\label{tab:runtime_protocol}
\small
\setlength{\tabcolsep}{3pt}
\begin{adjustbox}{max width=\linewidth}
\begin{tabular}{lll}
\toprule
Table & Runtime boundary & Hardware \\
\midrule
Table~\ref{tab:type1_n_calibration_main}, \ref{tab:cit_runtime} & per CI test & main server \\
Table~\ref{tab:sim_pc_main} & graph invocation (some GPC first runs also trained FM); PC time + post-hoc repair & H100 for GPC; CPU for PC-stable \\
Table~\ref{tab:component_ablation} & graph learning incl.\ completion; training excluded & RTX 6000 Ada \\
Table~\ref{tab:kci_screen_n_sweep} & not reported (\(n=200,1000\) include checkpoint creation) & main server \\
Table~\ref{tab:dgp_robustness} & not reported (FisherZ runs include checkpoint creation) & main server \\
Table~\ref{tab:budget_pool_ablation} & not reported (checkpoints created within runs) & H100 \\
Tables~\ref{tab:hesc_cpdag_main}, \ref{tab:hesc_alpha_sensitivity} & graph learning as recorded by the hESC runner & H100 for GPC; CPU for PC \\
\bottomrule
\end{tabular}
\end{adjustbox}
\end{table}

\section{Theoretical Results}
\label{app:fmcit_theory}

In this section, we provide proofs of the failure of vanilla RePaint to converge to the true conditional distribution, the convergence of our proposed Langevin-RePaint algorithm, and the type-I error control of FMCIT.

\subsection{Notation}
\label{sec_notation}
\paragraph{Distributions, dimensions, negative log-likelihood, and the conditional score.} For clarity and notational simplicity, we introduce
a slightly different notation for the theoretical
analysis in this section. Let $Y=X_{S^c}\in\mathbb R^d$
denote the missing variables to be imputed and
$Z=X_S\in\mathbb R^q$ denote the observed variables,
where the total dimension $p=d+q$. The distribution of $Z$ is defined as $P_Z$. Their corresponding variables along the FM interpolation path $X(t)$ are denoted by $Y(t)$ and $Z(t)$, respectively. We write the conditional density of
$Y$ given $Z=z$ as
    $\pi_z(y):=p_{X_{S^c}\mid X_S}(y\mid z)$. 
We define the negative log-likelihood of $\pi_z(y)$ as $\mathbf{NLL}_z(y) := -\log \pi_z(y)$, and the score of $\pi_z(y)$ as $b_z(y):= \nabla_y \log \pi_z(y)$.

\paragraph{Reordering of the coordinates and the true velocity.} Since this is merely a relabeling and reordering of
the coordinates of $X$, we also write the learned
velocity field $\widehat v_t(x)$ as
$\widehat v_t(y,z)$, with the input and output
coordinates reordered consistently.  Similarly, the true velocity field $v_t^*(x) := \mathbb E [X(1) - X(0) | X(t) =x ]$ in
(\ref{eq_thm_tv}) can also be written as $v_t^*(y,z)$.

\paragraph{The notations associated with sampling steps and time.} Throughout the analysis, the sampling grid is uniform:
\begin{equation}
\label{eq_t_ell}
t_\ell=\frac{\ell}{L},\qquad \ell=0,\ldots,L.    
\end{equation}
For a terminal-window width $\delta_L$, define
\begin{equation}
\label{eq_lr_grid_notation}
\ell_0
:=
\left\lceil L(1-\delta_L)\right\rceil,
\qquad
M_L:=L-\ell_0,
\qquad
n_L:=L-1-\ell_0,
\qquad
\eta_\ell:=\frac{(1-t_\ell)^2}{t_\ell^2},
\end{equation}
where $\eta_\ell$ is used only for $\ell\geq1$.
Thus, $M_L$ is the number of half-score predictions, whereas $n_L$
is the number of half-score predictions followed by Gaussian
re-noising. The implemented choice is $\delta_L=L^{-1/4}$.

\paragraph{Transformation between velocity, score, denoiser, and half-score correction.} Recall the Langevin-RePaint update in
(\ref{eq_lr_update}). We call $t\in[0, 1-\delta_L)$ the warm-up stage and $t \in [1-\delta_L, 1)$ the half-score correction stage, respectively. For $t_\ell<1-\delta_L$,
the terminal prediction is given by a denoiser
\cite{li_2026_jit}.
We define the learned and true denoisers as
\begin{equation}
\begin{aligned}
D_t(x) &= x+(1-t)\widehat v_t(x), \quad
D_t^*(x) = x+(1-t)v_t^*(x).
\end{aligned}
\end{equation}
For $t_\ell\geq 1-\delta_L$, we instead employ
the half-score correction. Accordingly, we define
\begin{equation}
\begin{aligned}
G_t(x) &=
\frac{1+t}{2t}x
+\frac{1-t}{2}\widehat v_t(x), \quad
G_t^*(x) =
\frac{1+t}{2t}x
+\frac{1-t}{2}v_t^*(x).
\end{aligned}
\end{equation}

To explain the term "half-score correction",
recall that, under the linear FM interpolation,
the true velocity field and score function satisfy
\begin{equation}
s_t^*(x)
:=\nabla \log p_t(x)
=\frac{t v_t^*(x)-x}{1-t},
\end{equation}
where $p_t$ denotes the marginal density of $X(t)$ \cite{lipman_2023_fm,li_2026_jit}.
Similarly, the estimated score is given by
$\widehat s_t(x)=(t\widehat v_t(x)-x)/(1-t)$.
Consequently, for $t\in(0,1)$, we can rewrite
\begin{equation}
\label{eq_denoiser_to_score}
\begin{aligned}
D_t(x)
&=\frac{x}{t}
+\frac{(1-t)^2}{t}\widehat s_t(x), \quad
G_t(x)
&=\frac{x}{t}
+\frac{(1-t)^2}{2t}\widehat s_t(x).
\end{aligned}
\end{equation}
Thus, $G_t$ retains the same leading term as
$D_t$ while reducing the score contribution
by half. The same relabeling of the input coordinates from $x$ to $(y,z)$ also applies to $D_t$, $D_t^*$, $G_t$, $G_t^*$, $\widehat s_t$, $s^*_t$, and $p_t$.

\paragraph{The dataset.} Let $\mathcal D =\{X^{(1)},\ldots,X^{(N)}\}$ consist of $N$ independent
observations from the joint data distribution with density $p_1$.
The learned velocity field $\widehat v_t$ is constructed from
$\mathcal D$.  The observations used for  CRT are independent of $\mathcal D$.
All Gaussian random noise generated during sampling are mutually
independent and independent of the training data and the observation.

\subsection{Assumptions for theoretical analysis}
\label{sec_assumptions_langevin_repaint}

In this section, we list the assumptions of our theoretical analysis.

\begin{assumption}[Joint density regularity]
\label{ass_lr_joint_reg}
The joint data density $p_1$ is strictly positive and belongs to
$C^2(\mathbb R^p)$, where $p=d+q$. Moreover,
\begin{equation}
\label{eq_2_moment_finite}
\mathbb E_{X\sim p_1}\|X\|^2<\infty.    
\end{equation}

The endpoint score $s_1^*(x):=\nabla\log p_1(x)$ is globally $K$-Lipschitz for some $K<\infty$:
\begin{equation}
\label{eq_score_1_lipschitz}
\|s_1^*(x)-s_1^*(x')\|
\leq K\|x-x'\|,
\qquad x,x'\in\mathbb R^p.    
\end{equation}

\end{assumption}

\begin{assumption}[Uniform conditional strong log-concavity]
\label{ass_lr_cond_reg}
For $P_Z$-almost every $z$, the conditional density $\pi_z$ is
strictly positive and twice continuously differentiable.
There exists $m>0$, independent of $z$, such that
\begin{equation}
\label{eq_lr_cond_hessian}
mI_d
\preceq
\nabla_y^2\mathbf{NLL}_z(y)
\preceq
KI_d,
\qquad y\in\mathbb R^d.
\end{equation}
\end{assumption}

The upper bound in \eqref{eq_lr_cond_hessian} also follows from
Assumption~\ref{ass_lr_joint_reg}, since
$b_z(y)
=
[s_1^*(y,z)]_Y$. 
We state it explicitly because both the lower and upper Hessian
bounds are used in the proof.

In particular, the conditional target satisfies the log-Sobolev
inequality \cite{Brigati_2026_log_sobolev}
\begin{equation}
\label{eq_lr_lsi}
D_{\mathrm{KL}}(\rho\|\pi_z)
\leq
\frac{1}{2m}I(\rho\|\pi_z),
\qquad
I(\rho\|\pi_z)
:=
\int_{\mathbb R^d}
\rho(y)
\left\|
\nabla_y\log\frac{\rho(y)}{\pi_z(y)}
\right\|^2\,dy,
\end{equation}
for probability densities for which the right-hand side is defined.
This is the standard log-Sobolev consequence of strong
log-concavity \cite{Vempala_2019_log_sobolev}.

\begin{assumption}[Terminal Score Consistency]
\label{ass_terminal_score}
There exists $C_{\mathrm{TS}}<\infty$ such that, for all
$t_\ell\in[1/2,1)$ and
$(y,z)\in\mathbb R^d\times\mathbb R^q$,
\begin{equation}
\label{eq_lr_terminal_consistency_weak}
\begin{aligned}
&\mathbb E_{e_\ell^Z\sim\mathcal N(0,I_q)}
\left[
\left\|
t_\ell
\left[
\nabla\log p_{t_\ell}
\left(
t_\ell y,\,
t_\ell z+(1-t_\ell)e_\ell^Z
\right)
\right]_Y
-b_z(y)
\right\|^2
\right]\\
&\qquad\leq
C_{\mathrm{TS}}(1-t_\ell)
\left(1+\|y\|^2+\|z\|^2\right).
\end{aligned}
\end{equation}
\end{assumption}

Assumption~\ref{ass_terminal_score} gives the mean-square drift
consistency used in the SDE discussion.
The quantitative convergence rate proved below uses a stronger
bound with $(1-t_\ell)^2$ on the right-hand side.
Importantly, this stronger bound is established in
Lemma~\ref{lem_lr_terminal_refinement} from
Assumption~\ref{ass_lr_joint_reg}; it is not inferred from
Assumption~\ref{ass_terminal_score} alone.

\begin{assumption}[Warm-up moment control]
\label{ass_lr_warmup}
Let $A_{L}^z\in\mathbb R^d$ denote the $Y$-coordinate
terminal prediction produced by the learned warm-up stage at
step $\ell_0-1$, before re-noising to time $t_{\ell_0}$.
There exists $B_0<\infty$ such that
\begin{equation}
\label{eq_lr_warmup_moment}
\mathbb E_{\mathcal D}
\mathbb E_{Z\sim P_Z}
\mathbb E_{\mathrm{warm}\mid\mathcal D,Z}
\|A_{L}^{Z}\|^2
\leq B_0
\end{equation}
uniformly over the considered pairs $(N, L)$.
\end{assumption}

This assumption does not require the warm-up distribution to be
close to $\pi_z$. It only controls the second moment of the
initialization supplied to the half-score correction stage.

\begin{assumption}[Regularity of the final prediction]
\label{ass_lr_final_reg}
The learned velocity field is measurable and finite-valued.
At $t_*=t_{L-1}=1-1/L$, the map
$x\mapsto\widehat v_{t_*}(x)$ is continuously differentiable for
almost every training realization. There exist constants
$B,B_v<\infty$ such that
\begin{equation}
\label{eq_lr_final_reg}
\sup_{x\in\mathbb R^p}
\|\nabla_x\widehat v_{t_*}(x)\|_{\mathrm{op}}
\leq B
\quad\text{almost surely},
\qquad
\mathbb E_{\mathcal D}
\|\widehat v_{t_*}(0)\|^2
\leq B_v.
\end{equation}
The constants are independent of $N$ and $L$.
\end{assumption}

Assumption~\ref{ass_lr_final_reg} is used only to control the last,
non-re-noised output map. It is not a uniform approximation
assumption on $\widehat v_t-v_t^*$.

\begin{assumption}[Oracle-trajectory velocity risk]
\label{ass_lr_trajectory_risk}
Let $U_{t_\ell}^{*,z}$ be the oracle half-score chain defined in
\eqref{eq_lr_oracle_chain_full}. It starts from the same learned
warm-up distribution as the actual sampler, but uses $v_t^*$
instead of $\widehat v_t$ during the terminal correction stage.
Define its input to the joint velocity field by
\begin{equation}
\label{eq_lr_oracle_input}
\widetilde X_{t_\ell}^{*,z}
:=
\left(
t_\ell U_{t_\ell}^{*,z},\,
t_\ell z+(1-t_\ell)e_\ell^Z
\right).
\end{equation}
The cumulative terminal risk
\begin{equation}
\label{eq_lr_trajectory_risk}
\begin{aligned}
\mathcal R_{N,L}^{*}
:=
\sum_{\ell=\ell_0}^{L-2}
\mathbb E_{\mathcal D}
\mathbb E_{Z\sim P_Z}
\mathbb E_{\widetilde X_{t_\ell}^{*,Z} \mid\mathcal D,Z}
\left[
\left\|
\widehat v_{t_\ell}(\widetilde X_{t_\ell}^{*,Z})
-
v_{t_\ell}^{*}(\widetilde X_{t_\ell}^{*,Z})
\right\|^2
\right]
\end{aligned}
\end{equation}
is finite. For the asymptotic consistency statement, we additionally
require
\[
\mathcal R_{N,L_N}^{*}\longrightarrow0
\]
along the chosen sequence $L_N\to\infty$.
\end{assumption}

The expectation in \eqref{eq_lr_trajectory_risk} includes the
oracle initialization and all oracle sampling noises. 
The risk is evaluated under the oracle sampling distribution,
which generally differs from the FM training distribution.

The following two assumptions are needed for
Corollary~\ref{cor_lr_population_fm} and the FM-risk bound in
Theorem~\ref{thm_LR_tv}.
They are not needed for Theorem~\ref{thm_lr_conditional_tv},
which is stated directly in terms of the oracle-trajectory risk.

\begin{assumption}[Oracle-to-FM distribution comparison]
\label{ass_lr_density_transfer}
For fixed $\mathcal D$, let
\[
Q_{\ell,\mathcal D}^{*}
:=
\mathcal L(\widetilde X_{t_\ell}^{*,Z}\mid\mathcal D),
\]
where the law includes averaging over $Z\sim P_Z$ and the oracle
sampling randomness. 
There exists a finite deterministic constant
$\Lambda_{\mathcal D,L}$, independent of the
realized training data, such that for almost
every $\mathcal D$ and every
$\ell=\ell_0,\ldots,L-2$,
\begin{equation}
\label{eq_lr_density_transfer}
Q_{\ell,\mathcal D}^{*}
\ll p_{t_\ell}(x)\,dx,
\qquad
\frac{dQ_{\ell,\mathcal D}^{*}}
     {d[p_{t_\ell}(x)\,dx]}(x)
\leq \Lambda_{\mathcal D,L} < \infty
\quad\text{for almost every }x.
\end{equation}
Besides, there exist a constant $\overline{\Lambda} < \infty$ such that $\sup_{\mathcal D, L} \Lambda_{\mathcal D,L} \leq \overline{\Lambda}$. 
\end{assumption}

\begin{assumption}[Terminal-time risk comparison]
\label{ass_lr_time_transfer}
Define
\begin{equation}
\label{eq_lr_population_risk}
\begin{aligned}
g_{N}(t)
&:=
\mathbb E_{\mathcal D}
\mathbb E_{X(t)\sim p_t}
\left[
\|\widehat v_t(X(t))-v_t^*(X(t))\|^2
\right],\\
\mathcal E_{N}^{\mathrm{FM}}
&:=
\int_0^1g_{N}(t)\,dt,
\end{aligned}
\end{equation}
where the FM interpolation variables in the inner expectation
are independent of $\mathcal D$.
There exists $C_{\mathrm{time}}<\infty$, independent of $N,L$,
such that
\begin{equation}
\label{eq_lr_time_transfer}
\frac{1}{n_L}
\sum_{\ell=\ell_0}^{L-2}g_{N}(t_\ell)
\leq
C_{\mathrm{time}}\mathcal E_{N}^{\mathrm{FM}}.
\end{equation}
\end{assumption}

Assumptions~\ref{ass_lr_density_transfer} and
\ref{ass_lr_time_transfer} are not consequences of conditional
strong log-concavity. In particular, convergence of the
time-integrated FM risk alone does not control its values at a
deterministic terminal grid.

\subsection{Repaint under the perspective of a stochastic differential equation (SDE)}
\label{sec_repaint_fails}
In this subsection, we first derive a stochastic differential equation (SDE) representation of the vanilla RePaint update in (\ref{eq_lr_update}), assuming access to the exact velocity field $v_t^*$. We then show that the resulting SDE converges to a biased distribution with density $\pi_z^2(y)/\int_{\mathbb R^d}\pi_z^2(u)\,du$, rather than the target conditional density $\pi_z(y)$.

According to (\ref{eq_lr_update}), for a fixed observation $X_S=Z=z$, the vanilla RePaint update with the exact velocity field $v_t^*$ can be expressed componentwise as
\begin{equation}
\label{eq_repaint_update}
\begin{aligned}
&Y(t_{\ell+1})
=t_{\ell+1}\left[D_{t_\ell}^*\big(Y(t_\ell),Z(t_\ell)\big)\right]_{Y}
+(1-t_{\ell+1})e^Y_{\ell+1}, \quad e^Y_{\ell+1} \sim N(0, I_d)\\
&Z(t_{\ell+1})
=t_{\ell+1}z
+(1-t_{\ell+1})e^Z_{\ell+1}, \quad e^Z_{\ell+1} \sim N(0, I_q), \quad \ell=0,\ldots,L-2, \\
&Y(t_{L})=[D_{t_{L-1}}^*\big(Y(t_{L-1}),Z(t_{L-1})\big)]_Y.
\end{aligned}
\end{equation}
Using the relationship between the denoiser and
the score function in (\ref{eq_denoiser_to_score}),
and dividing both sides of (\ref{eq_repaint_update})
by $t_{\ell+1}$ when $t_\ell >0$, we obtain
\begin{equation}
\label{eq_repaint_score_update}
\frac{Y(t_{\ell+1})}{t_{\ell+1}}
=
\frac{Y(t_\ell)}{t_\ell}
+\frac{(1-t_\ell)^2}{t_\ell}
\left[\nabla\log p_{t_\ell}\big(Y(t_\ell),Z(t_\ell)\big)\right]_Y
+\frac{1-t_{\ell+1}}{t_{\ell+1}}e_{\ell+1}^Y.
\end{equation}

To facilitate the SDE analysis, we introduce the
rescaled time parameter and state variable
\begin{equation}
\eta_\ell=\frac{(1-t_\ell)^2}{t_\ell^2},
\qquad
U_{t_\ell}:=\frac{Y(t_\ell)}{t_\ell}.
\end{equation}
Since
$Z(t_\ell)=t_\ell z+(1-t_\ell)e_\ell^Z$,
we further define the stochastic drift as
\begin{equation}
\label{eq_repaint_drift}
b_{t_\ell}^z(y;e_\ell^Z)
:=
t_\ell
\left[
\nabla\log p_{t_\ell}
\left(
t_\ell y,\,
t_\ell z+(1-t_\ell)e_\ell^Z
\right)
\right]_Y.
\end{equation}
Consequently, (\ref{eq_repaint_score_update})
can be rewritten as
\begin{equation}
\label{eq_repaint_rescaled}
U_{t_{\ell+1}}
=
U_{t_\ell}
+\eta_\ell b_{t_\ell}^z(U_{t_\ell};e_\ell^Z)
+\sqrt{\eta_{\ell+1}}\,e_{\ell+1}^Y, \quad \ell=1,\ldots,L-2.
\end{equation}
Under Assumption~\ref{ass_terminal_score}, the stochastic
drift $b_{t_\ell}^z(y;e_\ell^Z)$ converges to
$b_z(y)$ in mean square as $t_\ell\to 1$.
Consequently, near the terminal time,
the recursion in (\ref{eq_repaint_rescaled})
can be approximated by
\begin{equation}
\label{eq_repaint_approx}
U_{t_{\ell+1}}
\approx
U_{t_\ell}
+\eta_\ell b_z(U_{t_\ell})
+\sqrt{\eta_{\ell+1}}\,e_{\ell+1}^Y.
\end{equation}

To derive the continuous-time dynamics corresponding
to (\ref{eq_repaint_approx}), we assume a uniform
sampling grid $t_\ell=\ell/L$ in (\ref{eq_t_ell}) with step size
$h=1/L$. Define the time-dependent coefficient
\begin{equation}
a_L(t):=L\frac{(1-t)^2}{t^2}.
\end{equation}
Then $\eta_\ell=a_L(t_\ell)h$.
Since $e_{\ell+1}^Y\sim\mathcal N(0,I_d)$,
we can represent the Gaussian perturbation as
$\Delta B_\ell=\sqrt{h}\,e_{\ell+1}^Y$,
where $\Delta B_\ell$ denotes a Brownian increment.
Consequently, (\ref{eq_repaint_approx}) becomes
\begin{equation}
\begin{aligned}
U_{t_{\ell+1}}-U_{t_\ell}
\approx{}&
a_L(t_\ell)b_z(U_{t_\ell})h +\sqrt{a_L(t_{\ell+1})}\,\Delta B_\ell.
\end{aligned}
\end{equation}
For $t_\ell\geq 1-\delta_L$, the difference
between $\sqrt{\eta_{\ell+1}}$ and
$\sqrt{\eta_\ell}$ is of order $O(L^{-1})$.
Thus, the accumulated mean-square discrepancy
between the two noise discretizations vanishes
as $L\to\infty$.

This motivates the following time-inhomogeneous
Langevin SDE as a continuous-time approximation
of the vanilla RePaint dynamics:
\begin{equation}
\label{eq_repaint_sde}
\boxed{
dU_t
=
a_L(t)\nabla_y\log\pi_z(U_t)\,dt
+
\sqrt{a_L(t)}\,dB_t,
}
\end{equation}
where $B_t$ is a standard $d$-dimensional
Brownian motion. Under the regularity and ergodicity conditions
of \cite{roberts_tweedie_1996}, this diffusion
converges to its invariant distribution $\pi_z^2(y)/\int_{\mathbb R^d}\pi_z^2(u)\,du$. 
In practice, convergence to 
$\pi_z^2(y)/\int_{\mathbb R^d}\pi_z^2(u)\,du$
does not necessarily produce unrealistic samples.
Instead, squaring the density preserves its support
while assigning relatively greater probability mass
to high-density regions, resulting in overly
concentrated conditional samples.
This may partly explain the strong empirical
performance of vanilla RePaint despite its
theoretical bias.

Although vanilla RePaint does not converge to the
target conditional distribution $\pi_z(y)$,
this bias can be corrected by halving the score
term in (\ref{eq_repaint_sde}), which motivates
our \emph{half-score correction}.
Specifically, the corrected SDE becomes
\begin{equation}
\label{eq_half_score_sde}
dU_t
=
\frac{a_L(t)}{2}
\nabla_y\log\pi_z(U_t)\,dt
+
\sqrt{a_L(t)}\,dB_t.
\end{equation}
Under suitable regularity and ergodicity conditions,
this SDE converges to the desired conditional
distribution $\pi_z(y)$.
At the discrete level, the half-score correction
corresponds to replacing $b_{t_\ell}^z$ with
$\frac{1}{2}b_{t_\ell}^z$ in
(\ref{eq_repaint_rescaled}), which is equivalent
to the corrected denoiser $G_t(x)$ defined in
(\ref{eq_denoiser_to_score}).
This is precisely the terminal-stage update
of Langevin-RePaint in (\ref{eq_lr_update})
for $t_\ell\geq 1-\delta_L$.

\subsection{Theoretical analysis of Langevin-RePaint}
\label{sec_thm_langevin_repaint}

We establish convergence of the proposed Langevin-RePaint
sampler, rather than deducing convergence solely from the
invariant distribution of its effective SDE. The proof separates
three errors: the difference between the learned and oracle
terminal chains, the difference between the oracle chain and the
target conditional distribution, and the effect of the final
non-re-noised prediction.

\paragraph{Sampler states and terminal indexing.} 
We distinguish the FM interpolation or vanilla Repaint variables $Y(t),Z(t)$  from
the states generated by Langevin-RePaint.
We write
$\widetilde X_{t_\ell}=(\widetilde Y_{t_\ell},\widetilde Z_{t_\ell})$ for a
sampler input at time $t_\ell$. The sampler state need not have
density $p_{t_\ell}$.

Similar to (\ref{eq_repaint_update}), given $Z=z$, the update of Langevin-RePaint is 
\begin{equation}
\label{eq_langevin_repaint_update}
\begin{aligned}
&\widetilde Y_{t_{\ell+1}}
=
\begin{cases}
t_{\ell+1}
\left[D_{t_\ell}
\big(\widetilde Y_{t_\ell},\widetilde Z_{t_\ell}\big)\right]_Y
+(1-t_{\ell+1})e^Y_{\ell+1},
& t_\ell<1-\delta_L,\\[2mm]
t_{\ell+1}
\left[G_{t_\ell}
\big(\widetilde Y_{t_\ell},\widetilde Z_{t_\ell}\big)\right]_Y
+(1-t_{\ell+1})e^Y_{\ell+1},
& t_\ell\geq 1-\delta_L,
\end{cases}\\ 
&\widetilde Z_{t_{\ell+1}}
=t_{\ell+1}z+(1-t_{\ell+1})e^Z_{\ell+1},
\quad \ell=0,\ldots,L-2,\\[2mm]
& \widehat Y_{t_{L}}
=
[G_{t_{L-1}}(\widetilde Y_{t_{L-1}},\widetilde Z_{t_{L-1}})]_Y,
\end{aligned}
\end{equation}
where all $e^Y_{\ell+1}$ and
$e^Z_{\ell+1}$ are independently drawn from$ N(0,I_d)$ and $ N(0,I_q)$, respectively.


\paragraph{Learned and oracle terminal chains.}
For $y\in\mathbb R^d$, $z\in\mathbb R^q$, and $e\in\mathbb R^q$,
define
\begin{equation}
\label{eq_lr_joint_input_map}
x_{t_\ell}(y,z,e)
:=
\left(t_\ell y,\,
t_\ell z+(1-t_\ell)e\right).
\end{equation}
The learned half-score transition mean in the rescaled coordinates
is
\begin{equation}
\label{eq_lr_learned_mean}
m_{t_\ell}^{z}(y,e)
:=
[G_{t_\ell}(x_{t_\ell}(y,z,e))]_Y
=
\frac{1+t_\ell}{2}y
+
\frac{1-t_\ell}{2}
[\widehat v_{t_\ell}(x_{t_\ell} (y,z,e))]_Y.
\end{equation}
The corresponding oracle mean is
\begin{equation}
\label{eq_lr_oracle_mean}
m_{t_\ell}^{*,z}(y,e)
:=
[G_{t_\ell}^*(x_{t_\ell}(y,z,e))]_Y
=
\frac{1+t_\ell}{2}y
+
\frac{1-t_\ell}{2}
[v_{t_\ell}^{*}(x_{t_\ell}(y,z,e))]_Y.
\end{equation}

Conditional on $\mathcal D$ and $Z=z$, let both the learned and oracle chain start from
the same law:
\begin{equation}
\label{eq_lr_common_initialization}
U_{t_{\ell_0}}^{z}
\overset{d}{=}
U_{t_{\ell_0}}^{*,z}
\overset{d}{=}
A_{L}^{Z}+\sqrt{\eta_{\ell_0}}\,e_{\ell_0}^Y,
\qquad
e_{\ell_0}^Y\sim\mathcal N(0,I_d),
\end{equation}
where $A_{L}^{Z}$ is the warm-up stage defined in Assumption \ref{ass_lr_warmup}. For $\ell=\ell_0,\ldots,L-2$, the learned chain is
\begin{equation}
\label{eq_lr_learned_chain_full}
U_{t_{\ell+1}}^{z}
=
m_{t_{\ell}}^{z}(U_{t_{\ell}}^{z},e_\ell^Z)
+
\sqrt{\eta_{\ell+1}}\,e_{\ell+1}^Y,
\end{equation}
and the oracle chain is
\begin{equation}
\label{eq_lr_oracle_chain_full}
U_{t_{\ell+1}}^{*,z}
=
m_{t_{\ell}}^{*,z}(U_{t_{\ell}}^{*,z},e_\ell^Z)
+
\sqrt{\eta_{\ell+1}}\,e_{\ell+1}^Y.
\end{equation}
Each chain uses fresh independent Gaussian noises with
$e_\ell^Z\sim N(0,I_q)$ and
$e_{\ell+1}^Y\sim\mathcal N(0,I_d)$.
These formulas specify transition laws; no coupling between the
two chains is required. We write
\[
\mu_{t_\ell}^{z}
:=
\mathcal L(U_{t_\ell}^{z}\mid\mathcal D,Z=z),
\qquad
\mu_{t_\ell}^{*,z}
:=
\mathcal L(U_{t_\ell}^{*,z}\mid\mathcal D,Z=z).
\]
In particular, the oracle chain may depend on the training data
through the learned warm-up stage $A_{L}^{Z}$. The oracle mean also has the exact score representation
\begin{equation}
\label{eq_lr_oracle_score_mean}
m_{t_\ell}^{*,z}(y,e)
=
y+\frac{\eta_\ell}{2}b_{t_\ell}^z(y;e),
\end{equation}
where $b_{t_\ell}^z$ is defined in
\eqref{eq_repaint_drift}. Consequently,
\begin{equation}
\label{eq_lr_oracle_score_chain}
U_{t_{\ell+1}}^{*,z}
=
U_{t_{\ell}}^{*,z}
+\frac{\eta_\ell}{2}
b_{t_\ell}^z(U_{t_{\ell}}^{*,z};e_\ell^Z)
+\sqrt{\eta_{\ell+1}}\,e_{\ell+1}^Y.
\end{equation}
Let $\widehat P_{Y\mid Z=z}^{\mathcal D,L}$ denote the law of the final output
\eqref{eq_langevin_repaint_update}, conditional on
$\mathcal D$ and $Z=z$. This is the probability law induced by
the sampler, not an empirical measure formed from finitely many
generated samples.

\begin{theorem}[Conditional TV convergence of Langevin-RePaint]
\label{thm_lr_conditional_tv}
Suppose Assumptions~\ref{ass_lr_joint_reg},
\ref{ass_lr_cond_reg}, \ref{ass_lr_warmup},
\ref{ass_lr_final_reg}, and
\ref{ass_lr_trajectory_risk} hold.
Define
\begin{equation}
\label{eq_lr_effective_time}
T_L
:=
\sum_{\ell=\ell_0}^{L-2}\eta_{\ell+1}.
\end{equation}
For sufficiently small $\delta_L$, sufficiently large $L$,
and $L\delta_L\geq4$, there exist constants $C,c>0$,
independent of $N,L$, such that
\begin{equation}
\label{eq_lr_main_tv_bound}
\begin{aligned}
&\mathbb E_{\mathcal D}
\mathbb E_{Z\sim P_Z}
d_{\mathrm{TV}}
\left(
P_{Y\mid Z},
\widehat P_{Y\mid Z}^{\mathcal D,L}
\right)\\
&\quad\leq
C\left[
\sqrt{1+\log(1/\delta_L)}
\exp\left(-\frac{mT_L}{4}\right)
+\delta_L
+\sqrt{\frac{\delta_L}{L}}
+\frac1L
\right]
+\frac12\sqrt{\mathcal R_{N,L}^{*}}\\
&\quad\leq
C\left[
\sqrt{1+\log(1/\delta_L)}
e^{-cL\delta_L^3}
+\delta_L
+\sqrt{\frac{\delta_L}{L}}
+\frac1L
\right]
+\frac12\sqrt{\mathcal R_{N,L}^{*}},
\end{aligned}
\end{equation}
where $\widehat P_{Y\mid Z}^{\mathcal D,L}$ is the conditional distribution induced by learned Langevin-RePaint. In particular, for $\delta_L=L^{-1/4}$,
\begin{equation}
\label{eq_lr_main_tv_quarter}
\begin{aligned}
&\mathbb E_{\mathcal D}
\mathbb E_Z
d_{\mathrm{TV}}
\left(
P_{Y\mid Z},
\widehat P_{Y\mid Z}^{\mathcal D,L}
\right) \leq
C\left[
\sqrt{\log L}\,e^{-cL^{1/4}}
+L^{-1/4}
\right]
+\frac12\sqrt{\mathcal R_{N,L}^{*}}.
\end{aligned}
\end{equation}
Hence, if $L=L_N\to\infty$ and
$\mathcal R_{N,L_N}^{*}\to0$, then
\[
\mathbb E_{\mathcal D}
\mathbb E_Z
d_{\mathrm{TV}}
\left(
P_{Y\mid Z},
\widehat P_{Y\mid Z}^{\mathcal D,L_N}
\right)
\longrightarrow0.
\]
\end{theorem}

We prove the theorem through the following lemmas.

\begin{lemma}[Terminal-grid estimates]
\label{lem_lr_grid}
Suppose $\delta_L\leq1/2$ and $L\delta_L\geq4$. Then
\begin{equation}
\label{eq_lr_grid_basic}
\eta_{\ell_0}\asymp\delta_L^2,
\qquad
\max_{\ell_0\leq\ell\leq L-1}\eta_\ell
\leq4\delta_L^2,
\qquad
n_L\asymp L\delta_L.
\end{equation}
Writing
\[
a_\ell:=\frac{\eta_\ell}{\eta_{\ell+1}},
\qquad \ell_0\leq\ell\leq L-2,
\]
we have
\begin{equation}
\label{eq_lr_grid_ratio}
1\leq a_\ell\leq C,
\qquad
\sum_{\ell=\ell_0}^{L-2}
\eta_{\ell+1}(a_\ell-1)^2
\leq C\frac{\delta_L}{L}.
\end{equation}
Moreover, $T_L\asymp L\delta_L^3$. If
$\delta_L\to0$ and $L\delta_L\to\infty$, then
\begin{equation}
\label{eq_lr_grid_time_asymptotic}
T_L
=
\frac{L\delta_L^3}{3}(1+o(1)).
\end{equation}
\end{lemma}
\begin{proof}
Recall the definitions in (\ref{eq_lr_grid_notation}), since
\[
M_L=L-\lceil L(1-\delta_L)\rceil
=\lfloor L\delta_L\rfloor,
\qquad n_L=M_L-1,
\]
we have $M_L\asymp L\delta_L$ and
\[
\eta_{\ell_0}
=
\frac{M_L^2}{(L-M_L)^2}
\asymp\delta_L^2.
\]
The function $(1-t)^2/t^2$ is decreasing on $(0,1)$, and
$t_\ell\geq1-\delta_L$ in the terminal window. Thus
\[
\eta_\ell
\leq
\frac{\delta_L^2}{(1-\delta_L)^2}
\leq4\delta_L^2.
\]

For $\ell\leq L-2$,
\[
\frac{\sqrt{\eta_\ell}}{\sqrt{\eta_{\ell+1}}}
=
\frac{L-\ell}{L-\ell-1}\frac{\ell+1}{\ell}.
\]
Both factors are uniformly bounded on the terminal grid, proving
the bound on $a_\ell$. In addition,
\[
\sqrt{\eta_\ell}-\sqrt{\eta_{\ell+1}}
=
\frac{L-\ell}{\ell}
-\frac{L-\ell-1}{\ell+1}
=
\frac{L}{\ell(\ell+1)}
\leq\frac{C}{L}.
\]
Consequently,
\[
\begin{aligned}
\eta_{\ell+1}(a_\ell-1)^2
&=
\frac{(\eta_\ell-\eta_{\ell+1})^2}{\eta_{\ell+1}}\\
&=
\left(\sqrt{\eta_\ell}-\sqrt{\eta_{\ell+1}}\right)^2
\left(
\sqrt{\frac{\eta_\ell}{\eta_{\ell+1}}}+1
\right)^2
\leq\frac{C}{L^2}.
\end{aligned}
\]
Summing over $n_L\leq L\delta_L$ transitions yields
\eqref{eq_lr_grid_ratio}. Finally, setting $j=L-\ell-1$ gives
\[
T_L
=
\sum_{j=1}^{M_L-1}\frac{j^2}{(L-j)^2}.
\]
Therefore,
\[
\frac1{L^2}\sum_{j=1}^{M_L-1}j^2
\leq T_L
\leq
\frac1{(L-M_L)^2}\sum_{j=1}^{M_L-1}j^2.
\]
Since
\[
\sum_{j=1}^{M_L-1}j^2
=
\frac{(M_L-1)M_L(2M_L-1)}6,
\]
the stated comparisons and asymptotic formula follow.
\end{proof}

\begin{lemma}[Quadratic terminal score refinement]
\label{lem_lr_terminal_refinement}
Under Assumption~\ref{ass_lr_joint_reg}, there exists
$t_\dagger<1$ such that, for $t\in[t_\dagger,1)$,
\begin{equation}
\label{eq_lr_scaled_score_bound}
\|t\,s_t^*(tw)-s_1^*(w)\|
\leq
C(1-t)(1+\|w\|),
\qquad w\in\mathbb R^p.
\end{equation}
Consequently, for terminal grid points satisfying
$t_\ell\geq t_\dagger$,
\begin{equation}
\label{eq_lr_quadratic_drift_error}
\mathbb E_{e\sim\mathcal N(0,I_q)}
\|b_{t_\ell}^z(y;e)-b_z(y)\|^2
\leq
C(1-t_\ell)^2
(1+\|y\|^2+\|z\|^2).
\end{equation}
Moreover, for every fixed $z,e$, the map
$y\mapsto b_{t_\ell}^z(y;e)$ is globally Lipschitz, with a
constant independent of $\ell,z,e$ on this terminal interval.
\end{lemma}

\begin{proof}
Write
\[
\eta(t):=\frac{(1-t)^2}{t^2}.
\]
For fixed $t,w$, the posterior density of $X(1)$ given
$X(t)=tw$ is
\begin{equation}
\label{eq_post_x1_given_xt}
\nu_{t,w}(u)
=
\frac{
p_1(u)\exp(-\|u-w\|^2/(2\eta(t)))
}{
\int_{\mathbb R^p}
p_1(v)\exp(-\|v-w\|^2/(2\eta(t)))\,dv
}.    
\end{equation}

Its negative log-density, up to a constant independent of $u$, is
\[
V_{t,w}(u) :=
-\log p_1(u)+\frac{\|u-w\|^2}{2\eta(t)}.
\]
Assumption~\ref{ass_lr_joint_reg} implies
\begin{equation}
\label{eq_V_tw_strong_convex}
\nabla^2V_{t,w}(u)
\succeq
\big(\eta(t)^{-1}-K\big)I_p.    
\end{equation}

Choose $t_\dagger\geq1/2$ sufficiently close to $1$ that
$K\eta(t)\leq1/2$ whenever $t\geq t_\dagger$.
Then $V_{t,w}$ is strongly convex, with unique minimizer
$u_{t,w}^*$. The first-order condition is
\[
0
=
-s_1^*(u_{t,w}^*)
+\frac{u_{t,w}^*-w}{\eta(t)}.
\]
Hence
\[
\begin{aligned}
\|u_{t,w}^*-w\|
&=
\eta(t)\|s_1^*(u_{t,w}^*)\|\\
&\leq
\eta(t)\|s_1^*(w)\|
+\eta(t)K\|u_{t,w}^*-w\|,
\end{aligned}
\]
which yields
\begin{equation}
\label{eq_lr_posterior_mode_shift}
\|u_{t,w}^*-w\|
\leq2\eta(t)\|s_1^*(w)\|.
\end{equation}

Let $V\sim\nu_{t,w}$. Since
$\partial_j\nu_{t,w}
=-\nu_{t,w}\partial_jV_{t,w}$, integration by parts gives
\[
\begin{aligned}
0
&=
\int_{\mathbb R^p}
\partial_j\left[
(u_j-u_{t,w,j}^*)\nu_{t,w}(u)
\right]\,du\\
&=
1-
\mathbb E\left[
(V_j-u_{t,w,j}^*)\partial_jV_{t,w}(V)
\right].
\end{aligned}
\]
The boundary terms vanish because $V_{t,w}$ has a uniform
quadratic lower bound around its minimizer, while its gradient
has at most linear growth.
Summing over coordinates gives
\[
\mathbb E
\langle V-u_{t,w}^*,\nabla V_{t,w}(V)\rangle=p.
\]
Strong convexity and $\nabla V_{t,w}(u_{t,w}^*)=0$ imply
\[
\langle u-u_{t,w}^*,\nabla V_{t,w}(u)\rangle
\geq\frac1{2\eta(t)}\|u-u_{t,w}^*\|^2.
\]
Therefore,
\[
\mathbb E\|V-u_{t,w}^*\|^2\leq2p\eta(t).
\]
Together with \eqref{eq_lr_posterior_mode_shift},
\begin{equation}
\label{eq_lr_posterior_deviation}
\mathbb E\|V-w\|
\leq
\sqrt{2p\eta(t)}
+2\eta(t)\|s_1^*(w)\|.
\end{equation}

Let $\varphi_\eta$ denote the density of
$ N(0,\eta I_p)$ and define the convolution of $X(1)$ and $\frac{1-t}{t} X(0)$ as 
\[
r_t(w)
:=
\int p_1(u)\varphi_{\eta(t)}(w-u)\,du.
\]
Since $r_t$ is the density of $X(t)/t$,
\[
r_t(w)=t^p p_t(tw).
\]
Consequently,
\[
\nabla\log r_t(w)=t\,s_t^*(tw).
\]
Furthermore,
\begin{align}
\nabla r_t(w)
&=
-\int p_1(u)\nabla_u\varphi_{\eta(t)}(w-u)\,du\notag \\ 
&=
\int \nabla p_1(u)\varphi_{\eta(t)}(w-u)\,du \notag \\
&=
\int p_1(u)s_1^*(u)\varphi_{\eta(t)}(w-u)\,du. \label{eq_nabla_rt}
\end{align}
The integration by parts is justified by
$\nabla p_1=p_1s_1^*$, the linear growth of $s_1^*$, and the
finite second moment of $p_1$.
Dividing (\ref{eq_nabla_rt}) by $r_t(w)$ and (\ref{eq_post_x1_given_xt}) gives
\[
t\,s_t^*(tw)
=
\mathbb E_{\nu_{t,w}}s_1^*(V).
\]
It follows from \eqref{eq_lr_posterior_deviation} that
\[
\begin{aligned}
\|t\,s_t^*(tw)-s_1^*(w)\|
&\leq
K\mathbb E_{\nu_{t,w}}\|V-w\|\\
&\leq
K\sqrt{2p\eta(t)}
+2K\eta(t)\|s_1^*(w)\|\\
&\leq
C(1-t)(1+\|w\|),
\end{aligned}
\]
where we used $t\geq1/2$ and
$\|s_1^*(w)\|\leq\|s_1^*(0)\|+K\|w\|$.
This proves \eqref{eq_lr_scaled_score_bound}.

For the conditional drift, set
\[
w=(y,z+\sqrt{\eta_\ell}e).
\]
Then $t_\ell w=x_{t_\ell}(y,z,e)$, and
\[
\begin{aligned}
b_{t_\ell}^z(y;e)-b_z(y)
={}&
[t_\ell s_{t_\ell}^*(t_\ell w)-s_1^*(w)]_Y\\
&+[s_1^*(w)-s_1^*(y,z)]_Y.
\end{aligned}
\]
The first term is controlled by
\eqref{eq_lr_scaled_score_bound}. By (\ref{eq_score_1_lipschitz}) in Assumption \ref{ass_lr_joint_reg}, the second satisfies
\[
\|[s_1^*(w)-s_1^*(y,z)]_Y\|
\leq K\sqrt{\eta_\ell}\|e\|.
\]
Thus,
\[
\|b_{t_\ell}^z(y;e)-b_z(y)\|
\leq
C(1-t_\ell)(1+\|y\|+\|z\|+\|e\|).
\]
Squaring and using $\mathbb E\|e\|^2=q$ proves
\eqref{eq_lr_quadratic_drift_error}. 

We finally verify spatial regularity of the oracle drift.
Differentiating the convolution twice gives
\[
\nabla^2\log r_t(w)
=
\mathbb E_{\nu_{t,w}}\nabla s_1^*(V)
+
\operatorname{Cov}_{\nu_{t,w}}(s_1^*(V)).
\]
By (\ref{eq_V_tw_strong_convex}), the posterior $\nu_{t,w}$ is
$\lambda_t$-strongly log-concave, where
$\lambda_t:=\eta(t)^{-1}-K$.
Its Poincare inequality implies, for every unit vector $a$,
\[
\begin{aligned}
\operatorname{Var}_{\nu_{t,w}}(a^\top s_1^*(V))
&\leq
\lambda_t^{-1}
\mathbb E_{\nu_{t,w}}
\|[\nabla s_1^*(V)]^\top a\|^2\\
&\leq K^2/\lambda_t.
\end{aligned}
\]
Therefore,
\[
-KI_p
\preceq
\nabla^2\log r_t(w)
\preceq
\left(K+\frac{K^2}{\lambda_t}\right)I_p
\preceq2KI_p.
\]
Since
\[
b_{t_\ell}^z(y;e)
=
[\nabla\log r_{t_\ell}
(y,z+\sqrt{\eta_\ell}e)]_Y,
\]
the Lipschitz constant in $y$ is at most $2K$.
\end{proof}

\begin{lemma}[Oracle moment and initialization bounds]
\label{lem_lr_oracle_moments}
Under Assumptions~\ref{ass_lr_joint_reg},
\ref{ass_lr_cond_reg}, and \ref{ass_lr_warmup}, for sufficiently
small $\delta_L$,
\begin{equation}
\label{eq_lr_oracle_uniform_moment}
\sup_{\ell_0\leq\ell\leq L-1}
\mathbb E_{\mathcal D}
\mathbb E_Z
\mathbb E_{U_{t_\ell}^{*,Z}\mid\mathcal D,Z}
\|U_{t_\ell}^{*,Z}\|^2
\leq C.
\end{equation}
Moreover,
\begin{equation}
\label{eq_lr_initial_kl}
\mathbb E_{\mathcal D}\mathbb E_Z
D_{\mathrm{KL}}(\mu_{t_{\ell_0}}^{*,Z}\|\pi_Z)
\leq
C\big(1+\log(1/\delta_L)\big).
\end{equation}
\end{lemma}

\begin{proof}
For this proof, all expectations involving the oracle chain include
the training data $\mathcal D$, the observation $Z$, and oracle sampling randomness,
unless conditioning is explicitly displayed.

First, by (\ref{eq_lr_cond_hessian}) in Assumption \ref{ass_lr_cond_reg},
\[
\langle y,b_z(y)-b_z(0)\rangle
=
-\int_0^1
y^\top\nabla^2\mathbf{NLL}_z(ry)y\,dr
\leq-m\|y\|^2.
\]
Also,
\begin{align}
    \|b_z(0)\|& \leq || s^*_1(0,z)|| \notag \\
    & \leq \| s^*_1(0,0) \| + \|s^*_1(0,z) - s^*_1(0,0)\| \notag \\
    & \leq\|s_1^*(0,0)\|+K\|z\|. \notag 
\end{align}

Young's inequality therefore gives
\begin{align}
\label{eq_lr_true_dissipativity}
\langle y,b_z(y)\rangle & = \langle y,b_z(y) - b_z(0)\rangle + \langle y,b_z(0)\rangle \notag \\
& \leq -m\|y\|^2 + \| y \| \| b_z(0) \| \notag \\
& \leq
-\frac{3m}{4}\|y\|^2+C(1+\|z\|^2).
\end{align}

Set
\[
r_{t_\ell}^z(y;e):=b_{t_\ell}^z(y;e)-b_z(y).
\]
By (\ref{eq_lr_quadratic_drift_error}) in Lemma~\ref{lem_lr_terminal_refinement},
\[
\mathbb E_e\|r_{t_\ell}^z(y;e)\|^2
\leq
C\delta_L^2(1+\|y\|^2+\|z\|^2).
\]
Using Cauchy–Schwarz inequality and Young's inequality, we have
\[
\langle y,r_{t_\ell}^z(y;e)\rangle
\leq
\frac m4\|y\|^2+\frac1m\|r_{t_\ell}^z(y;e)\|^2.
\]
Choosing 
sufficiently small $\delta_L$ yields
\begin{equation}
\label{eq_lr_oracle_dissipativity}
\mathbb E_e
\langle y,b_{t_\ell}^z(y;e)\rangle
\leq
-c_0\|y\|^2+C_1(1+\|z\|^2)
\end{equation}
for some $c_0>0$.
The same drift-error bound and the linear growth of $s_1^*$ give
\begin{equation}
\label{eq_lr_oracle_growth}
\mathbb E_e\|b_{t_\ell}^z(y;e)\|^2
\leq C_2(1+\|y\|^2+\|z\|^2).
\end{equation}

Write
\[
M_\ell
:=
\mathbb E_{\mathcal D}\mathbb E_Z
\mathbb E_{U_{t_\ell}^{*,Z} \mid\mathcal D,Z}\|U_{t_\ell}^{*,Z}\|^2,
\qquad
A_Z:=1+\mathbb E\|Z\|^2.
\]
By calculating $\|U_{t_{\ell+1}}^{*,Z}\|^2$ in \eqref{eq_lr_oracle_score_chain} and taking the expectation, the cross terms
containing the fresh centered Gaussian $e_{\ell+1}^Y$ vanish.
Consequently,
\[
\begin{aligned}
M_{\ell+1}
={}&M_\ell
+\eta_\ell
\mathbb E
\langle U_{t_\ell}^{*,Z},
b_{t_\ell}^Z(U_{t_\ell}^{*,Z};e_\ell^Z)\rangle +
\frac{\eta_\ell^2}{4}
\mathbb E
\|b_{t_\ell}^Z(U_{t_\ell}^{*,Z};e_\ell^Z)\|^2
+d\eta_{\ell+1}\\
\leq{}&
\left(
1-c_0\eta_\ell+\frac{C_2}{4}\eta_\ell^2
\right)M_\ell
+
C_1A_Z\eta_\ell
+\frac{C_2A_Z}{4}\eta_\ell^2
+d\eta_{\ell+1}.
\end{aligned}
\]
Since $\eta_{\ell+1}\leq\eta_\ell$ and
$\max_\ell\eta_\ell\leq4\delta_L^2$, the positive quadratic
terms can be absorbed for sufficiently small $\delta_L$:
\[
M_{\ell+1}
\leq
(1-c_1\eta_\ell)M_\ell+C_3\eta_\ell,
\qquad 0\leq c_1\eta_\ell\leq1.
\]
The right-hand side is a convex combination of $M_\ell$ and
$C_3/c_1$, so
\[
M_\ell
\leq
\max\{M_{\ell_0},C_3/c_1\}.
\]
By \eqref{eq_lr_common_initialization},
\[
M_{\ell_0}
=
\mathbb E\|A_{L}^Z\|^2+d\eta_{\ell_0}\leq C.
\]
This proves \eqref{eq_lr_oracle_uniform_moment}.

We next control the initial KL divergence.
Let $y_z^*$ be the unique minimizer of
$\mathbf{NLL}_z$.
By the strong convexity condition
(\ref{eq_lr_cond_hessian}) in
Assumption~\ref{ass_lr_cond_reg},
$\nabla\mathbf{NLL}_z$ is $m$-strongly monotone.
Since $\nabla\mathbf{NLL}_z(y_z^*)=0$, we have
\[
m\|y_z^*\|^2
\leq
\left\langle
\nabla\mathbf{NLL}_z(0),-y_z^*
\right\rangle
\leq
\|\nabla\mathbf{NLL}_z(0)\|\|y_z^*\|.
\]
Moreover, using
$\nabla\mathbf{NLL}_z(0)=-b_z(0)$
and the $K$-Lipschitz continuity of $s_1^*$,
\[
\|y_z^*\|
\leq
\frac{\|\nabla\mathbf{NLL}_z(0)\|}{m}
=
\frac{\|b_z(0)\|}{m}
\leq
\frac{\|s_1^*(0,0)\|+K\|z\|}{m}
\leq C(1+\|z\|).
\]
Furthermore, the Hessian bounds in
(\ref{eq_lr_cond_hessian}), together with
$\nabla\mathbf{NLL}_z(y_z^*)=0$, imply
\[
\frac m2\|y-y_z^*\|^2
\leq
\mathbf{NLL}_z(y)-\mathbf{NLL}_z(y_z^*)
\leq
\frac K2\|y-y_z^*\|^2.
\]

Using the normalization
$\int_{\mathbb R^d}e^{-\mathbf{NLL}_z(y)}\,dy=1$
and the quadratic bounds implied by
(\ref{eq_lr_cond_hessian}) in
Assumption~\ref{ass_lr_cond_reg}, we obtain
\begin{equation}
\label{eq_lr_nll_min_bound}
\frac d2\log\frac{2\pi}{K}
\leq
\mathbf{NLL}_z(y_z^*)
\leq
\frac d2\log\frac{2\pi}{m}.
\end{equation}
Combining (\ref{eq_lr_nll_min_bound}) with the
upper quadratic bound from
(\ref{eq_lr_cond_hessian}) and the preceding
estimate $\|y_z^*\|\leq C(1+\|z\|)$ yields
\begin{equation}
\label{eq_lr_nll_quadratic_upper}
\mathbf{NLL}_z(y)
\leq C(1+\|y\|^2+\|z\|^2).
\end{equation}

We now bound the initial KL divergence.
Recall that $A_L^z$ denotes the $Y$-coordinate
terminal prediction produced by the learned
warm-up stage at step $\ell_0-1$, as defined
in Assumption~\ref{ass_lr_warmup}.
By the Gaussian re-noising step in
(\ref{eq_lr_update}), the initial state of
the rescaled oracle chain satisfies
\[
U_{t_{\ell_0}}^{*,z} {=}
A_L^z+\sqrt{\eta_{\ell_0}}e_{\ell_0}^Y,
\qquad
e_{\ell_0}^Y\sim  N(0,I_d),
\]
where $e_{\ell_0}^Y$ is independent of the
warm-up prediction $A_L^z$.
Consequently, conditional on $\mathcal D$
and $Z=z$, the initial law
$\mu_{t_{\ell_0}}^{*,z}$ is a Gaussian mixture
with mixing distribution given by $A_L^z$. By the convexity of KL divergence in its first
argument,
\begin{equation}
\begin{aligned}
&D_{\mathrm{KL}}
(\mu_{t_{\ell_0}}^{*,z}\|\pi_z) \leq
\mathbb E_{A_L^z\mid\mathcal D,z}
D_{\mathrm{KL}}\left(
 N(A_L^z,\eta_{\ell_0}I_d)
\|\pi_z
\right).
\end{aligned}
\end{equation}

For any fixed $a\in\mathbb R^d$, using
(\ref{eq_lr_nll_quadratic_upper}) and
$\mathbb E_{\mathcal N(a,\eta_{\ell_0}I_d)}
\|Y\|^2=\|a\|^2+d\eta_{\ell_0}$, we have
\begin{equation}
\begin{aligned}
D_{\mathrm{KL}}
\left( N(a,\eta_{\ell_0}I_d)\|\pi_z\right) &=
-\frac d2\log(2\pi e\eta_{\ell_0})
+
\mathbb E_{ N(a,\eta_{\ell_0}I_d)}
\mathbf{NLL}_z(Y)\\
&\leq
C(1+\|a\|^2+\|z\|^2+\eta_{\ell_0})
+\frac d2\log(1/\eta_{\ell_0}).
\end{aligned}
\end{equation}

Taking expectations over $\mathcal D$, $Z$,
and $A_L^Z$, and applying the warm-up moment
bound (\ref{eq_lr_warmup_moment}) in
Assumption~\ref{ass_lr_warmup}, we obtain
\begin{equation}
\begin{aligned}
\mathbb E_{\mathcal D,Z}
D_{\mathrm{KL}}
(\mu_{t_{\ell_0}}^{*,Z}\|\pi_Z)
&\leq
C\left(
1+B_0+\mathbb E\|Z\|^2
+\eta_{\ell_0}
+|\log\eta_{\ell_0}|
\right)\\
&\leq C(1+|\log\eta_{\ell_0}|),
\end{aligned}
\end{equation}
where the last inequality holds for
sufficiently large $L$. Finally, by Lemma~\ref{lem_lr_grid},
$\eta_{\ell_0}\asymp\delta_L^2$.
Therefore,
\[
\mathbb E_{\mathcal D,Z}
D_{\mathrm{KL}}
(\mu_{t_{\ell_0}}^{*,Z}\|\pi_Z)
\leq C(1+\log(1/\delta_L)),
\]
which proves (\ref{eq_lr_initial_kl}).
\end{proof}

\begin{lemma}[Convergence of the oracle half-score chain]
\label{lem_lr_oracle_tv}
Under the assumptions of
Lemma~\ref{lem_lr_oracle_moments},
\begin{equation}
\label{eq_lr_oracle_tv}
\begin{aligned}
&\mathbb E_{\mathcal D}\mathbb E_Z
d_{\mathrm{TV}}(\mu_{t_{L-1}}^{*,Z},\pi_Z) \leq
C\left[
\sqrt{1+\log(1/\delta_L)}
\exp\left(-\frac{mT_L}{4}\right)
+\delta_L+\sqrt{\frac{\delta_L}{L}}
\right].
\end{aligned}
\end{equation}
\end{lemma}

\begin{proof}
Fix $\mathcal D$ and $Z=z$ temporarily.
Unless otherwise specified, expectations in the
following calculations are conditional on
$\mathcal D$ and $Z=z$.

For each $\ell=\ell_0,\ldots,L-2$, we construct
an exact frozen-drift interpolation of the
oracle half-score recursion
(\ref{eq_lr_oracle_score_chain}) on the interval
$[t_\ell,t_{\ell+1}]$:
\begin{equation}
\label{eq_lr_frozen_interpolation}
\begin{aligned}
\overline U_t
={} U_{t_\ell}^{*,z}
+\frac{L\eta_\ell}{2}(t-t_\ell)
b_{t_\ell}^z(U_{t_\ell}^{*,z};e_\ell^Z) +\sqrt{L\eta_{\ell+1}}(B_t-B_{t_\ell}),
\end{aligned}
\end{equation}
where the Brownian increment is independent of
$U_{t_\ell}^{*,z}$ and $e_\ell^Z$.
By the uniform time grid in
Lemma~\ref{lem_lr_grid},
$t_{\ell+1}-t_\ell=1/L$.
Therefore,
\[
\begin{aligned}
\overline U_{t_{\ell+1}}
=U_{t_\ell}^{*,z}
+\frac{\eta_\ell}{2}
b_{t_\ell}^z(U_{t_\ell}^{*,z};e_\ell^Z) +\sqrt{\eta_{\ell+1}}e_{\ell+1}^Y,
\end{aligned}
\]
where
$e_{\ell+1}^Y
=\sqrt L(B_{t_{\ell+1}}-B_{t_\ell})
\sim\mathcal N(0,I_d)$.
This coincides with
(\ref{eq_lr_oracle_score_chain}), so the
terminal law of the interpolation is exactly
$\mu_{t_{\ell+1}}^{*,z}$.

Let $\rho_t$ denote the conditional density
of $\overline U_t$. Recall the definition of $a_\ell$ in Lemma \ref{lem_lr_grid} and define
\[
a_\ell=\frac{\eta_\ell}{\eta_{\ell+1}},
\qquad
\beta_t(y)
:=
\mathbb E\left[
a_\ell b_{t_\ell}^z
(U_{t_\ell}^{*,z};e_\ell^Z)
\mid \overline U_t=y,\mathcal D,Z=z
\right].
\]
For any smooth compactly supported test function
$\varphi$, It\^o's formula applied to
(\ref{eq_lr_frozen_interpolation}) gives
\[
\begin{aligned}
\frac{d}{dt}\mathbb E\varphi(\overline U_t)
={}&
\frac{L\eta_{\ell+1}}2
\mathbb E\Big[
\big\langle
a_\ell b_{t_\ell}^z
(U_{t_\ell}^{*,z};e_\ell^Z),
\nabla\varphi(\overline U_t)
\big\rangle 
+\Delta\varphi(\overline U_t)
\Big]\\
={}&
\frac{L\eta_{\ell+1}}2
\int_{\mathbb R^d}
\Big[
\langle\beta_t(y),\nabla\varphi(y)\rangle
+\Delta\varphi(y)
\Big]\rho_t(y)\,dy.
\end{aligned}
\]
Thus, the marginal density satisfies the
Fokker--Planck equation
\begin{equation}
\label{eq_lr_interpolation_fp}
\partial_t\rho_t
=
\frac{L\eta_{\ell+1}}2
\left[
\Delta\rho_t-\nabla\cdot(\rho_t\beta_t)
\right].
\end{equation}
The conditional expectation in $\beta_t$
is essential because the drift in
(\ref{eq_lr_frozen_interpolation}) is frozen
at the random step-starting state rather
than evaluated at $\overline U_t$.

Define the relative entropy and relative
score as
\[
\mathsf H_z(t)
:=D_{\mathrm{KL}}(\rho_t\|\pi_z),
\qquad
g_t(y):=\nabla\log\frac{\rho_t(y)}{\pi_z(y)}.
\]
Recall that $b_z=\nabla\log\pi_z$ defined in Appendix \ref{sec_notation}.
Consequently,
\[
\nabla\rho_t-\rho_tb_z=\rho_tg_t.
\]
Substituting this identity into
(\ref{eq_lr_interpolation_fp}) yields
\[
\partial_t\rho_t
=
\frac{L\eta_{\ell+1}}2
\nabla\cdot(\rho_tg_t)
-
\frac{L\eta_{\ell+1}}2
\nabla\cdot\big(\rho_t(\beta_t-b_z)\big).
\]

By mass conservation and integration by parts,
\[
\begin{aligned}
\frac{d}{dt}\mathsf H_z(t)
={}&
\int_{\mathbb R^d}
\log\frac{\rho_t(y)}{\pi_z(y)}
\partial_t\rho_t(y)\,dy\\
={}&
-\frac{L\eta_{\ell+1}}2
\int_{\mathbb R^d}\rho_t(y)\|g_t(y)\|^2\,dy +
\frac{L\eta_{\ell+1}}2
\int_{\mathbb R^d}\rho_t(y)
\langle g_t(y),\beta_t(y)-b_z(y)\rangle\,dy.
\end{aligned}
\]
Using the definition of relative Fisher
information and the tower property of
conditional expectation, we obtain
\begin{equation}
\label{eq_lr_entropy_identity}
\begin{aligned}
\frac{d}{dt}\mathsf H_z(t)
={}&
-\frac{L\eta_{\ell+1}}2
I(\rho_t\|\pi_z) +
\frac{L\eta_{\ell+1}}2
\mathbb E\Big[
\big\langle
g_t(\overline U_t),
a_\ell b_{t_\ell}^z
(U_{t_\ell}^{*,z};e_\ell^Z)
-b_z(\overline U_t)
\big\rangle
\Big].
\end{aligned}
\end{equation}
Here, the tower property allows us to replace
$\beta_t(\overline U_t)$ with the frozen
random drift inside the expectation.

We next justify the entropy calculation at
the interval endpoints.
For $t\in(t_\ell,t_{\ell+1}]$, the law
$\rho_t$ is a convolution with a nondegenerate
Gaussian of covariance
$L\eta_{\ell+1}(t-t_\ell)I_d$.
It is therefore smooth and strictly positive.
By conditioning the Gaussian score and applying
Jensen's inequality, its ordinary Fisher
information satisfies
\[
\int_{\mathbb R^d}
\|\nabla\log\rho_t(y)\|^2\rho_t(y)\,dy
\leq
\frac{d}{L\eta_{\ell+1}(t-t_\ell)}.
\]
Together with the second-moment bound
(\ref{eq_lr_oracle_uniform_moment}) and the
linear growth of $b_z$, this gives finite
relative Fisher information away from $t_\ell$.
The entropy calculation can thus be justified
using spatial cutoffs and bounded approximations
of $\log(\rho_t/\pi_z)$, followed by removal
of the regularization.

We also verify that there is no entropy jump
at $t_\ell$.
For $r=t-t_\ell$, define the conditional
deterministic transport map
\[
\mathcal T_{r,e}(y)
:=
y+\frac{L\eta_\ell r}{2}
b_{t_\ell}^z(y;e).
\]
Recall that
\[
b_t^z(y;e)
=
\left[
\nabla\log r_t\left(
y,z+\frac{1-t}{t}e
\right)
\right]_Y,
\]
where $r_t$ denotes the density of
$X(t)/t$.
By the smoothed-score Hessian bound
established in
Lemma~\ref{lem_lr_terminal_refinement},
\[
-KI_p\preceq
\nabla^2\log r_t(w)
\preceq 2KI_p
\]
on the terminal interval, for sufficiently
large $L$.
Consequently,
\[
\sup_{y,z,e}
\|\nabla_y b_{t_\ell}^z(y;e)\|_{\mathrm{op}}
\leq 2K.
\]
Hence, using $0\leq r\leq1/L$,
\[
\begin{aligned}
\|\nabla\mathcal T_{r,e}(y)-I_d\|_{\mathrm{op}}
&\leq KL\eta_\ell r\\
&\leq K\eta_\ell\\
&\leq C\delta_L^2,
\end{aligned}
\]
where the last inequality follows from
Lemma~\ref{lem_lr_grid}.
For sufficiently large $L$, this bound is
strictly less than one, ensuring that
$\mathcal T_{r,e}$ is a bi-Lipschitz
perturbation of the identity.

Since $e_\ell^Z$ is independent of
$U_{t_\ell}^{*,z}$, conditional on
$(\mathcal D,Z=z)$, the change-of-variables
formula gives
\[
\begin{aligned}
D_{\mathrm{KL}}\left(
(\mathcal T_{r,e})_\#
\mu_{t_\ell}^{*,z}\,\middle\|\,\pi_z
\right)
={}&
D_{\mathrm{KL}}(\mu_{t_\ell}^{*,z}\|\pi_z)\\
&+
\mathbb E_{U\sim\mu_{t_\ell}^{*,z}}
\Big[
\mathbf{NLL}_z(\mathcal T_{r,e}(U))
-\mathbf{NLL}_z(U)\\
&\hspace{29mm}
-\log|\det\nabla\mathcal T_{r,e}(U)|
\Big].
\end{aligned}
\]
By the Hessian bound
(\ref{eq_lr_cond_hessian}), the quadratic
upper bound
(\ref{eq_lr_nll_quadratic_upper}), and the
drift growth and moment bounds from
Lemmas~\ref{lem_lr_terminal_refinement}
and \ref{lem_lr_oracle_moments}, the
expectation on the right is bounded by
$C_{\mathcal D,z,\ell,e}r$,
where the constant has finite expectation
over $e=e_\ell^Z$.
Therefore,
\[
\begin{aligned}
D_{\mathrm{KL}}\left(
(\mathcal T_{r,e})_\#
\mu_{t_\ell}^{*,z}\,\middle\|\,\pi_z
\right) \leq
D_{\mathrm{KL}}(\mu_{t_\ell}^{*,z}\|\pi_z)
+C_{\mathcal D,z,\ell,e}r.
\end{aligned}
\]

Moreover, adding an independent Gaussian
perturbation of covariance $\sigma^2I_d$
increases the KL upper bound by at most
$Kd\sigma^2/2$.
Indeed, convexity of KL and the upper
Hessian bound in
(\ref{eq_lr_cond_hessian}) imply, for any
probability law $\nu$ with finite KL,
\[
D_{\mathrm{KL}}\left(
\nu*\mathcal N(0,\sigma^2I_d)\,\middle\|\,\pi_z
\right)
\leq
D_{\mathrm{KL}}(\nu\|\pi_z)
+\frac{Kd\sigma^2}{2}.
\]
Applying this with
$\sigma^2=L\eta_{\ell+1}r$ and averaging
over $e_\ell^Z$ yields
\[
\mathsf H_z(t)
\leq
D_{\mathrm{KL}}(\mu_{t_\ell}^{*,z}\|\pi_z)
+C_{\mathcal D,z,\ell}r.
\]
On the other hand, since
$\overline U_t\to U_{t_\ell}^{*,z}$
in distribution as $t\downarrow t_\ell$,
lower semicontinuity of relative entropy
gives
\[
D_{\mathrm{KL}}(\mu_{t_\ell}^{*,z}\|\pi_z)
\leq
\liminf_{t\downarrow t_\ell}\mathsf H_z(t).
\]
Combining these inequalities establishes
\[
\lim_{t\downarrow t_\ell}\mathsf H_z(t)
=
D_{\mathrm{KL}}(\mu_{t_\ell}^{*,z}\|\pi_z).
\]
Therefore, the integrated entropy inequality
below applies over the entire step. Applying Young's inequality to the second
term in (\ref{eq_lr_entropy_identity}) gives
\[
\begin{aligned}
\frac{d}{dt}\mathsf H_z(t)
\leq{} 
-\frac{L\eta_{\ell+1}}4
I(\rho_t\|\pi_z) +
\frac{L\eta_{\ell+1}}4
\mathbb E\left\|
a_\ell b_{t_\ell}^z
(U_{t_\ell}^{*,z};e_\ell^Z)
-b_z(\overline U_t)
\right\|^2.
\end{aligned}
\]
By the logarithmic Sobolev inequality
(\ref{eq_lr_lsi}),
$I(\rho_t\|\pi_z)\geq
2mD_{\mathrm{KL}}(\rho_t\|\pi_z)$.
Thus,
\begin{equation}
\label{eq_lr_entropy_differential}
\begin{aligned}
\frac{d}{dt}\mathsf H_z(t)
\leq{}&
-\frac m2L\eta_{\ell+1}\mathsf H_z(t) +
\frac{L\eta_{\ell+1}}4
\mathbb E\left\|
a_\ell b_{t_\ell}^z
(U_{t_\ell}^{*,z};e_\ell^Z)
-b_z(\overline U_t)
\right\|^2.
\end{aligned}
\end{equation}

We next bound the drift discrepancy.
Decompose it as
\begin{equation}
\label{eq_lr_entropy_error_decomposition}
\begin{aligned}
a_\ell b_{t_\ell}^z
(U_{t_\ell}^{*,z};e_\ell^Z)
-b_z(\overline U_t)
={}&
a_\ell\left[
b_{t_\ell}^z
(U_{t_\ell}^{*,z};e_\ell^Z)
-b_z(U_{t_\ell}^{*,z})
\right]\\
&+(a_\ell-1)b_z(U_{t_\ell}^{*,z})\\
&+\left[
b_z(U_{t_\ell}^{*,z})
-b_z(\overline U_t)
\right].
\end{aligned}
\end{equation}

In the following error estimates, expectations
are taken over $\mathcal D$, $Z$, and all
sampling randomness.

For the first term, the quadratic drift error
bound (\ref{eq_lr_quadratic_drift_error})
from Lemma~\ref{lem_lr_terminal_refinement},
together with the uniform moment bound
(\ref{eq_lr_oracle_uniform_moment}), yields
\[
\begin{aligned}
\mathbb E\left\|
b_{t_\ell}^Z
(U_{t_\ell}^{*,Z};e_\ell^Z)
-b_Z(U_{t_\ell}^{*,Z})
\right\|^2 
&\leq
C\delta_L^2
\mathbb E\left[
1+\|U_{t_\ell}^{*,Z}\|^2+\|Z\|^2
\right]\\
&\leq C\delta_L^2.
\end{aligned}
\]
Here, $e_\ell^Z$ is independent of
$U_{t_\ell}^{*,Z}$ conditional on
$\mathcal D$ and $Z$, allowing the drift
error bound to be applied conditionally.
Since $a_\ell$ is uniformly bounded by
(\ref{eq_lr_grid_ratio}), the squared norm
of the first term in
(\ref{eq_lr_entropy_error_decomposition})
is bounded by $C\delta_L^2$.

For the second term, the $K$-Lipschitz
continuity and linear growth of $b_z$,
together with
(\ref{eq_lr_oracle_uniform_moment}), give
\[
\begin{aligned}
\mathbb E\left\|
(a_\ell-1)b_Z(U_{t_\ell}^{*,Z})
\right\|^2 
&=(a_\ell-1)^2
\mathbb E\|b_Z(U_{t_\ell}^{*,Z})\|^2\\
&\quad\leq C(a_\ell-1)^2.
\end{aligned}
\]

For the third term, put
$r=t-t_\ell\leq1/L$.
By (\ref{eq_lr_frozen_interpolation}),
independence and centering of the
Brownian increment give
\[
\begin{aligned}
\mathbb E
\|\overline U_t-U_{t_\ell}^{*,Z}\|^2
&=
\frac{L^2\eta_\ell^2r^2}{4}
\mathbb E\left\|
b_{t_\ell}^Z
(U_{t_\ell}^{*,Z};e_\ell^Z)
\right\|^2
+dL\eta_{\ell+1}r\\
&\leq
C(\eta_\ell^2+\eta_{\ell+1})\\
&\leq C\delta_L^2,
\end{aligned}
\]
where the last two inequalities follow
from Lemmas~\ref{lem_lr_terminal_refinement},
\ref{lem_lr_oracle_moments}, and
\ref{lem_lr_grid}.
Since $b_z$ is $K$-Lipschitz,
\[
\begin{aligned}
\mathbb E
\|b_Z(U_{t_\ell}^{*,Z})
-b_Z(\overline U_t)\|^2
&\leq
K^2\mathbb E
\|U_{t_\ell}^{*,Z}-\overline U_t\|^2\\
&\leq C\delta_L^2.
\end{aligned}
\]

Combining the three estimates using
(\ref{eq_lr_entropy_error_decomposition})
and
$\|x_1+x_2+x_3\|^2
\leq3\sum_{i=1}^3\|x_i\|^2$,
we obtain
\begin{equation}
\label{eq_lr_entropy_error_bound}
\begin{aligned}
\mathbb E\left\|
a_\ell b_{t_\ell}^Z
(U_{t_\ell}^{*,Z};e_\ell^Z)
-b_Z(\overline U_t)
\right\|^2
\leq
C\left[
\delta_L^2+(a_\ell-1)^2
\right].
\end{aligned}
\end{equation}

Define
\[
H_\ell
:=
\mathbb E_{\mathcal D}\mathbb E_Z
D_{\mathrm{KL}}
(\mu_{t_\ell}^{*,Z}\|\pi_Z).
\]
Integrating
(\ref{eq_lr_entropy_differential})
over $[t_\ell,t_{\ell+1}]$,
then averaging over $\mathcal D,Z$
and using (\ref{eq_lr_entropy_error_bound}),
we obtain, with $\lambda=m/2$,
\begin{equation}
\label{eq_lr_oracle_entropy_recursion}
\begin{aligned}
H_{\ell+1}
\leq{}&
e^{-\lambda\eta_{\ell+1}}H_\ell +
C\left(1-e^{-\lambda\eta_{\ell+1}}\right)
\left[\delta_L^2+(a_\ell-1)^2\right].
\end{aligned}
\end{equation}
Here we used
$L\eta_{\ell+1}(t_{\ell+1}-t_\ell)
=\eta_{\ell+1}$. Iterating
(\ref{eq_lr_oracle_entropy_recursion})
from $\ell=\ell_0$ to $L-2$ gives
\[
\begin{aligned}
H_{L-1}
\leq{}&
e^{-\lambda T_L}H_{\ell_0} +
C\delta_L^2
\sum_{\ell=\ell_0}^{L-2}
e^{-\lambda\sum_{j=\ell+1}^{L-2}\eta_{j+1}}
\left(1-e^{-\lambda\eta_{\ell+1}}\right)\\
&+
C\sum_{\ell=\ell_0}^{L-2}
e^{-\lambda\sum_{j=\ell+1}^{L-2}\eta_{j+1}}
\left(1-e^{-\lambda\eta_{\ell+1}}\right)
(a_\ell-1)^2,
\end{aligned}
\]
where
\[
T_L:=\sum_{\ell=\ell_0}^{L-2}\eta_{\ell+1}.
\]

The first error sum telescopes:
\[
\begin{aligned}
\sum_{\ell=\ell_0}^{L-2}
e^{-\lambda\sum_{j=\ell+1}^{L-2}\eta_{j+1}}
\left(1-e^{-\lambda\eta_{\ell+1}}\right) =
1-e^{-\lambda T_L}
\leq1.
\end{aligned}
\]

For the second error sum, using
$1-e^{-x}\leq x$ for $x\geq0$ and
discarding the exponential weights, we obtain
\[
\begin{aligned}
\sum_{\ell=\ell_0}^{L-2}
e^{-\lambda\sum_{j=\ell+1}^{L-2}\eta_{j+1}}
\left(1-e^{-\lambda\eta_{\ell+1}}\right)
(a_\ell-1)^2 
&\leq
\lambda\sum_{\ell=\ell_0}^{L-2}
\eta_{\ell+1}(a_\ell-1)^2\\
&\leq C\frac{\delta_L}{L},
\end{aligned}
\]
where the last inequality follows directly
from (\ref{eq_lr_grid_ratio}). Moreover, the initial KL bound
(\ref{eq_lr_initial_kl}) gives
\[
H_{\ell_0}
\leq C\left(1+\log(1/\delta_L)\right).
\]
Combining these bounds and recalling
$\lambda=m/2$, we conclude that
\begin{equation}
\label{eq_lr_oracle_entropy_final}
\begin{aligned}
H_{L-1}
\leq{}&
C\left(1+\log(1/\delta_L)\right)
e^{-mT_L/2} +
C\left(\delta_L^2+\frac{\delta_L}{L}\right).
\end{aligned}
\end{equation}

Finally, Pinsker's inequality and
Jensen's inequality yield
\[
\begin{aligned}
\mathbb E_{\mathcal D}\mathbb E_Z
d_{\mathrm{TV}}
(\mu_{t_{L-1}}^{*,Z},\pi_Z) 
& \leq
\mathbb E_{\mathcal D}\mathbb E_Z
\sqrt{
\frac12
D_{\mathrm{KL}}
(\mu_{t_{L-1}}^{*,Z}\|\pi_Z)
}\\
&\leq\sqrt{\frac{H_{L-1}}2}.
\end{aligned}
\]
Substituting
(\ref{eq_lr_oracle_entropy_final})
and using $\sqrt{a+b}\leq\sqrt a+\sqrt b$
proves (\ref{eq_lr_oracle_tv}).
\end{proof}

\begin{lemma}[Learned-oracle Markov-chain comparison]
\label{lem_lr_markov_comparison}
Under Assumption~\ref{ass_lr_trajectory_risk},
suppose the learned and oracle terminal chains
share the common initialization
(\ref{eq_lr_common_initialization}). Then
\begin{equation}
\label{eq_lr_markov_tv}
\mathbb E_{\mathcal D}\mathbb E_Z
d_{\mathrm{TV}}
\left(
\mu_{t_{L-1}}^{Z},
\mu_{t_{L-1}}^{*,Z}
\right)
\leq
\frac12\sqrt{\mathcal R_{N,L}^{*}}.
\end{equation}
\end{lemma}

\begin{proof}
Fix $\mathcal D$ and $Z=z$ temporarily.
Unless otherwise specified, expectations below
are conditional on $(\mathcal D,Z=z)$.

Recall (\ref{eq_lr_learned_mean}) and (\ref{eq_lr_oracle_mean}). The conditional transition means of Langevin-RePaint update
(\ref{eq_lr_update}) and the oracle recursion
(\ref{eq_lr_oracle_chain_full}) are
\[
\begin{aligned}
m_{t_\ell}^{z}(y,e)
=
\left[
G_{t_\ell}
\big(x_{t_\ell}(y,z,e)\big)
\right]_Y \quad \text{and} \quad
m_{t_\ell}^{*,z}(y,e)
=
\left[
G_{t_\ell}^{*}
\big(x_{t_\ell}(y,z,e)\big)
\right]_Y.
\end{aligned}
\]

Conditional on the current state
$U_{t_\ell}=y$ and auxiliary observation
noise $e_\ell^Z=e$, the learned and oracle
transition kernels are
\[
k_{t_\ell}^{z}(\cdot\mid y,e)
=
\mathcal N\left(
m_{t_\ell}^{z}(y,e),
\eta_{\ell+1}I_d
\right) \quad \text{and}\quad 
k_{t_\ell}^{*,z}(\cdot\mid y,e)
=
\mathcal N\left(
m_{t_\ell}^{*,z}(y,e),
\eta_{\ell+1}I_d
\right).
\]
The conditioning on $e$ is important:
after integrating out $e$, the transition
is generally a Gaussian mixture rather
than a single Gaussian.

By the definitions of $G_t$ and $G_t^*$
in (\ref{eq_denoiser_to_score}), the
transition means differ by
\begin{equation}
\label{eq_lr_mean_difference}
\begin{aligned}
m_{t_\ell}^{z}(y,e)
-m_{t_\ell}^{*,z}(y,e) =
\frac{1-t_\ell}{2}
\left[
\widehat v_{t_\ell}
\big(x_{t_\ell}(y,z,e)\big)
-
v_{t_\ell}^{*}
\big(x_{t_\ell}(y,z,e)\big)
\right]_Y.
\end{aligned}
\end{equation}

For two Gaussian distributions with the
same covariance matrix $\eta I_d$,
\[
D_{\mathrm{KL}}
\left(
\mathcal N(a,\eta I_d)
\|
\mathcal N(b,\eta I_d)
\right)
=
\frac{\|a-b\|^2}{2\eta}.
\]
Indeed, the log-density ratio is
\[
\frac{\|x-b\|^2-\|x-a\|^2}{2\eta}.
\]
Under $x=a+\sqrt{\eta}\xi$ with
$\xi\sim\mathcal N(0,I_d)$, the cross
term has zero expectation, leaving
$\|a-b\|^2/(2\eta)$.

Consequently, substituting
(\ref{eq_lr_mean_difference}) into the
Gaussian KL identity gives
\begin{equation}
\label{eq_lr_one_step_kl_exact}
\begin{aligned}
D_{\mathrm{KL}}
\left(
k_{t_\ell}^{*,z}(\cdot\mid y,e)
\|
k_{t_\ell}^{z}(\cdot\mid y,e)
\right) =
\frac{(1-t_\ell)^2}{8\eta_{\ell+1}}
\left\|
\left[
\widehat v_{t_\ell}
\big(x_{t_\ell}(y,z,e)\big)
-
v_{t_\ell}^{*}
\big(x_{t_\ell}(y,z,e)\big)
\right]_Y
\right\|^2.
\end{aligned}
\end{equation}
By the uniform time grid in
Lemma~\ref{lem_lr_grid}, we have
$t_\ell=\ell/L$ and
$\eta_\ell=(1-t_\ell)^2/t_\ell^2$.
Therefore, for $\ell\leq L-2$,
\[
\begin{aligned}
\frac{(1-t_\ell)^2}{8\eta_{\ell+1}}
=
\frac{t_{\ell+1}^2}{8}
\left(
\frac{L-\ell}{L-\ell-1}
\right)^2 \leq\frac12,
\end{aligned}
\]
where we used
$t_{\ell+1}\leq1$ and
$(L-\ell)/(L-\ell-1)\leq2$.

Thus, since the Euclidean norm of a
coordinate subvector cannot exceed
that of the full vector,
\begin{equation}
\label{eq_lr_one_step_kl_bound}
\begin{aligned}
&D_{\mathrm{KL}}
\left(
k_{t_\ell}^{*,z}(\cdot\mid y,e)
\|
k_{t_\ell}^{z}(\cdot\mid y,e)
\right)\\
&\quad\leq
\frac12
\left\|
\widehat v_{t_\ell}
\big(x_{t_\ell}(y,z,e)\big)
-
v_{t_\ell}^{*}
\big(x_{t_\ell}(y,z,e)\big)
\right\|^2.
\end{aligned}
\end{equation}

We next compare the full Markov-chain
path distributions.
Consider the augmented paths
\[
\left(
U_{t_{\ell_0}},e_{\ell_0}^Z,
U_{t_{\ell_0+1}},\ldots,
e_{L-2}^Z,U_{t_{L-1}}
\right).
\]
Let their conditional laws under the
oracle and learned chains be denoted
by $\mathsf P_{\mathrm{path}}^{*,z}$
and $\mathsf P_{\mathrm{path}}^{z}$,
respectively.

At every step, the auxiliary Gaussian
noise $e_\ell^Z$ is sampled independently
of the current state, conditional on
$\mathcal D$ and $Z=z$, and has the same
distribution in both chains.
Consequently, the augmented path densities
factorize as
\[
\begin{aligned}
p_{\mathrm{path}}^{*,z}
={}&
\mu_{t_{\ell_0}}^{*,z}(u_{t_{\ell_0}})
\prod_{\ell=\ell_0}^{L-2}
\phi_q(e_\ell)
k_{t_\ell}^{*,z}
(u_{t_{\ell+1}}\mid u_{t_\ell},e_\ell),\\
p_{\mathrm{path}}^{z}
={}&
\mu_{t_{\ell_0}}^{z}(u_{t_{\ell_0}})
\prod_{\ell=\ell_0}^{L-2}
\phi_q(e_\ell)
k_{t_\ell}^{z}
(u_{t_{\ell+1}}\mid u_{t_\ell},e_\ell),
\end{aligned}
\]
where $\phi_q$ denotes the standard
Gaussian density on $\mathbb R^q$.

By the common initialization
(\ref{eq_lr_common_initialization}),
\[
\mu_{t_{\ell_0}}^{z}
=
\mu_{t_{\ell_0}}^{*,z}.
\]
Hence, the initial density terms and
the auxiliary-noise densities cancel
in the pathwise log-density ratio.

Taking expectation under the oracle
path measure and applying the KL
chain rule yields
\begin{equation}
\label{eq_lr_path_kl}
\begin{aligned}
D_{\mathrm{KL}}
\left(
\mathsf P_{\mathrm{path}}^{*,z}
\|
\mathsf P_{\mathrm{path}}^{z}
\right)
&=
\sum_{\ell=\ell_0}^{L-2}
\mathbb E_{*\mid\mathcal D,z}
D_{\mathrm{KL}}
\left(
k_{t_\ell}^{*,z}
(\cdot\mid U_{t_\ell}^{*,z},e_\ell^Z)
\|
k_{t_\ell}^{z}
(\cdot\mid U_{t_\ell}^{*,z},e_\ell^Z)
\right)\\
&\leq
\frac12
\sum_{\ell=\ell_0}^{L-2}
\mathbb E_{*\mid\mathcal D,z}
\left\|
\widehat v_{t_\ell}
(\widetilde X_{t_\ell}^{*,z})
-
v_{t_\ell}^{*}
(\widetilde X_{t_\ell}^{*,z})
\right\|^2.
\end{aligned}
\end{equation}
The last inequality follows from
(\ref{eq_lr_one_step_kl_bound}), where
$\widetilde X_{t_\ell}^{*,z}$ is the oracle
velocity-field input defined in
(\ref{eq_lr_oracle_input}).
The direction of KL divergence is
deliberate: it places the learning
error under the oracle trajectory
law rather than the learned one.

Since marginalization cannot increase
KL divergence, the data-processing
inequality gives
\[
\begin{aligned}
D_{\mathrm{KL}}
\left(
\mu_{t_{L-1}}^{*,z}
\|
\mu_{t_{L-1}}^{z}
\right) \leq
D_{\mathrm{KL}}
\left(
\mathsf P_{\mathrm{path}}^{*,z}
\|
\mathsf P_{\mathrm{path}}^{z}
\right).
\end{aligned}
\]

Applying Pinsker's inequality and
(\ref{eq_lr_path_kl}), we obtain
\[
\begin{aligned}
d_{\mathrm{TV}}
\left(
\mu_{t_{L-1}}^{*,z},
\mu_{t_{L-1}}^{z}
\right)
&\leq
\sqrt{
\frac12
D_{\mathrm{KL}}
\left(
\mu_{t_{L-1}}^{*,z}
\|
\mu_{t_{L-1}}^{z}
\right)
}\\
&\leq
\frac12
\left[
\sum_{\ell=\ell_0}^{L-2}
\mathbb E_{*\mid\mathcal D,z}
\left\|
\widehat v_{t_\ell}
(\widetilde X_{t_\ell}^{*,z})
-
v_{t_\ell}^{*}
(\widetilde X_{t_\ell}^{*,z})
\right\|^2
\right]^{1/2}.
\end{aligned}
\]

Finally, taking expectations over
$\mathcal D,Z$ and applying Jensen's
inequality gives
\[
\begin{aligned}
\mathbb E_{\mathcal D}\mathbb E_Z
d_{\mathrm{TV}}
\left(
\mu_{t_{L-1}}^{Z},
\mu_{t_{L-1}}^{*,Z}
\right)
&\leq
\frac12
\left[
\sum_{\ell=\ell_0}^{L-2}
\mathbb E_{\mathcal D}\mathbb E_Z
\mathbb E_{*\mid\mathcal D,Z}
\left\|
\widehat v_{t_\ell}
(\widetilde X_{t_\ell}^{*,Z})
-
v_{t_\ell}^{*}
(\widetilde X_{t_\ell}^{*,Z})
\right\|^2
\right]^{1/2}\\
&=
\frac12\sqrt{\mathcal R_{N,L}^{*}},
\end{aligned}
\]
where the last equality follows from
the cumulative trajectory risk definition
(\ref{eq_lr_trajectory_risk}).
This proves (\ref{eq_lr_markov_tv}).
\end{proof}

\begin{lemma}[Final non-re-noised prediction]
\label{lem_lr_final_prediction}
Under Assumptions~\ref{ass_lr_joint_reg},
\ref{ass_lr_cond_reg}, and
\ref{ass_lr_final_reg}, the final learned
prediction kernel $Q_{\mathcal D,L}^z$
defined below satisfies
\begin{equation}
\label{eq_lr_final_kernel_tv}
\mathbb E_{\mathcal D}\mathbb E_Z
d_{\mathrm{TV}}
\left(
\pi_Z Q_{\mathcal D,L}^Z,\pi_Z
\right)
\leq \frac{C}{L}
\end{equation}
for sufficiently large $L$.
\end{lemma}

\begin{proof}
Set $t_*=t_{L-1}=1-1/L$ and
$h_L=(1-t_*)/2=1/(2L)$.
For fixed $\mathcal D$, $z$, and $e$, define
\begin{equation}
\label{eq_lr_final_map}
\begin{aligned}
H_{\mathcal D,L,z,e}(y)
&:=
\left[
\widehat v_{t_*}
\left(t_*y,t_*z+\frac{e}{L}\right)
\right]_Y-y,\\
F_{\mathcal D,L,z,e}(y)
&:=
y+h_LH_{\mathcal D,L,z,e}(y).
\end{aligned}
\end{equation}

By the definition of the half-score
correction $G_t$ in
(\ref{eq_denoiser_to_score}), we have
\[
\begin{aligned}
\left[
G_{t_*}
\left(t_*y,t_*z+\frac{e}{L}\right)
\right]_Y
&=
\frac{1+t_*}{2}y
+\frac{1-t_*}{2}
\left[
\widehat v_{t_*}
\left(t_*y,t_*z+\frac{e}{L}\right)
\right]_Y\\
&=
F_{\mathcal D,L,z,e}(y).
\end{aligned}
\]
Therefore, the final output in
(\ref{eq_langevin_repaint_update})
can be written as
\[
\widehat Y_{t_L}^z
=
F_{\mathcal D,L,z,e_{L-1}^Z}
\left(U_{t_{L-1}}^z\right),
\]
where $e_{L-1}^Z\sim\mathcal N(0,I_q)$
is independent of $U_{t_{L-1}}^z$
conditional on $\mathcal D$ and $Z=z$.

We define the final prediction kernel by
\[
Q_{\mathcal D,L}^z(y,A)
:=
\mathbb P_{e\sim\mathcal N(0,I_q)}
\left\{
F_{\mathcal D,L,z,e}(y)\in A
\right\}.
\]
Consequently, the conditional distribution
of the final output is
$\mu_{t_{L-1}}^z Q_{\mathcal D,L}^z$.

We next show that this kernel perturbs
the target distribution $\pi_z$ by at most
$O(L^{-1})$ in total variation distance. Fix $\mathcal D$, $z$, and $e$, and abbreviate
\[
H=H_{\mathcal D,L,z,e},
\qquad
F=F_{\mathcal D,L,z,e},
\qquad
h=h_L.
\]
By the Jacobian bound
(\ref{eq_lr_final_reg}) in
Assumption~\ref{ass_lr_final_reg},
\[
\begin{aligned}
DH(y)
={}&
t_*
\left[
\nabla_x\widehat v_{t_*}
\left(t_*y,t_*z+\frac{e}{L}\right)
\right]_{YY}
-I_d.
\end{aligned}
\]
Therefore,
\[
\|DH(y)\|_{\mathrm{op}}
\leq 1+t_*B
\leq 1+B.
\]

For sufficiently large $L$,
$h(1+B)<1/2$.
Hence $F$ is a global $C^1$
diffeomorphism.
Indeed, for every $x\in\mathbb R^d$,
the equation
\[
y=x-hH(y)
\]
has a unique solution by the contraction
mapping theorem, since
$h\|DH\|_{\mathrm{op}}<1$.
Moreover,
\[
DF(y)=I_d+hDH(y)
\]
is invertible everywhere.
Its determinant is positive because
$I_d+rhDH(y)$ remains invertible for
all $r\in[0,1]$ and its determinant
equals one at $r=0$.

Recall that
$\mathbf{NLL}_z(y)=-\log\pi_z(y)$.
By the change-of-variables formula,
\begin{equation}
\label{eq_lr_final_map_kl_identity}
\begin{aligned}
D_{\mathrm{KL}}(F_\#\pi_z\|\pi_z)
=
\mathbb E_{Y\sim\pi_z}
\Big[
\mathbf{NLL}_z(Y+hH(Y))
-\mathbf{NLL}_z(Y) -\log\det(I_d+hDH(Y))
\Big].
\end{aligned}
\end{equation}

By the Hessian upper bound
(\ref{eq_lr_cond_hessian}) in
Assumption~\ref{ass_lr_cond_reg},
Taylor's theorem gives
\[
\begin{aligned}
\mathbf{NLL}_z(y+hH(y))
-\mathbf{NLL}_z(y)
\leq
h\langle
\nabla\mathbf{NLL}_z(y),H(y)
\rangle
+\frac K2h^2\|H(y)\|^2.
\end{aligned}
\]

Furthermore, since
$h\|DH(y)\|_{\mathrm{op}}<1/2$,
the convergent matrix logarithm expansion
yields
\[
\begin{aligned}
-\log\det(I_d+hDH(y))
\leq{}&
-h\operatorname{div}H(y) +Cdh^2(1+B)^2.
\end{aligned}
\]
This estimate does not require $DH(y)$
to be symmetric. Substituting these two inequalities into
(\ref{eq_lr_final_map_kl_identity}) gives
\[
\begin{aligned}
D_{\mathrm{KL}}(F_\#\pi_z\|\pi_z)
&\leq
h\mathbb E_{\pi_z}
\Big[
\langle\nabla\mathbf{NLL}_z(Y),H(Y)\rangle
-\operatorname{div}H(Y)
\Big]\\
&\qquad+
Ch^2\left[
\mathbb E_{\pi_z}\|H(Y)\|^2
+d(1+B)^2
\right].
\end{aligned}
\]

Since
$\nabla\pi_z=-\pi_z\nabla\mathbf{NLL}_z$,
integration by parts yields
\[
\begin{aligned}
\mathbb E_{\pi_z}
\langle\nabla\mathbf{NLL}_z(Y),H(Y)\rangle
&=
-\int_{\mathbb R^d}
\langle\nabla\pi_z(y),H(y)\rangle\,dy\\
&=
\int_{\mathbb R^d}
\pi_z(y)\operatorname{div}H(y)\,dy\\
&=
\mathbb E_{\pi_z}\operatorname{div}H(Y).
\end{aligned}
\]

The boundary terms vanish because the
strong convexity condition
(\ref{eq_lr_cond_hessian}) implies
Gaussian-type tails for $\pi_z$,
while $H$ has at most linear growth
and a bounded Jacobian.
Consequently, the first-order terms
in $h$ cancel exactly, giving
\begin{equation}
\label{eq_lr_final_map_kl_bound}
D_{\mathrm{KL}}(F_\#\pi_z\|\pi_z)
\leq
Ch^2\left[
\mathbb E_{\pi_z}\|H(Y)\|^2
+d(1+B)^2
\right].
\end{equation}

We next control the second moment of
$H_{\mathcal D,L,z,e}$.
By the global Jacobian bound
(\ref{eq_lr_final_reg}) and the
mean-value inequality,
\[
\begin{aligned}
\|H_{\mathcal D,L,z,e}(y)\|
\leq{}&
\|\widehat v_{t_*}(0)\|
+(1+B)\|y\| +B\|z\|+\frac BL\|e\|.
\end{aligned}
\]

Using $(a_1+\cdots+a_4)^2
\leq4\sum_{i=1}^4a_i^2$, we obtain
\[
\begin{aligned}
\mathbb E_{\mathcal D}\mathbb E_Z
\mathbb E_{e\sim\mathcal N(0,I_q)}
\mathbb E_{Y\sim\pi_Z}
\|H_{\mathcal D,L,Z,e}(Y)\|^2
&\leq
C\Big[
\mathbb E_{\mathcal D}
\|\widehat v_{t_*}(0)\|^2
+\mathbb E\|Y\|^2 +\mathbb E\|Z\|^2
+\frac{1}{L^2}\mathbb E\|e\|^2
\Big]\\
&\leq C.
\end{aligned}
\]
Here, the final inequality follows from
the velocity moment bound
(\ref{eq_lr_final_reg}) in
Assumption~\ref{ass_lr_final_reg}
and the finite joint second moment in
Assumption~\ref{ass_lr_joint_reg}.
In particular, when $Z\sim P_Z$ and
$Y\mid Z\sim\pi_Z$, the pair $(Y,Z)$
follows the true joint data distribution.

Taking expectations in
(\ref{eq_lr_final_map_kl_bound}) over
$\mathcal D$, $Z$, and $e$, and recalling
$h=1/(2L)$, we obtain
\begin{equation}
\label{eq_lr_final_expected_kl}
\begin{aligned}
\mathbb E_{\mathcal D}\mathbb E_Z
\mathbb E_e
D_{\mathrm{KL}}
\left(
(F_{\mathcal D,L,Z,e})_\#\pi_Z
\middle\|\pi_Z
\right)
\leq \frac{C}{L^2}.
\end{aligned}
\end{equation}

Finally, by the definition of
$Q_{\mathcal D,L}^Z$, convexity of
total variation distance under mixtures
gives
\[
\begin{aligned}
d_{\mathrm{TV}}
\left(
\pi_ZQ_{\mathcal D,L}^Z,\pi_Z
\right)
\leq
\mathbb E_e
d_{\mathrm{TV}}
\left(
(F_{\mathcal D,L,Z,e})_\#\pi_Z,
\pi_Z
\right).
\end{aligned}
\]

Applying Pinsker's inequality and
Jensen's inequality, we conclude that
\[
\begin{aligned}
\mathbb E_{\mathcal D}\mathbb E_Z
d_{\mathrm{TV}}
\left(
\pi_ZQ_{\mathcal D,L}^Z,\pi_Z
\right)
&\leq
\mathbb E_{\mathcal D}\mathbb E_Z\mathbb E_e
d_{\mathrm{TV}}
\left(
(F_{\mathcal D,L,Z,e})_\#\pi_Z,\pi_Z
\right)\\
&\leq
\left[
\frac12
\mathbb E_{\mathcal D}\mathbb E_Z\mathbb E_e
D_{\mathrm{KL}}
\left(
(F_{\mathcal D,L,Z,e})_\#\pi_Z
\middle\|\pi_Z
\right)
\right]^{1/2}\\
&\leq \frac{C}{L},
\end{aligned}
\]
where the last inequality follows from
(\ref{eq_lr_final_expected_kl}).
This proves (\ref{eq_lr_final_kernel_tv}).
\end{proof}

\subsection{Proof of Theorem~\ref{thm_lr_conditional_tv}}

\begin{proof}
Fix $\mathcal D$ and $Z=z$ temporarily.
By the final output convention in
(\ref{eq_langevin_repaint_update})
and the definition of the final prediction
kernel $Q_{\mathcal D,L}^z$ in
Lemma~\ref{lem_lr_final_prediction},
the final observation noise $e_{L-1}^Z$
is independent of $U_{t_{L-1}}^z$
conditional on $\mathcal D$ and $Z=z$.
Therefore, the conditional distribution
generated by Langevin-RePaint is
\begin{equation}
\label{eq_lr_output_distribution}
\widehat P_{Y\mid Z=z}^{\mathcal D,L}
=
\mu_{t_{L-1}}^z Q_{\mathcal D,L}^z.
\end{equation}
Applying the triangle inequality and
the contraction of total variation distance
under a common Markov kernel, we obtain
\begin{equation}
\label{eq_lr_main_tv_decomposition}
\begin{aligned}
d_{\mathrm{TV}}
\left(
\widehat P_{Y\mid Z=z}^{\mathcal D,L},
\pi_z
\right)
&\leq
d_{\mathrm{TV}}
\left(
\mu_{t_{L-1}}^z Q_{\mathcal D,L}^z,
\mu_{t_{L-1}}^{*,z} Q_{\mathcal D,L}^z
\right)\\
&\quad+
d_{\mathrm{TV}}
\left(
\mu_{t_{L-1}}^{*,z}Q_{\mathcal D,L}^z,
\pi_z Q_{\mathcal D,L}^z
\right)\\
&\quad+
d_{\mathrm{TV}}
\left(
\pi_z Q_{\mathcal D,L}^z,\pi_z
\right)\\
&\leq
d_{\mathrm{TV}}
\left(
\mu_{t_{L-1}}^z,
\mu_{t_{L-1}}^{*,z}
\right)\\
&\quad+
d_{\mathrm{TV}}
\left(
\mu_{t_{L-1}}^{*,z},\pi_z
\right)
+
d_{\mathrm{TV}}
\left(
\pi_zQ_{\mathcal D,L}^z,\pi_z
\right).
\end{aligned}
\end{equation}

For completeness, the contraction property
follows from
\[
(\mu Q-\nu Q)(A)
=
\int Q(y,A)(\mu-\nu)(dy),
\]
where $0\leq Q(y,A)\leq1$.
By the variational characterization of
total variation distance,
\[
d_{\mathrm{TV}}(\mu Q,\nu Q)
\leq d_{\mathrm{TV}}(\mu,\nu).
\]

We now take expectations over $\mathcal D$
and $Z$ in (\ref{eq_lr_main_tv_decomposition}).

First, the learned-oracle Markov-chain
comparison bound (\ref{eq_lr_markov_tv})
in Lemma~\ref{lem_lr_markov_comparison}
gives
\[
\begin{aligned}
\mathbb E_{\mathcal D}\mathbb E_Z
d_{\mathrm{TV}}
\left(
\mu_{t_{L-1}}^Z,
\mu_{t_{L-1}}^{*,Z}
\right)
&\leq
\frac12\sqrt{\mathcal R_{N,L}^{*}},
\end{aligned}
\]
where the cumulative oracle-trajectory
velocity risk $\mathcal R_{N,L}^{*}$
is defined in (\ref{eq_lr_trajectory_risk}).

Second, the oracle convergence bound
(\ref{eq_lr_oracle_tv}) in
Lemma~\ref{lem_lr_oracle_tv} yields
\[
\begin{aligned}
\mathbb E_{\mathcal D}\mathbb E_Z
d_{\mathrm{TV}}
\left(
\mu_{t_{L-1}}^{*,Z},\pi_Z
\right)
&\leq
C\left[
\sqrt{1+\log(1/\delta_L)}
\exp\left(-\frac{mT_L}{4}\right)
+\delta_L+\sqrt{\frac{\delta_L}{L}}
\right].
\end{aligned}
\]
Third, the final prediction bound
(\ref{eq_lr_final_kernel_tv}) in
Lemma~\ref{lem_lr_final_prediction}
gives
\[
\mathbb E_{\mathcal D}\mathbb E_Z
d_{\mathrm{TV}}
\left(
\pi_ZQ_{\mathcal D,L}^Z,\pi_Z
\right)
\leq\frac{C}{L}.
\]
Combining these three bounds, we obtain
\begin{equation}
\label{eq_lr_main_tv_combined}
\begin{aligned}
\mathbb E_{\mathcal D}\mathbb E_Z
d_{\mathrm{TV}}
\left(
\widehat P_{Y\mid Z}^{\mathcal D,L},
\pi_Z
\right)
&\leq
C\left[
\sqrt{1+\log(1/\delta_L)}
\exp\left(-\frac{mT_L}{4}\right)
+\delta_L
+\sqrt{\frac{\delta_L}{L}}
+\frac1L
\right]\\
&\quad+
\frac12\sqrt{\mathcal R_{N,L}^{*}}.
\end{aligned}
\end{equation}

Furthermore, Lemma~\ref{lem_lr_grid}
establishes that
\[
T_L\geq c_0 L\delta_L^3
\]
for some constant $c_0>0$.
Consequently,
\[
\exp\left(-\frac{mT_L}{4}\right)
\leq
\exp\left(-cL\delta_L^3\right),
\qquad c=\frac{mc_0}{4}>0.
\]
Substituting this inequality into
(\ref{eq_lr_main_tv_combined}) gives
the bound in (\ref{eq_lr_main_tv_bound}).

Now choose $\delta_L=L^{-1/4}$.
By Lemma~\ref{lem_lr_grid},
\[
T_L\asymp L^{1/4},
\qquad
\sqrt{\frac{\delta_L}{L}}=L^{-5/8},
\qquad
L^{-1}\leq L^{-1/4}.
\]
Moreover,
\[
\sqrt{1+\log(1/\delta_L)}
=
\sqrt{1+\frac14\log L}.
\]
Therefore, absorbing lower-order terms
into the constant $C$, we obtain
\begin{equation}
\label{eq_lr_main_tv_quarter_proof}
\begin{aligned}
\mathbb E_{\mathcal D}\mathbb E_Z
d_{\mathrm{TV}}
\left(
\widehat P_{Y\mid Z}^{\mathcal D,L},
\pi_Z
\right)
&\leq
C\left[
\sqrt{1+\log L}\,
e^{-cL^{1/4}}
+L^{-1/4}
\right]
+\frac12\sqrt{\mathcal R_{N,L}^{*}}.
\end{aligned}
\end{equation}
This proves (\ref{eq_lr_main_tv_quarter}).

Finally, suppose $L=L_N\to\infty$
as the training sample size $N\to\infty$.
Under the oracle-trajectory velocity risk
condition in
Assumption~\ref{ass_lr_trajectory_risk},
\[
\mathcal R_{N,L_N}^{*}
\longrightarrow0.
\]
Since
\[
\sqrt{1+\log L_N}\,
e^{-cL_N^{1/4}}\longrightarrow0,
\qquad
L_N^{-1/4}\longrightarrow0,
\]
it follows from
(\ref{eq_lr_main_tv_quarter_proof}) that
\[
\mathbb E_{\mathcal D}\mathbb E_Z
d_{\mathrm{TV}}
\left(
\widehat P_{Y\mid Z}^{\mathcal D,L_N},
\pi_Z
\right)
\longrightarrow0.
\]
This establishes the claimed
asymptotic consistency of Langevin-RePaint
in total variation distance.
\end{proof}

\paragraph{Choice of the terminal correction window.}

The preceding analysis reveals two competing
requirements for the terminal correction window.
The window must shrink so that the terminal
drift approximation error vanishes, while
containing sufficient effective sampling time
to eliminate the bias inherited from the
warm-up stage.

More precisely, by the oracle convergence
bound (\ref{eq_lr_oracle_tv}) in
Lemma~\ref{lem_lr_oracle_tv} and the
effective-time bound in
Lemma~\ref{lem_lr_grid}, sufficient
conditions for convergence of the oracle
chain are
\begin{equation}
\label{eq_lr_window_conditions}
\delta_L\longrightarrow0,
\qquad
\sqrt{1+\log(1/\delta_L)}
e^{-cL\delta_L^3}\longrightarrow0.
\end{equation}
Indeed, the remaining error term
$\sqrt{\delta_L/L}$ in
(\ref{eq_lr_oracle_tv}) vanishes
automatically under these conditions
as $L\to\infty$.

For the polynomial choice
$\delta_L=L^{-\gamma}$, every fixed
$0<\gamma<1/3$ satisfies
(\ref{eq_lr_window_conditions}).
We choose $\gamma=1/4$, which gives
\[
\delta_L=L^{-1/4},
\qquad
T_L\asymp L^{1/4},
\qquad
n_L\asymp L^{3/4},
\]
where $T_L$ is the accumulated effective
sampling time and $n_L$ is the number
of terminal correction steps, as defined
in Lemma~\ref{lem_lr_grid}.

The terminal correction stage begins at
\[
t_{\ell_0}
=
\frac{\lceil L(1-L^{-1/4})\rceil}{L}.
\]

Importantly, the proof retains the
exponential contraction weights in the
entropy recursion
(\ref{eq_lr_oracle_entropy_recursion}).
Consequently, the convergence argument
does not require an additional condition
that the unweighted sum
$\sum_{\ell=\ell_0}^{L-2}\eta_\ell^2$
tend to zero.

\begin{corollary}[Bound in terms of FM population excess risk]
\label{cor_lr_population_fm}
Under the assumptions of
Theorem~\ref{thm_lr_conditional_tv}
and Assumptions~\ref{ass_lr_density_transfer}
and \ref{ass_lr_time_transfer}, the cumulative
oracle-trajectory risk defined in
(\ref{eq_lr_trajectory_risk}) satisfies
\begin{equation}
\label{eq_lr_risk_conversion}
\mathcal R_{N,L}^{*}
\leq
\Lambda_{\mathcal D,L}
C_{\mathrm{time}}
n_L\mathcal E_{N}^{\mathrm{FM}}.
\end{equation}

Consequently, for $\delta_L=L^{-1/4}$
and sufficiently large $L$,
\begin{equation}
\label{eq_lr_tv_population_fm}
\begin{aligned}
\mathbb E_{\mathcal D}\mathbb E_Z
d_{\mathrm{TV}}
\left(
\widehat P_{Y\mid Z}^{\mathcal D,L},
\pi_Z
\right)
\leq
C\left[
\sqrt{\log L}\,e^{-cL^{1/4}}
+L^{-1/4}
\right]
+
\frac12
\sqrt{
\Lambda_{\mathcal D,L}
C_{\mathrm{time}}
n_L\mathcal E_{N}^{\mathrm{FM}}
}.
\end{aligned}
\end{equation}
In particular, if $\Lambda_{\mathcal D,L_N}$
remains uniformly bounded along the selected
sequence, then
\[
L_N\to\infty,
\qquad
L_N^{3/4}\mathcal E_{N}^{\mathrm{FM}}
\to0
\]
are sufficient for convergence in
expected total variation distance.
\end{corollary}

\begin{proof}
For each $\ell=\ell_0,\ldots,L-2$
and fixed training data $\mathcal D$,
define the nonnegative function
\[
f_{\ell,\mathcal D}(x)
:=
\left\|
\widehat v_{t_\ell}(x)
-v_{t_\ell}^{*}(x)
\right\|^2.
\]

Recall that the oracle velocity-field
input $\widetilde X_{t_\ell}^{*,Z}$
is defined in (\ref{eq_lr_oracle_input}),
and that $Q_{\ell,\mathcal D}^{*}$
denotes its conditional law given
$\mathcal D$, after averaging over
$Z\sim P_Z$ and the oracle sampling
randomness, as defined in
Assumption~\ref{ass_lr_density_transfer}.

By the density domination condition
(\ref{eq_lr_density_transfer}), for
almost every $\mathcal D$,
\[
\begin{aligned}
\mathbb E_Z
\mathbb E_{\widetilde X_{t_\ell}^{*,Z}\mid\mathcal D,Z}
f_{\ell,\mathcal D}
(\widetilde X_{t_\ell}^{*,Z})
&=
\int_{\mathbb R^p}
f_{\ell,\mathcal D}(x)
Q_{\ell,\mathcal D}^{*}(dx)\\
&\leq
\Lambda_{\mathcal D,L}
\int_{\mathbb R^p}
f_{\ell,\mathcal D}(x)
p_{t_\ell}(x)\,dx.
\end{aligned}
\]

Since $\Lambda_{\mathcal D,L}$ is a
deterministic bound that holds uniformly
over almost every training realization,
we may take expectations over
$\mathcal D$ and obtain
\[
\begin{aligned}
\mathbb E_{\mathcal D}\mathbb E_Z
\mathbb E_{\widetilde X_{t_\ell}^{*,Z} \mid\mathcal D,Z}
f_{\ell,\mathcal D}
(\widetilde X_{t_\ell}^{*,Z})
&\leq
\Lambda_{\mathcal D,L}
\mathbb E_{\mathcal D}
\int_{\mathbb R^p}
f_{\ell,\mathcal D}(x)
p_{t_\ell}(x)\,dx\\
&=
\Lambda_{\mathcal D,L}
g_{N}(t_\ell),
\end{aligned}
\]
where the last equality follows from
the population risk definition
(\ref{eq_lr_population_risk})
in Assumption~\ref{ass_lr_time_transfer}.

Summing over the terminal correction
steps and using the definition of
$\mathcal R_{N,L}^{*}$
in (\ref{eq_lr_trajectory_risk}), we have
\[
\begin{aligned}
\mathcal R_{N,L}^{*}
&=
\sum_{\ell=\ell_0}^{L-2}
\mathbb E_{\mathcal D}\mathbb E_Z
\mathbb E_{*\mid\mathcal D,Z}
f_{\ell,\mathcal D}
(\widetilde X_{t_\ell}^{*,Z})\\
&\leq
\Lambda_{\mathcal D,L}
\sum_{\ell=\ell_0}^{L-2}
g_{N}(t_\ell).
\end{aligned}
\]
Applying the terminal-time risk
comparison bound
(\ref{eq_lr_time_transfer}) gives
\[
\begin{aligned}
\mathcal R_{N,L}^{*}
&\leq
\Lambda_{\mathcal D,L}
\sum_{\ell=\ell_0}^{L-2}
g_{N}(t_\ell)\\
&\leq
\Lambda_{\mathcal D,L}
C_{\mathrm{time}}
n_L\mathcal E_{N}^{\mathrm{FM}}.
\end{aligned}
\]
This proves (\ref{eq_lr_risk_conversion}).

Next, substituting
(\ref{eq_lr_risk_conversion})
into the conditional TV bound
(\ref{eq_lr_main_tv_quarter})
in Theorem~\ref{thm_lr_conditional_tv}
yields
\[
\begin{aligned}
\mathbb E_{\mathcal D}\mathbb E_Z
d_{\mathrm{TV}}
\left(
\widehat P_{Y\mid Z}^{\mathcal D,L},
\pi_Z
\right)
\leq
C\left[
\sqrt{1+\log L}\,
e^{-cL^{1/4}}
+L^{-1/4}
\right] +
\frac12
\sqrt{
\Lambda_{\mathcal D,L}
C_{\mathrm{time}}
n_L\mathcal E_{N}^{\mathrm{FM}}
}.
\end{aligned}
\]
Since
$\sqrt{1+\log L}\leq C\sqrt{\log L}$
for sufficiently large $L$, this proves
(\ref{eq_lr_tv_population_fm}).

Finally, Lemma~\ref{lem_lr_grid}
gives $n_L\asymp L^{3/4}$ when
$\delta_L=L^{-1/4}$.
If $\Lambda_{\mathcal D,L_N}$
is uniformly bounded, then
\[
\begin{aligned}
\Lambda_{\mathcal D,L_N}
C_{\mathrm{time}}
n_{L_N}\mathcal E_{N}^{\mathrm{FM}}\leq
C L_N^{3/4}
\mathcal E_{N}^{\mathrm{FM}}
\longrightarrow0.
\end{aligned}
\]
The remaining terms in
(\ref{eq_lr_tv_population_fm})
vanish as $L_N\to\infty$.
Hence,
\[
\mathbb E_{\mathcal D}\mathbb E_Z
d_{\mathrm{TV}}
\left(
\widehat P_{Y\mid Z}^{\mathcal D,L_N},
\pi_Z
\right)
\longrightarrow0,
\]
which establishes the stated
consistency condition.
\end{proof}

\paragraph{A conditional sample-size consequence.}

Suppose that an applicable
learning theorem establishes a deterministic
upper bound on the FM population excess
risk defined in (\ref{eq_lr_population_risk}):
\[
\mathcal E_{N}^{\mathrm{FM}}
\leq a_N,
\qquad
a_N>0,
\qquad
a_N\to0,
\]
where $N$ denotes the training sample size.
Assume further that the constants in the
risk-comparison assumptions, including
$\Lambda_{\mathcal D,L_N}$, remain uniformly
bounded along the selected sequence.

By the TV bound
(\ref{eq_lr_tv_population_fm}) and
$n_L\asymp L^{3/4}$ from
Lemma~\ref{lem_lr_grid}, we obtain
\[
\begin{aligned}
\mathbb E_{\mathcal D}\mathbb E_Z
d_{\mathrm{TV}}
\left(
\widehat P_{Y\mid Z}^{\mathcal D,L},
\pi_Z
\right)\leq
C\left[
\sqrt{\log L}\,e^{-cL^{1/4}}
+L^{-1/4}
+L^{3/8}\sqrt{a_N}
\right].
\end{aligned}
\]

The two leading non-exponential terms are
\[
L^{-1/4}
\quad\text{and}\quad
L^{3/8}\sqrt{a_N}.
\]
Balancing these terms gives
\[
L^{-1/4}
\asymp
L^{3/8}\sqrt{a_N},
\]
which is equivalent to
\[
L^{5/8}\asymp a_N^{-1/2}.
\]
We therefore choose
\begin{equation}
\label{eq_lr_optimal_steps}
L_N\asymp a_N^{-4/5}.
\end{equation}

With this choice,
\[
L_N^{-1/4}\asymp a_N^{1/5},
\qquad
L_N^{3/8}\sqrt{a_N}
\asymp a_N^{1/5}.
\]
Moreover, the exponential term is
asymptotically negligible.
Consequently,
\begin{equation}
\label{eq_lr_sample_rate}
\begin{aligned}
\mathbb E_{\mathcal D}\mathbb E_Z
d_{\mathrm{TV}}
\left(
\widehat P_{Y\mid Z}^{\mathcal D,L_N},
\pi_Z
\right)=O(a_N^{1/5}).
\end{aligned}
\end{equation}

For example, if the FM population excess
risk satisfies
\[
\mathcal E_{N}^{\mathrm{FM}}
=O(N^{-\kappa})
\]
for some $\kappa>0$, choosing
\[
L_N\asymp N^{4\kappa/5}
\]
gives
\[
\begin{aligned}
\mathbb E_{\mathcal D}\mathbb E_Z
d_{\mathrm{TV}}
\left(
\widehat P_{Y\mid Z}^{\mathcal D,L_N},
\pi_Z
\right) =O(N^{-\kappa/5}).
\end{aligned}
\]

These rates are conditional consequences
of an independently established FM
population learning rate and the
density-transfer and time-transfer
assumptions. They do not assert that
an arbitrary FM training procedure
automatically satisfies these conditions.

\paragraph{Extraction of the coordinate required by FMCIT.}

For a fixed conditional-sampling query
with conditioning set $S$, let
$Z=X_S$ and $Y=X_{S^c}$.
For any $i\in S^c$, let
$\operatorname{pr}_i:\mathbb R^d\to\mathbb R$
denote the coordinate projection that
extracts $X_i$ from $Y$.

Since $\operatorname{pr}_i$ is measurable,
total variation distance cannot increase
under this projection.
In particular, for every fixed
$\mathcal D$ and $z$,
\[
\begin{aligned}
d_{\mathrm{TV}}
\left(
P_{X_i\mid X_S=z},
\widehat P_{X_i\mid X_S=z}^{\mathcal D,L}
\right)
&=
d_{\mathrm{TV}}
\left(
(\operatorname{pr}_i)_\#\pi_z,
(\operatorname{pr}_i)_\#
\widehat P_{Y\mid Z=z}^{\mathcal D,L}
\right)\\
&\leq
d_{\mathrm{TV}}
\left(
\pi_z,
\widehat P_{Y\mid Z=z}^{\mathcal D,L}
\right).
\end{aligned}
\]

Taking expectations over $\mathcal D$
and $X_S$ yields
\begin{equation}
\label{eq_lr_coordinate_tv}
\begin{aligned}
\mathbb E_{\mathcal D}\mathbb E_{X_S}
d_{\mathrm{TV}}
\left(
P_{X_i\mid X_S},
\widehat P_{X_i\mid X_S}^{\mathcal D,L}
\right) \leq
\mathbb E_{\mathcal D}\mathbb E_Z
d_{\mathrm{TV}}
\left(
\pi_Z,
\widehat P_{Y\mid Z}^{\mathcal D,L}
\right).
\end{aligned}
\end{equation}

Therefore, the conditional-sampling
bounds in
Theorem~\ref{thm_lr_conditional_tv}
and Corollary~\ref{cor_lr_population_fm}
also apply to the coordinate-level
conditional distributions required by
FMCIT.

This conclusion concerns a fixed
conditional-sampling query.
Additional uniformity or selection
conditions are required when applying
these bounds to an adaptively selected
sequence of conditional independence tests.

\subsection{Proof of Theorem \ref{thm_FMCIT_typeI}}
\begin{proof}
Under $H_0$, since the training dataset
$\mathcal D$ is independent of the CRT samples, the type-I error
bound of the CRT
\cite{berrett2020conditional} gives
\begin{equation}
\begin{aligned}
\mathbb P(p_{\mathrm{FMCIT}}\leq\alpha)
\leq \alpha
+n\mathbb E_{\mathcal D}\mathbb E_{X_S}
d_{\mathrm{TV}}\left(
P_{X_i\mid X_S},
\widehat P_{X_i\mid X_S}^{\mathcal D,L}
\right).
\end{aligned}
\end{equation}
By \eqref{eq_lr_coordinate_tv}, we further have
\begin{equation}
\begin{aligned}
\mathbb P(p_{\mathrm{FMCIT}}\leq\alpha)
&\leq \alpha
+n\mathbb E_{\mathcal D}\mathbb E_Z
d_{\mathrm{TV}}\left(
\pi_Z,
\widehat P_{Y\mid Z}^{\mathcal D,L}
\right)\\
&=\alpha
+n\mathbb E_{\mathcal D}\mathbb E_{X_S}
d_{\mathrm{TV}}\left(
P_{X_{S^c}\mid X_S},
\widehat P_{X_{S^c}\mid X_S}^{\mathcal D,L}
\right),
\end{aligned}
\end{equation}
where $Y=X_{S^c}$ and $Z=X_S$.
This completes the proof.
\end{proof}


\section{Empirical Performance of Langevin-RePaint Sampler}
\label{sec_empiri_sampler}

To empirically evaluate the effectiveness of the proposed
Langevin-RePaint sampler, we conduct an image imputation
experiment on the ImageNet validation set using the
JiT model~\cite{li_2026_jit}.
We compare the original RePaint sampler with
Langevin-RePaint, which introduces the proposed
half-score correction during the terminal sampling stage.

\paragraph{Experimental setup.}
We use the pretrained JiT-H/16 model with the
\texttt{model\_ema1} checkpoint.
The experiment covers all 50,000 images in the
ImageNet validation set at a resolution of
$256\times256\times3$.
For each image, we randomly mask $40\%$ of the spatial
pixels (26,214 pixels).
Different images are assigned different random masks.
We use 200 sampling steps.

We evaluate two sampling configurations:
the original RePaint sampler without half-score
correction (\ref{eq_repaint_update}) and the proposed Langevin-RePaint sampler
with half-score correction (\ref{eq_langevin_repaint_update}).
For Langevin-RePaint, the half-score correction
is activated when
\[
t\geq1-L^{-1/4}
=1-200^{-1/4}
\approx0.7341.
\]
All other sampling settings are kept identical
between the two configurations.

\begin{table}[t]
\centering
\caption{ImageNet-256 imputation results
(mean $\pm$ std). The paired win rate indicates
the percentage of images where Langevin-RePaint
outperforms RePaint.}
\label{tab_lr_imagenet_imputation}
\begin{tabular}{lccc}
\toprule
Method
& LPIPS ($\times 10^{-2}$) $\downarrow$
& PSNR $\uparrow$
& SSIM ($\times 10^{-1}$) $\uparrow$ \\
\midrule
RePaint
& $3.073\pm2.377$
& $27.118\pm5.213$
& $\mathbf{9.226\pm0.547}$ \\
Langevin-RePaint (Ours)
& $\mathbf{2.621\pm2.040}$
& $\mathbf{27.550\pm4.980}$
& $9.211\pm0.536$ \\
\midrule
Paired win rate (Ours)
& $90.892\%$
& $81.282\%$
& $34.008\%$ \\
\bottomrule
\end{tabular}
\end{table}

\begin{figure}[t]
\centering
\includegraphics[width=0.75\linewidth]{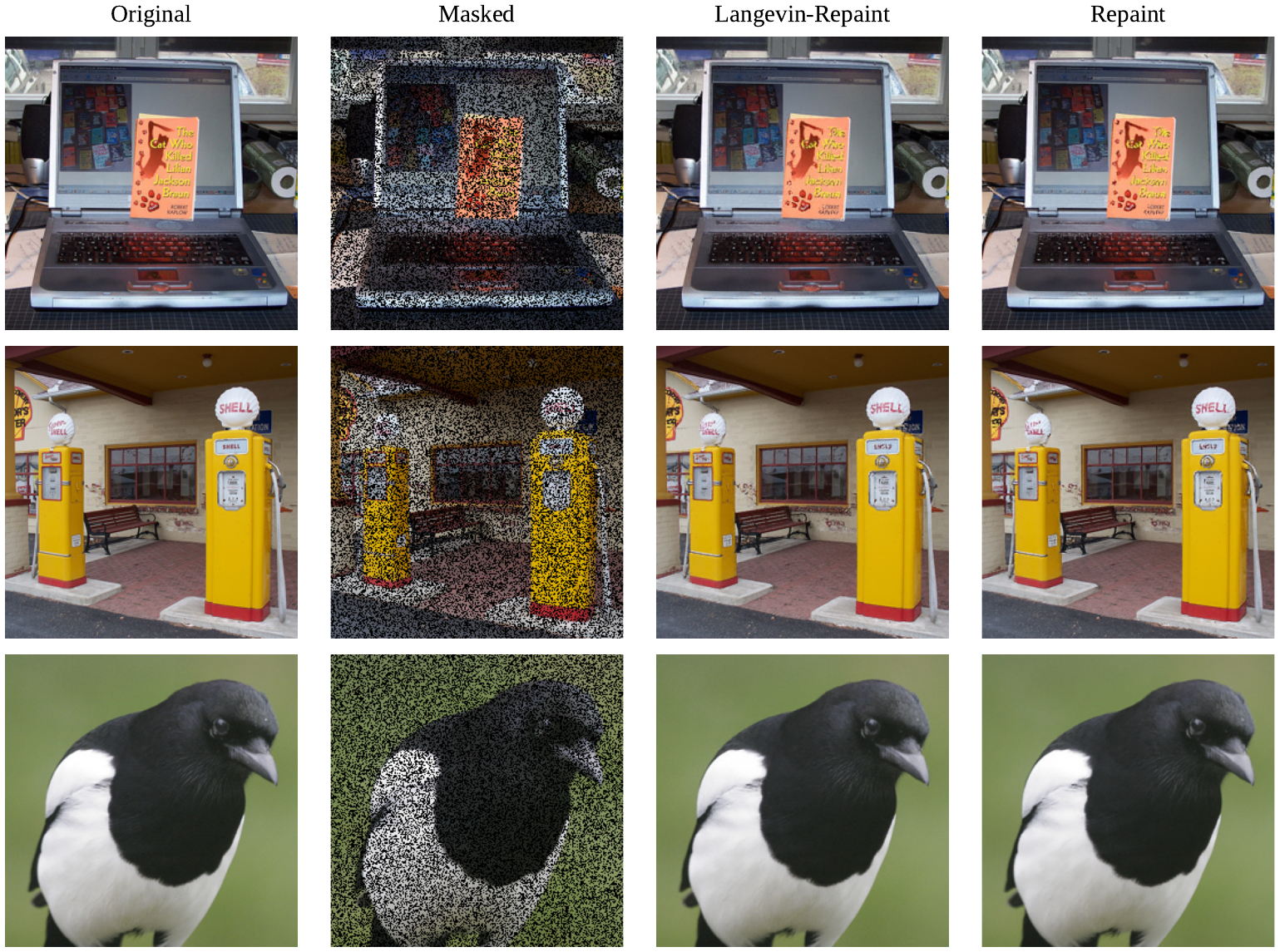}
\caption{Comparison of
Langevin-RePaint and RePaint on ImageNet-256 imputation
with $40\%$ random masking.}
\label{fig_lr_halfscore_comparison}
\end{figure}

\paragraph{Quantitative results.}
Table~\ref{tab_lr_imagenet_imputation} reports
the imputation performance on all 50,000
ImageNet validation images.
Langevin-RePaint reduces the mean LPIPS by
$14.7\%$ and improves masked PSNR by $0.432$ dB
compared with RePaint.
It outperforms RePaint on $90.892\%$ and
$81.282\%$ of the images in LPIPS and PSNR,
respectively.
However, masked SSIM decreases slightly by
$0.001532$, with Langevin-RePaint outperforming
RePaint on only $34.008\%$ of the images.

\paragraph{Discussion.}
These results demonstrate that the proposed
half-score correction improves perceptual
similarity and pixel-level reconstruction
accuracy, at the cost of a slight reduction
in structural similarity.
Figure~\ref{fig_lr_halfscore_comparison}
provides a qualitative comparison of the
two sampling methods.
Notably, these improvements are achieved
without additional training or modifications
to the pretrained generative model.

\clearpage

\section*{NeurIPS Paper Checklist}

\begin{enumerate}

\item {\bf Claims}
    \item[] Question: Do the main claims made in the abstract and introduction accurately reflect the paper's contributions and scope?
    \item[] Answer: \answerYes{}
    \item[] Justification: The abstract and introduction summarize the proposed reusable FM-based conditional resampling framework, Langevin-Repaint, the FMCIT CI test, and its integration into PC-style causal discovery. The claims are supported by the empirical null-calibration study, synthetic causal discovery experiments, ablations, scaling studies, and the hESC case study reported in the main text and appendix. The appendix provides convergence of Langevin-Repaint and Type-I error of FMCIT.

\item {\bf Limitations}
    \item[] Question: Does the paper discuss the limitations of the work performed by the authors?
    \item[] Answer: \answerYes{}
    \item[] Justification: The paper discusses that the real-data hESC analysis is a sanity check rather than ground-truth causal validation. It also reports runtime and scaling behavior, showing that the KCI-screened variant is accuracy-oriented and computationally more expensive than FisherZ-based baselines. 

\item {\bf Theory assumptions and proofs}
    \item[] Question: For each theoretical result, does the paper provide the full set of assumptions and a complete (and correct) proof?
    \item[] Answer: \answerYes{}
    \item[] Justification: The paper provides the full set of assumptions and complete proofs in Appendix \ref{app:fmcit_theory}.

    \item {\bf Experimental result reproducibility}
    \item[] Question: Does the paper fully disclose all the information needed to reproduce the main experimental results of the paper to the extent that it affects the main claims and/or conclusions of the paper (regardless of whether the code and data are provided or not)?
    \item[] Answer: \answerYes{}
    \item[] Justification: The paper specifies the synthetic SEM generation procedure, benchmark dimensions, sample sizes, edge densities, noise distributions, compared methods, evaluation metrics, number of trials, and relevant hyperparameters. The hESC case study reports the BEELINE data source, raw matrix dimensions, preprocessing steps, gene-selection protocol, and reference-network usage.

\item {\bf Open access to data and code}
    \item[] Question: Does the paper provide open access to the data and code, with sufficient instructions to faithfully reproduce the main experimental results, as described in supplemental material?
    \item[] Answer: \answerYes{}
    \item[] Justification: The code can be found in \url{https://github.com/Yanfeng-Yang-0316/GPC-FMCIT/}.

\item {\bf Experimental setting/details}
    \item[] Question: Does the paper specify all the training and test details (e.g., data splits, hyperparameters, how they were chosen, type of optimizer) necessary to understand the results?
    \item[] Answer: \answerYes{}
    \item[] Justification: The appendix specifies the flow-matching architecture, optimizer, learning rate, batch size, number of epochs, CRT randomization count, sampling steps, CI statistics, screening thresholds, refinement thresholds, conditioning-depth settings, and graph-search hyperparameters used in the experiments.

\item {\bf Experiment statistical significance}
    \item[] Question: Does the paper report error bars suitably and correctly defined or other appropriate information about the statistical significance of the experiments?
    \item[] Answer: \answerYes{}
    \item[] Justification: The main causal discovery results are reported as mean \(\pm\) standard deviation over ten trials, and the component ablations, dimensional scaling, and sample-size sensitivity studies report mean \(\pm\) standard deviation over multiple independent trials. The tables state the number of trials used for each experiment.

\item {\bf Experiments compute resources}
    \item[] Question: For each experiment, does the paper provide sufficient information on the computer resources (type of compute workers, memory, time of execution) needed to reproduce the experiments?
    \item[] Answer: \answerYes{}
    \item[] Justification: The computing environment and runtime measurement protocol are reported in Appendix~\ref{app:compute_environment}. The paper specifies the CPU, GPU, system memory, GPU memory, software versions, and wall-clock runtime measurement protocol. The experimental tables report per-run runtimes, and for repeated experiments the total compute can be estimated by multiplying the reported runtime by the number of independent trials.

\item {\bf Code of ethics}
    \item[] Question: Does the research conducted in the paper conform, in every respect, with the NeurIPS Code of Ethics \url{https://neurips.cc/public/EthicsGuidelines}?
    \item[] Answer: \answerYes{}
    \item[] Justification: The work develops algorithms for conditional independence testing and causal discovery using synthetic data and a public processed gene-expression benchmark. We are not aware of any aspect of the research that violates the NeurIPS Code of Ethics.

\item {\bf Broader impacts}
    \item[] Question: Does the paper discuss both potential positive societal impacts and negative societal impacts of the work performed?
    \item[] Answer: \answerYes{}
    \item[] Justification: The work is foundational methodological research. Its main potential positive impact is to support nonlinear conditional independence testing and causal structure exploration in scientific data analysis. The paper does not involve automated decision making, sensitive personal data modeling, or systems intended for public deployment, so we view its direct societal risk as low. For real observational data, the learned CPDAG should be interpreted as an exploratory graph estimate rather than as an unvalidated true causal mechanism. 

\item {\bf Safeguards}
    \item[] Question: Does the paper describe safeguards that have been put in place for responsible release of data or models that have a high risk for misuse (e.g., pre-trained language models, image generators, or scraped datasets)?
    \item[] Answer: \answerNA{}
    \item[] Justification: The paper does not release high-risk pretrained generative models, scraped datasets, or dual-use foundation models. The empirical evaluation uses synthetic data and a public processed gene-expression benchmark.

\item {\bf Licenses for existing assets}
    \item[] Question: Are the creators or original owners of assets (e.g., code, data, models), used in the paper, properly credited and are the license and terms of use explicitly mentioned and properly respected?
    \item[] Answer: \answerYes{}
    \item[] Justification: The paper cites the BEELINE benchmark paper for the processed hESC single-cell gene-expression data and ChIP-seq-derived reference network. The dataset is used only for research benchmarking and external-reference sanity checks, and no new redistributed version of the dataset is introduced in the paper.

\item {\bf New assets}
    \item[] Question: Are new assets introduced in the paper well documented and is the documentation provided alongside the assets?
    \item[] Answer: \answerNA{}
    \item[] Justification: The paper does not introduce a new dataset or pretrained model asset. The main contribution is an algorithmic framework and empirical evaluation using synthetic data and an existing public benchmark.

\item {\bf Crowdsourcing and research with human subjects}
    \item[] Question: For crowdsourcing experiments and research with human subjects, does the paper include the full text of instructions given to participants and screenshots, if applicable, as well as details about compensation (if any)? 
    \item[] Answer: \answerNA{}
    \item[] Justification: The paper does not involve crowdsourcing experiments or direct research with human subjects.
    
\item {\bf Institutional review board (IRB) approvals or equivalent for research with human subjects}
    \item[] Question: Does the paper describe potential risks incurred by study participants, whether such risks were disclosed to the subjects, and whether Institutional Review Board (IRB) approvals (or an equivalent approval/review based on the requirements of your country or institution) were obtained?
    \item[] Answer: \answerNA{}
    \item[] Justification: The paper does not involve new data collection from human subjects. The real-data experiment uses a previously processed public gene-expression benchmark.

\item {\bf Declaration of LLM usage}
    \item[] Question: Does the paper describe the usage of LLMs if it is an important, original, or non-standard component of the core methods in this research? Note that if the LLM is used only for writing, editing, or formatting purposes and does \emph{not} impact the core methodology, scientific rigor, or originality of the research, declaration is not required.
    \item[] Answer: \answerNA{}
    \item[] Justification: LLMs are not used as part of the proposed method, experiments, data generation, evaluation, or scientific analysis.

\end{enumerate}
\end{document}